\documentclass{article}
\usepackage{float}
\usepackage[final]{murad_style}
\usepackage[table]{xcolor}
\usepackage{booktabs}       
\usepackage{amsfonts}       
\usepackage{nicefrac}       
\usepackage{microtype}      
\usepackage{hyperref}       
\hypersetup{
    colorlinks=true,
    urlcolor=blue,
}
\usepackage{url}            
\usepackage{amsmath}
\usepackage{amssymb}        
\usepackage{graphicx}  
\usepackage{bbm}
\usepackage{siunitx}
\usepackage{subfig}
\usepackage{multicol}
\usepackage{multirow}
\title{Probabilistic Deep Learning to Quantify Uncertainty in Air Quality Forecasting}
\renewcommand\footnotemark{}
\author{
  Abdulmajid Murad$^{1}$\thanks{{\small \texttt{$^{1}$\{abdulmajid.a.murad, kraemer, kerstin.bach\}@ntnu.no}}}  
  \enskip Frank Alexander Kraemer$^{1}$
  \enskip Kerstin Bach$^{1}$
  \enskip Gavin Taylor$^{2}$\thanks{{\small \texttt{$^{2}$\{taylor\}@usna.edu}}} \\
  \AND
  \\ \vspace{-3em} \\
  $^{1}$Norwegian University of Science and Technology, \quad $^{2}$United States Naval Academy
}

\begin{document}
\maketitle

\begin{abstract}

Data-driven forecasts of air quality have recently achieved more accurate short-term predictions. Despite their success, most of the current data-driven solutions lack proper quantifications of model uncertainty that communicate how much to trust the forecasts. Recently, several practical tools to estimate uncertainty have been developed in probabilistic deep learning. However, there have not been empirical applications and extensive comparisons of these tools in the domain of air quality forecasts. Therefore, this work applies state-of-the-art techniques of uncertainty quantification in a real-world setting of air quality forecasts. Through extensive experiments, we describe training probabilistic models and evaluate their predictive uncertainties based on empirical performance, reliability of confidence estimate, and practical applicability. We also propose improving these models using "free" adversarial training and exploiting temporal and spatial correlation inherent in air quality data. Our experiments demonstrate that the proposed models perform better than previous works in quantifying uncertainty in data-driven air quality forecasts. Overall, Bayesian neural networks provide a more reliable uncertainty estimate but can be challenging to implement and scale. Other scalable methods, such as deep ensemble, Monte Carlo (MC) dropout, and stochastic weight averaging-Gaussian (SWAG), can perform well if applied correctly but with different tradeoffs and slight variations in performance metrics. Finally, our results show the practical impact of uncertainty estimation and demonstrate that, indeed, probabilistic models are more suitable for making informed decisions. Code and dataset are available at \url{https://github.com/Abdulmajid-Murad/deep_probabilistic_forecast}.
\end{abstract}

\section{Introduction}
Monitoring and forecasting real-world phenomena are fundamental use cases for many practical applications in the Internet of Things (IoT). For example, policymakers in municipalities can use forecasts of ambient air quality to make decisions about actions, such as informing the public or starting emission-reduction measures. The problem is that forecasts are both uncertain and intended for human interpretation, so decisions about specific actions should take forecast confidence into account. For instance, it may be best to only start a costly initiative for cleaning streets of dust when the forecast of air pollutants exceeds a certain threshold and the reported confidence is high. Therefore, quantifying the predictive confidence is crucial for learning, providing, and interpreting reliable forecasting models.

Progress in probabilistic machine learning \cite{ghahramani2015probabilistic} and, more recently, in probabilistic deep learning led to the development of practical tools to estimate uncertainty about models and predictions \cite{mackay1992practical,blundell2015weight, gal2016dropout, lakshminarayanan2017simple, zhu2017deep, maddox2019simple}. These tools have been successfully used in various domains, such as computer vision \cite{kendall2018multi, kendall2017uncertainties}, language modeling \cite{chien2016bayesian, xiao2019quantifying}, machine translation \cite{ott2018analyzing} and autonomous driving \cite{meyer2020learning}.
All of these tools address quantifying predictive uncertainty but differ in techniques, approximations, and assumptions when representing and manipulating uncertainty.

The successful application of probabilistic models to real human problems requires us to bridge the gap from the theory of these disparate approaches to practical concerns.  In particular, we need an empirical evaluation and comparison of these many techniques. We want to know how they perform with respect to prediction accuracy, uncertainty quantification, and other requirements. Specifically, we want to know the reliability of their confidence estimate, meaning if they are actually more accurate and trustworthy when their confidence is high. This is practically important for policymakers to make risk-informed decisions. Such an empirical evaluation, especially in the domain of air quality, is currently lacking.

\begin{figure}[!htp]
    \centering
    \includegraphics[width=0.8\linewidth]{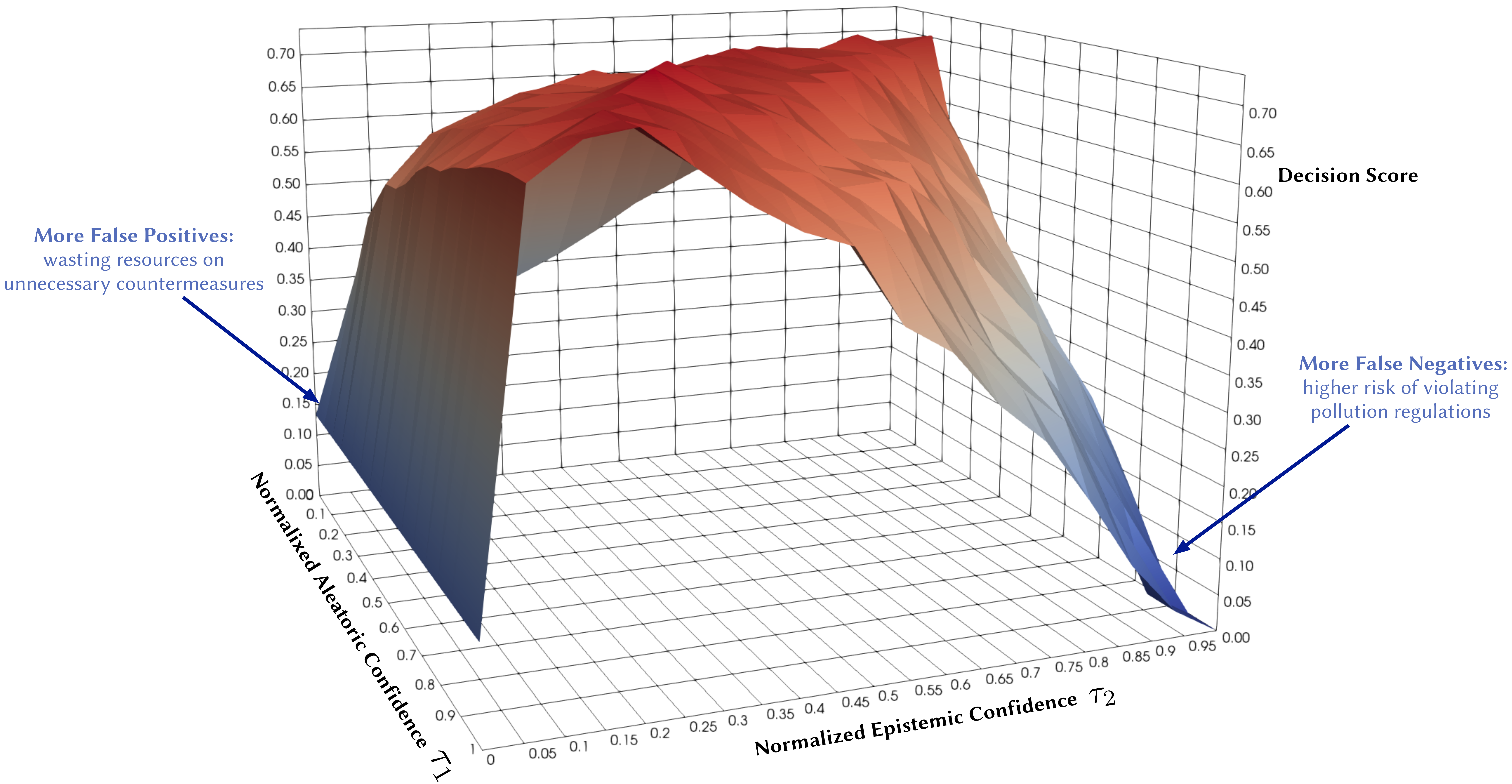}
    \caption{Illustrative example: decision F1 score as a function of normalized aleatoric and epistemic confidence thresholds. }
    \label{fig:epist_vs_aleat}
\end{figure}

Existing studies already incorporate aspects of uncertainty into their data-driven forecasts, but they often only quantify aleatoric uncertainty. This uncertainty is inherent in the observed data and can be quantified by a distribution over the model's output using softmax or Gaussian. There is also epistemic uncertainty that we can and should quantify as well. This type of uncertainty represents how much the model does not know, for instance, in regions of the input space with little data. Probabilistic models can capture both types of uncertainties, which makes them practically appealing since they convey more information about the reliability of the forecast. They also provide a wider area of control over the decision boundary and the risk profile of a decision. For instance, Figure \ref{fig:epist_vs_aleat} illustrates the potential of quantifying both types of uncertainties in decision making. It shows the decision F1 score as a function of normalized aleatoric and epistemic confidence thresholds in a probabilistic model. Each point on the surface represents the resulting score of accepting the model predictions only where its confidence is above specific thresholds ($\tau_1$ and $\tau_2$). Expressing both uncertainty dimensions gives more control regarding false positives and false negatives and their associated risks.  We will come back to this Figure in Section \ref{sec:informed_decision_making}.

In this work, we develop a set of deep probabilistic models for air quality forecasting that quantify both aleatoric and epistemic uncertainties and study how to represent and manipulate their predictive uncertainties. Our contributions are the following:
\begin{itemize}
    \item We conduct a broad empirical comparison and exploratory assessment of state-of-the-art techniques in deep probabilistic learning applied to air quality forecasting. Through exhaustive experiments, we describe training these models and evaluating their predictive uncertainties using various metrics for regression and classification tasks.
    \item We improve uncertainty estimation using adversarial training to smooth the conditional output distribution locally around training data points.
    \item  We apply uncertainty-aware models that exploit the temporal and spatial correlation inherent in air quality data using recurrent and graph neural networks.
    \item We introduce a new state-of-the-art example for air quality forecasting by defining the problem setup and selecting proper input features and models.
\end{itemize}

Our results show that all the considered probabilistic models perform well on our application, with slight variations in different performance metrics. Bayesian neural networks perform slightly better in proper scoring rules that measure the quality of probabilistic predictions. In addition, we show that smoothing predictive distributions by adversarial training improves metrics that punish incorrect, overconfident predictions, especially in regression tasks, since the forecasted phenomena are inherently smooth.

The rest of this paper is organized as follows: In Section \ref{sec:related_work}, we outline the related work in uncertainty estimation and air quality forecasting. We then define the problem setup and introduce non-probabilistic baselines and evaluation metrics in Section \ref{sec:preliminaries}. In Section \ref{sec:deep_probabilistic_forecast}, we present deep probabilistic models applied to the air quality forecast, describe their training, evaluation, and how to improve their uncertainty estimate. We close with a comparative analysis to compare the selected  models in Section \ref{sec:discussion} and a conclusion in Section \ref{sec:conclusion}. 

\section{Related Work}
\label{sec:related_work}

Forecasting ambient air quality is crucial to support decision-making in urban management. Therefore, a sizeable body of work addresses building air quality forecasting models. For example, MACC (Monitoring Atmospheric Composition and Climate) \cite{marecal2015regional} is a European project that provides air quality analysis and forecasting services for the European Continent. It uses physics-based modeling that combines transport, chemistry, and satellite data to provide a multi-model ensemble forecast of atmospheric composition (daily forecasts with hourly outputs of 10 chemical species/aerosols). Walker et al.~\cite{walkermodel} used the output of MACC ensemble-based probabilistic forecasting of air pollution and performed statistical post-processing, calibrated predictive distribution using Box-Cox transformation for correcting the skewness of air pollution data. Additionally, they discussed model selection and verification using Akaike and Bayesian information criteria. To obtain the ensemble forecast from MACC, they introduced stochastic perturbations to the emissions. Garaud et al.~\cite{garaud2011automatic} performed a posterior calibration of multi-model ensembles using a mixed optimization algorithm to extract a sub‐ensemble.

The official air quality forecasting service in Norway \cite{norway_air_forecast_url} \footnote{\href{https://luftkvalitet.miljodirektoratet.no/kart/59/10/5/aqi}{https://luftkvalitet.miljodirektoratet.no/kart/59/10/5/aqi}} uses a Gaussian dispersion modeling that provides a 2-day hourly forecast with high-resolution coverage (between 250 and 50 m grid) over the entire country \cite{denby2020description, mu2021downscaling}. The forecast is based on weather conditions, polluting emissions, and terrain. In particular, it uses the chemical transport model uEMEP (urban European Monitoring and Evaluation Program)  \cite{denby2020description} and a road dust emission model \cite{norman2016modelling}.  Denby et al. \cite{denby2018thenorwegian} analyze the accuracy of the Norwegian air quality forecasting by comparing model calculations with measurements. They show that the model´s forecast on particle dust is marginally better than the persistent forecast and give some assumptions about why model calculations deviate from the observations.

Although physics-based models \cite{simpson2012emep, denby2020description, mu2021downscaling} can provide long-range air pollution information, they require significant domain knowledge and complex modeling of dispersion, chemical transport, and meteorological processes. Additionally, physics-based models involve structural uncertainty, low spatial resolution, and do not capture abrupt and short-term changes in air pollution. Data-driven modeling based on historical data \cite{lepperod2019air, veiga2021low, zhou2019explore} can complement physics-based modeling by learning directly from air quality measurements and providing a more reliable short-term prediction. For example, Lepperod et al. \cite{lepperod2019air} deployed stationary and mobile micro-sensor devices and used Narrowband IoT (NB-IoT) to aggregate air quality data from these sensors. Then, they applied machine learning methods to predict air quality in the next 48 hours using observations of sensors' measurements, traffic, and weather data. Zhou et al. \cite{zhou2019explore} used long short-term memory (LSTM) to forecast multi-step time-series of air quality, while Mokhtari et al. \cite{mokhtari2021uncertainty} proposed combining a convolutional neural network (CNN) with LSTM for air quality prediction and quantified the uncertainty using quantile regression and MC dropout. Tao et al. \cite{tao2019air} used 1D CNNs and a Bidirectional gated recurrent unit (GRU) for a short-term forecast of fine air particles. Pucer et al. \cite{pucer2018bayesian} used Gaussian Processes (GP) to forecast daily air-pollutant levels, and Aznarte et al. \cite{aznarte2017probabilistic} proposed using quantile regression for probabilistic forecasting of extreme nitrogen dioxide ($NO_2$) pollution.

Most of the current data-driven forecasts give point predictions of a deterministic nature. Thus, they lack useful estimates of their predictive uncertainty that convey more information about how much to trust the forecast. Recently, quantifying prediction uncertainty has garnered increasing attention in machine learning fields, including deep learning. Bayesian methods are among the most used approaches for uncertainty estimation in neural networks. Given a prior distribution over the parameters, Bayesian methods use training data to compute a posterior distribution. Using the obtained distribution, we can easily quantify the predictive uncertainty. This approach has been extended to neural networks, theoretically allowing for the accuracy of modern prediction methods with valid uncertainty measures; however, modern neural networks contain many parameters, and obtaining explicit posterior densities through Bayesian inference is computationally intractable. Instead, there exist a variety of approximation methods that estimate the posterior distributions. These methods can be decomposed into three main categories, either based on variational inference \cite{graves2011practical, blundell2015weight, louizos2017multiplicative}, Markov chain Monte Carlo (MCMC) \cite{ neal2012bayesian, welling2011bayesian, chen2014stochastic}, or Laplace approximation \cite{mackay1992practical, ritter2018scalable}. In this paper, we will use variational Bayesian methods and approximate Bayesian inference methods.

\section{Air Quality Prediction, Base Models and Metrics}
\label{sec:preliminaries}

\subsection{Problem Setup}

\begin{figure}[!htp]
    \centering
    \includegraphics[width=\linewidth]{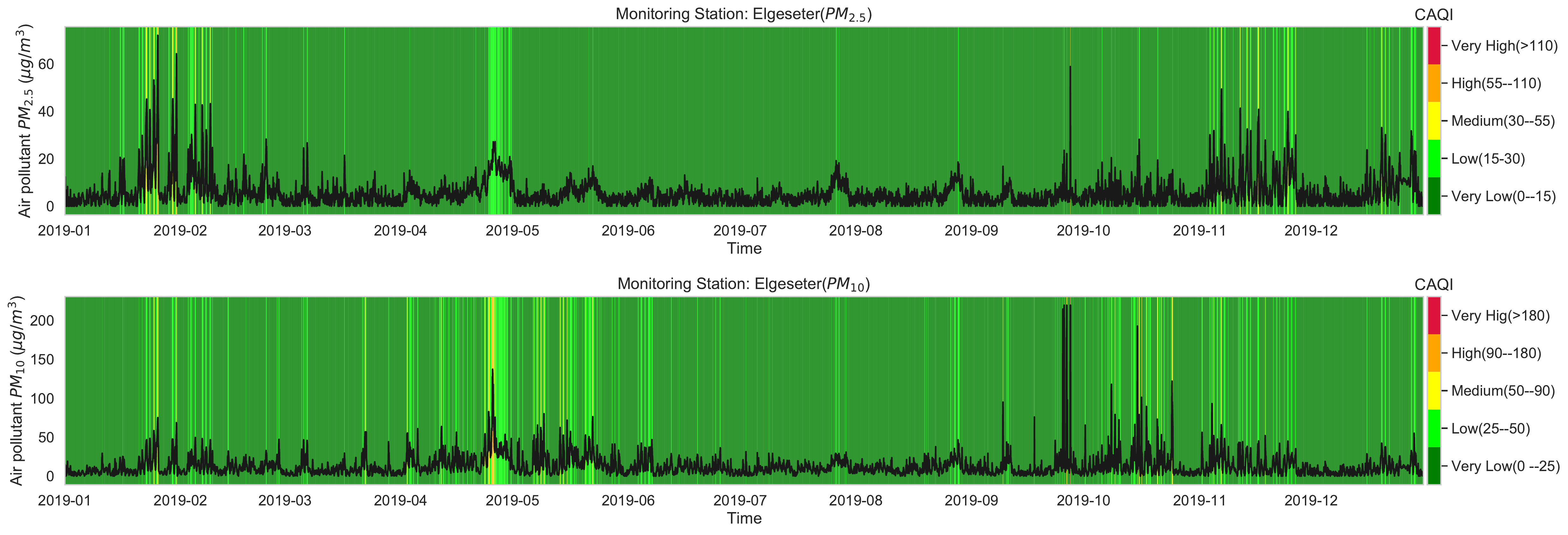}
    \caption{Air quality level over one year of one representative monitoring station in Trondheim, where the air pollutant is commonly at a \textit{Very Low} level and rarely exceeds \textit{Low}.}
    \label{fig:CAQI}
\end{figure}

We are trying to build a model for forecasting air quality trends at pre-defined locations and for a specified forecast horizon. In our case study, we want to forecast the level of microscopic particles in the air, known as particulate matter (PM). These are inhalable particles with two types: coarse particles with a diameter less than \SI{10}{\micro\metre} ($PM_{10}$) and fine particles with a diameter less than \SI{2.5}{\micro\metre} ($PM_{2.5}$). The forecast predicts pollutant levels for the next 24h at four monitoring stations in the city of Trondheim, as shown in Figure \ref{fig:NILU_Sensors}. The stakeholders, policymakers of the municipality, would like to estimate if the concentration of air particles exceeds certain thresholds following the Common Air Quality Index (CAQI) used in Europe \cite{CiteairII2012}, as shown in Table \ref{tab:CAQI}. This can be achieved through value regression or by directly classifying threshold exceedance levels. We will explore probabilistic models that forecast air quality values and predict threshold exceedance events. Using probabilistic models provides more qualitative information since decision-making largely depends on the credibility intervals of specific predictions. The forecast is based on explanatory variables, such as historical air quality measurements, meteorological data, traffic, and street-cleaning reports from the municipality. 
\definecolor{darkpastelgreen}{rgb}{0.01, 0.75, 0.24}
\definecolor{brightgreen}{rgb}{0.4, 1.0, 0.0}
\definecolor{cadmiumyellow}{rgb}{1.0, 0.96, 0.0}
\definecolor{cadmiumorange}{rgb}{0.93, 0.53, 0.18}
\definecolor{amaranth}{rgb}{0.9, 0.17, 0.31}
\begin{table}[H]
\small
\captionsetup{justification=centering}
	\begin{minipage}{0.5\linewidth}
		\centering
		\includegraphics[width=0.75\linewidth]{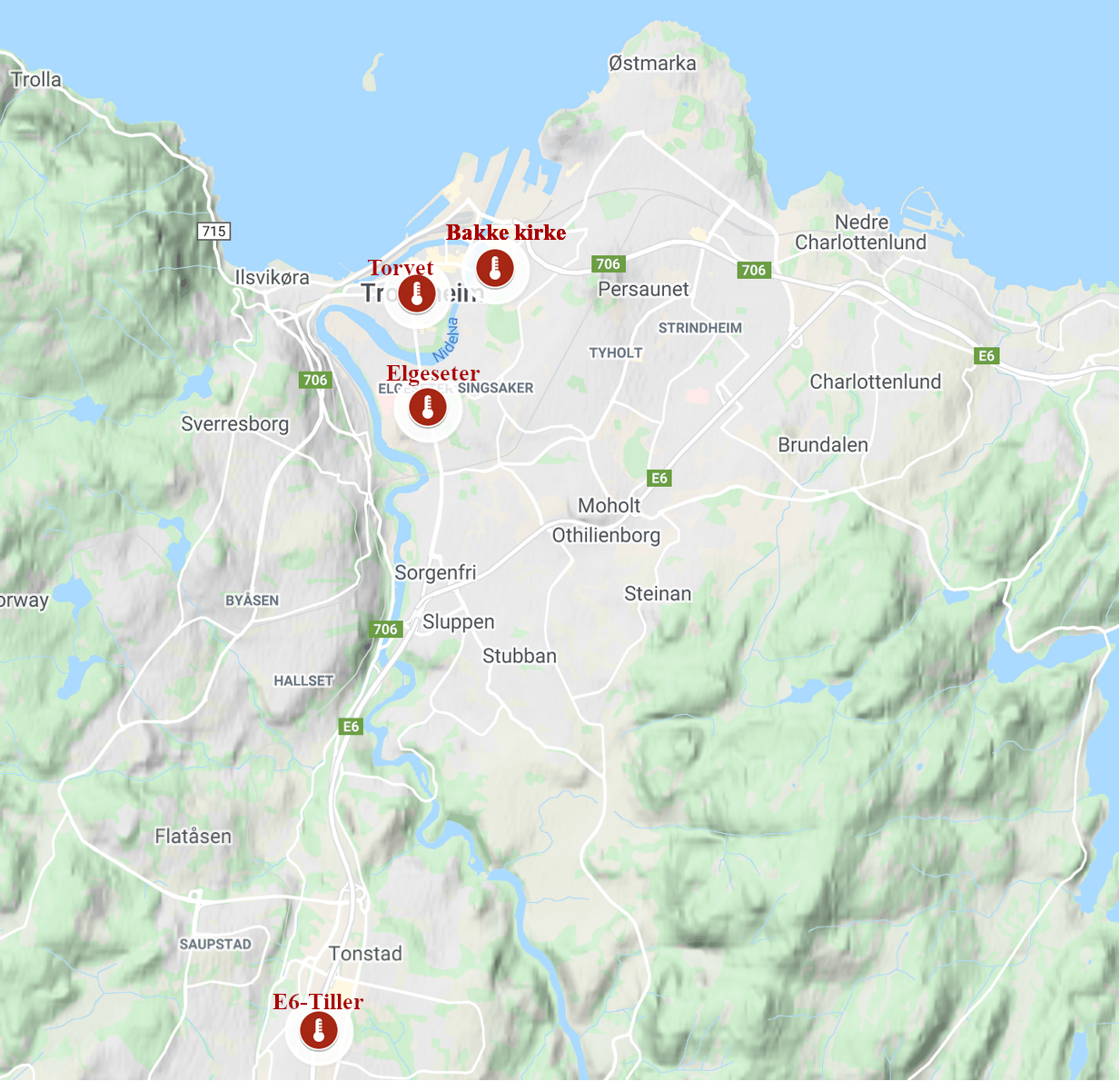}
		\captionof{figure}{Air quality monitoring stations in Trondheim, Norway.}
		\label{fig:NILU_Sensors}
	\end{minipage}
	\begin{minipage}{0.42\linewidth}
		\caption{European Common Air Quality Index}
		\label{tab:CAQI}
		\centering

            \begin{tabular}{lcc}
             \hline
            
             \bf Index & \bf $\pmb{PM_{10} (\mu g/m^3)}$  & \bf $\pmb{PM_{2.5} (\mu g/m^3)}$  \\
              \hline
                        \rowcolor{darkpastelgreen}
                         Very low  & 0–25 & 0–15 \\ 
                         \rowcolor{brightgreen}
                          Low  & 25–50 & 15–30 \\ 
                          \rowcolor{cadmiumyellow}
                          Medium  & 50-90 & 30-55 \\ 
                          \rowcolor{cadmiumorange}
                          High  & 90-180 & 55-110 \\ 
                         \rowcolor{amaranth}
                          Very High  & $>$80 & $>$110 \\ 
                         \hline
                         
            \end{tabular}
	\end{minipage}
\end{table}
Although the CAQI (Table \ref{tab:CAQI}) specifies five levels of air pollutants, the air quality in the city of Trondheim is usually at \textit{Very Low} and rarely exceeds the \textit{Low} level. 
For example, Figure \ref{fig:CAQI} shows the air quality level over one year of a representative monitoring station (Elgeseter). Therefore, instead of a multinomial classification task (with five classes), we transform the problem into a threshold exceedance forecast task in which we try to predict the points in time where the air quality exceeds the \textit{Very Low} level.

The air quality dataset we use is a part of the official measuring network in Europe. Specifically, we use the open database of air quality measurements offered by the Norwegian Institute for Air Research (NILU) \cite{nilu_url}\footnote{\href{https://www.nilu.com/open-data/}{https://www.nilu.com/open-data/}}. The meteorological data are based on historical weather and  
climate data offered by the Norwegian Meteorological Institute \cite{frost_url}\footnote{\href{https://frost.met.no}{https://frost.met.no}}. The traffic data are based on aggregated traffic volumes offered by the  
Norwegian Public Roads Administration \cite{vegvesen_url}\footnote{\href{https://www.vegvesen.no/trafikkdata/start/om-api}{https://www.vegvesen.no/trafikkdata/start/om-api}}. A more detailed description of the used datasets can be found in Appendix \ref{sec:appendix_dataset}.

\subsection{Epistemic and Aleatoric Uncertainty}

Before quantifying the predictive uncertainty, it is worth distinguishing the different sources of uncertainty and the appropriate actions to reduce them. The first source is model or epistemic uncertainty, which is uncertainty in the model parameters in regions of the input space with little data (i.e., data sparsity). This type of uncertainty can be reduced given enough data. By estimating the epistemic uncertainty of a model, we can obtain its confidence interval (CI). The second source of uncertainty is data or aleatoric uncertainty. It is essentially a noise inherent in the observations (i.e., input-dependent) due to either sensor noise or entropy in the true data generating process.
By estimating the aleatoric and epistemic uncertainties, we can obtain the prediction interval (PI) \cite{heskes1997practical, kendall2017uncertainties}. Accordingly, prediction intervals are wider than confidence intervals. The third source of uncertainty is model misspecification, i.e., uncertainty about the general structure of the model, such as model type, number of nodes, number of layers. It is also related to the bias-variance tradeoff \cite{dar2021farewell}.

\subsection{Non-probabilistic Baselines}

\begin{figure}[!htb]
    \centering
    \includegraphics[width=\linewidth]{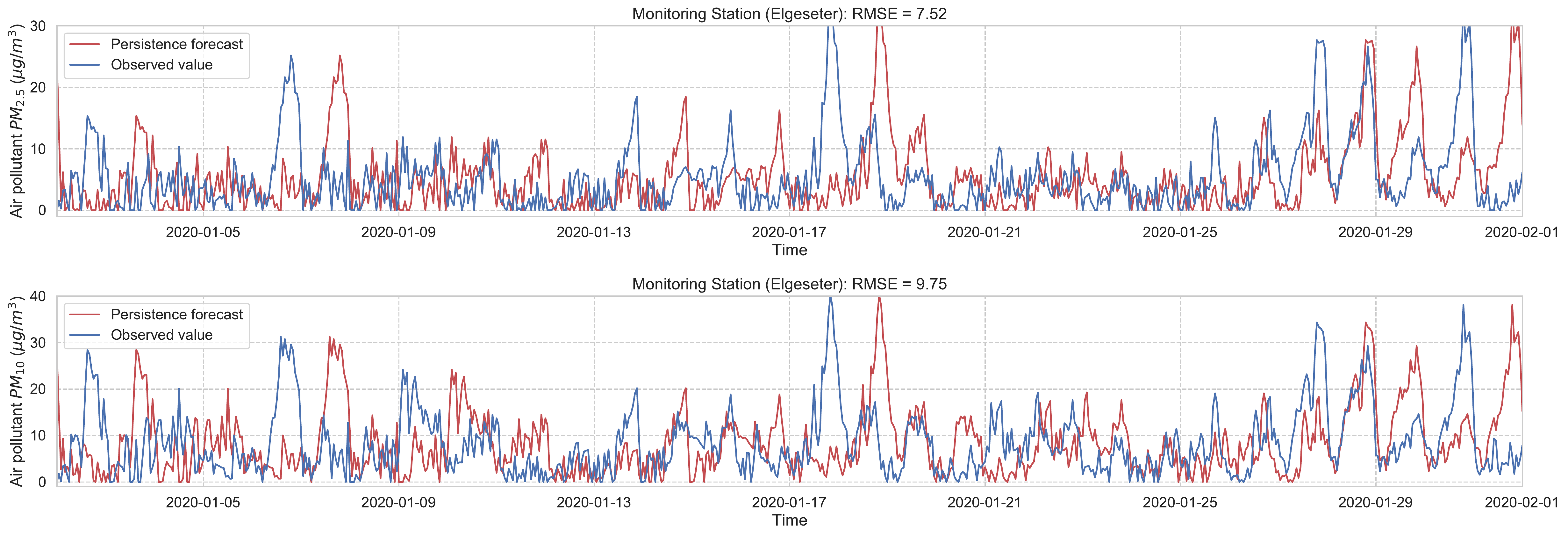}
    \caption{Persistence forecast of air pollutant over one month in one representative monitoring station.}
    \label{fig:persistance_reg}
\end{figure}

As a baseline for comparison and to motivate the use of evaluation metrics, we test the performance of non-probabilistic models, such as persistence, XGBoost, and gradient boosting models. A simple baseline predictor is a persistence forecast, which uses the diurnal patterns of the observations. To predict a value in the future, we use the value observed 24 hours earlier. Figure \ref{fig:persistance_reg} shows the results of the persistence model when forecasting the PM-value over one month (January 2020). Suppose $\hat{y}_t \in \mathbb{R}$ is the forecast value, while $y_t \in \mathbb{R}$ is the true observed value at time $t$, then we can evaluate the aggregated accuracy over a time period $T$ using the root-mean-square error:
\begin{equation}
    \mathit{RMSE} = \sqrt{\smash[b]{\frac{1}{T}\sum_{t=1}^{T} (y_t - \hat{y}_t)}}.
    \label{eq:rmse}
\end{equation}

\begin{figure}[!htb]
    \centering
    \includegraphics[width=\linewidth]{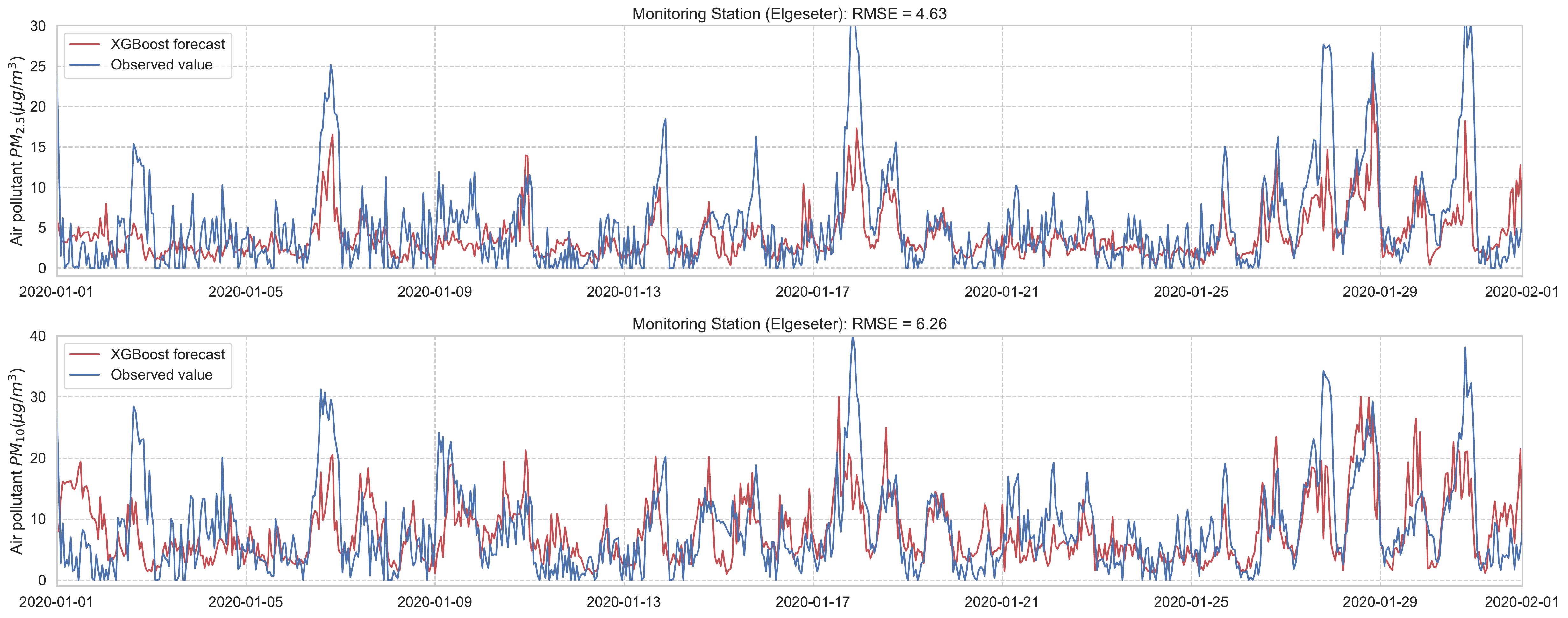}
    \caption{PM-value regression using the XGBoost model over one month in one representative monitoring station.}
    \label{fig:xgboost_reg}
\end{figure}

Next we evaluate the performance of XGBoost (\textit{eXtreme Gradient Boosting}) \cite{chen2016xgboost} as a non-probabilistic baseline in our problem setting. XGBoost uses the same model representation and inference as Random forests (i.e.,  gradient-boosted decision trees) but has a different training mechanism since it uses the second-order approximation of the training objective. We train the model over one year of data (2019) and test its performance over one month (January 2020) of a 24-hour forecast horizon. Figure \ref{fig:xgboost_reg} shows the results of the XGBoost model in one representative monitoring station. The results show an improved prediction accuracy compared to the persistence forecast, which illustrates the value of a learned predictor for time-series forecasting of air quality.

\begin{figure}[!htb]
    \centering
    \includegraphics[width=\linewidth]{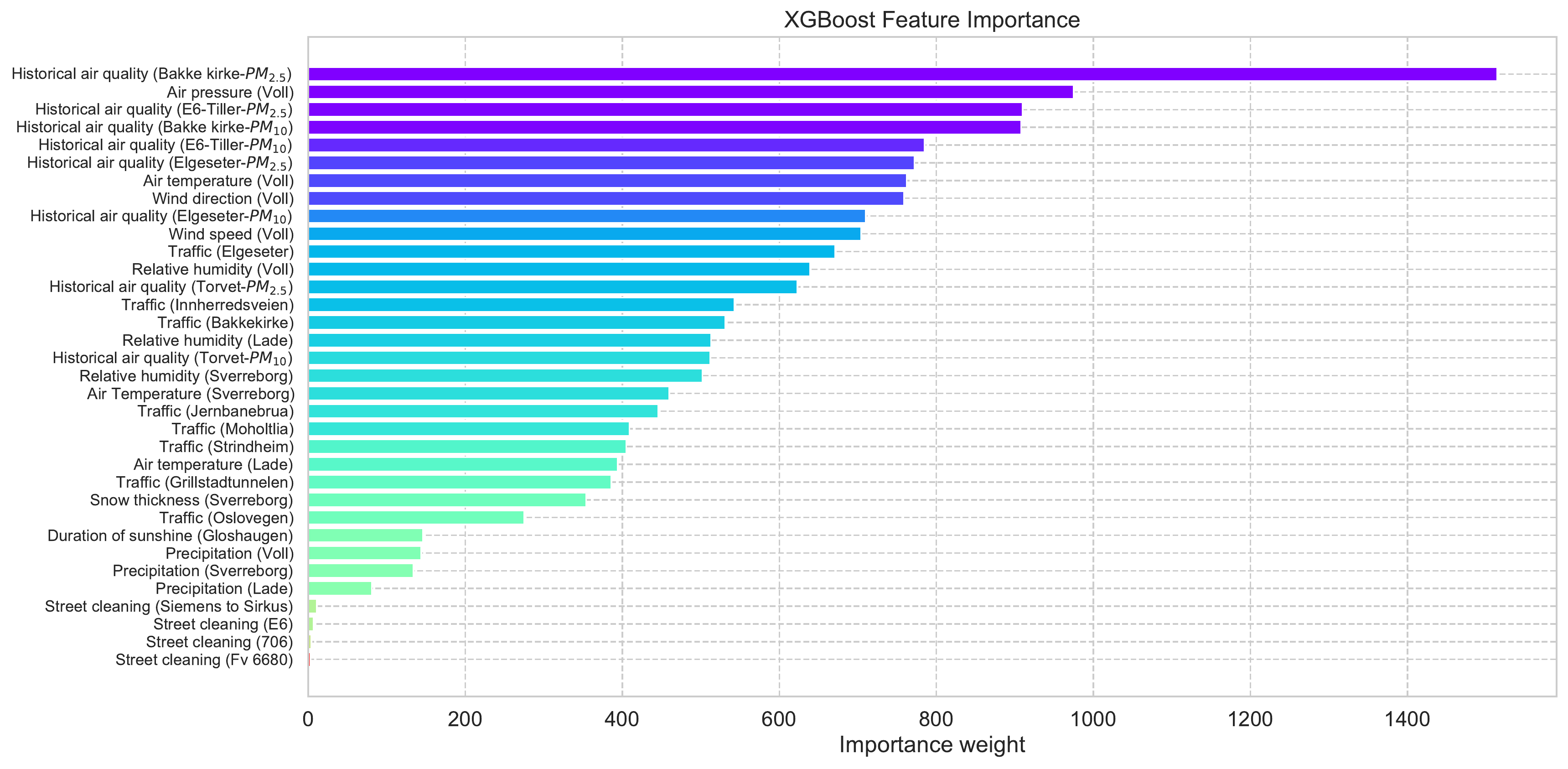}
    \caption{Feature importance of a trained XGBoost model indicating how useful each input feature when making a prediction.}
    \label{fig:xgboost_feature_importance}
\end{figure}

One advantage of using an XGBoost model is the feasibility of retrieving the feature importance by assigning a score to each input feature, which indicates how useful that feature is when making a prediction, thus contributing to prediction interpretation. Figure \ref{fig:xgboost_feature_importance} shows the feature importance of the XGBoost forecast shown in Figure \ref{fig:xgboost_reg} (a more detailed description of the features can be found in Appendix \ref{sec:appendix_dataset}.

For the threshold exceedance prediction task, we can use XGBoost to train a binary classifier to predict the probability of threshold exceedance. By using a proper scoring rule \cite{gneiting2007strictly}, such as the cross-entropy as a training criterion, we obtain predictive probabilities. Notably, these probabilities do not represent the model (epistemic) uncertainty. Rather, they represent data (aleatoric) uncertainty. In addition to cross-entropy, we need to evaluate performance using another metric that the model was not optimizing. We can use the Brier score \cite{brier1950verification}, which is a strictly proper scoring rule \cite{gneiting2007strictly} and metric to evaluate the reliability of predictive probabilities. Suppose at a time, $t$, the true class label is represented by $o_t$ (1 when pollutant level exceeds the threshold, 0 if not), while the predicted probability is represented by $\hat{p}_t \in [0, 1]$. Then we calculate the cross-entropy (CE) and Brier Score (BS) as follows:
\begin{equation}
  \mathit{CE} = - \frac{1}{T}\sum_{t=1}^{T} (o_t \times \log (\hat{p}_t))
  \label{eq:ce}
\end{equation}
\begin{equation}
  \mathit{BS} = \frac{1}{T} \sum_{t=1}^{T} (o_t - \hat{p}_t)^2
  \label{eq:bs}
\end{equation}
Both metrics heavily punish overconfident, incorrect predictions. Cross-entropy uses exponential punishment (heavily emphasizes tail probabilities) since it is a negative log-likelihood loss. Thus, it is sensitive to the predicted probabilities of the infrequent class (i.e., threshold exceedance events). In contrast, the Brier score uses quadratic punishment since it is a mean square error in the probability space. Thus, it treats the predicted probabilities of both classes equally.

\begin{figure}[!htb]
    \centering
    \includegraphics[width=\linewidth]{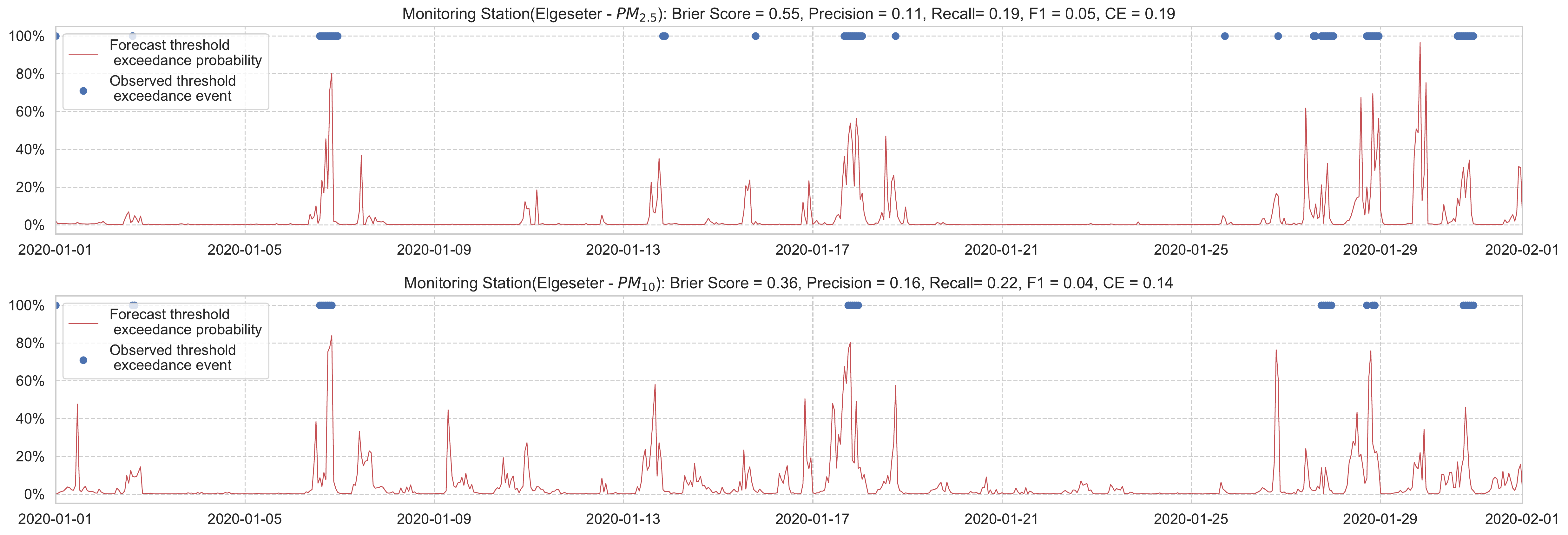}
    \caption{Predicting the threshold exceedance probability of the air pollutant level using an XGBoost model in one representative monitoring station.}
    \label{fig:xgboost_class}
\end{figure}

Additionally, we can use the commonly used metrics for classification tasks \cite{sokolova2009systematic} to evaluate the performance of a deterministic prediction. By converting the predictive probabilities into predictive class labels, we can calculate the true-positive rate $tp$, the false-positive rate $fp$, and the false-negative rate $fn$. Then, we can use the metrics of $Precision :\mathbb{R} \rightarrow [0, 1] $, $Recall :\mathbb{R} \rightarrow [0, 1]$, and $F1 :\mathbb{R} \rightarrow [0, 1]$ score to evaluate the threshold exceedance prediction as following:
\begin{equation}
    Precision = \frac{tp}{tp + fp}
    \label{eq:precision}
\end{equation}
\begin{equation}
    Recall = \frac{tp}{tp + fn}
    \label{eq:recall}
\end{equation}
\begin{equation}
    F1= \frac{tp}{tp + 0.5\times (fp + fn)}
    \label{eq:f1}
\end{equation}
Figure \ref{fig:xgboost_class} shows the results of the XGBoost model when predicting threshold exceedance probability. The blue dots indicate the points in time (hours) when air pollutants actually exceeded the threshold level. The red line represents the predicted probability in percentages. We see that with higher probability, the model predicts a more likely event of threshold exceedance at that specific time.

\subsection{Quantile Regression}
\begin{figure}[!htb]
    \centering
    \includegraphics[width=\linewidth]{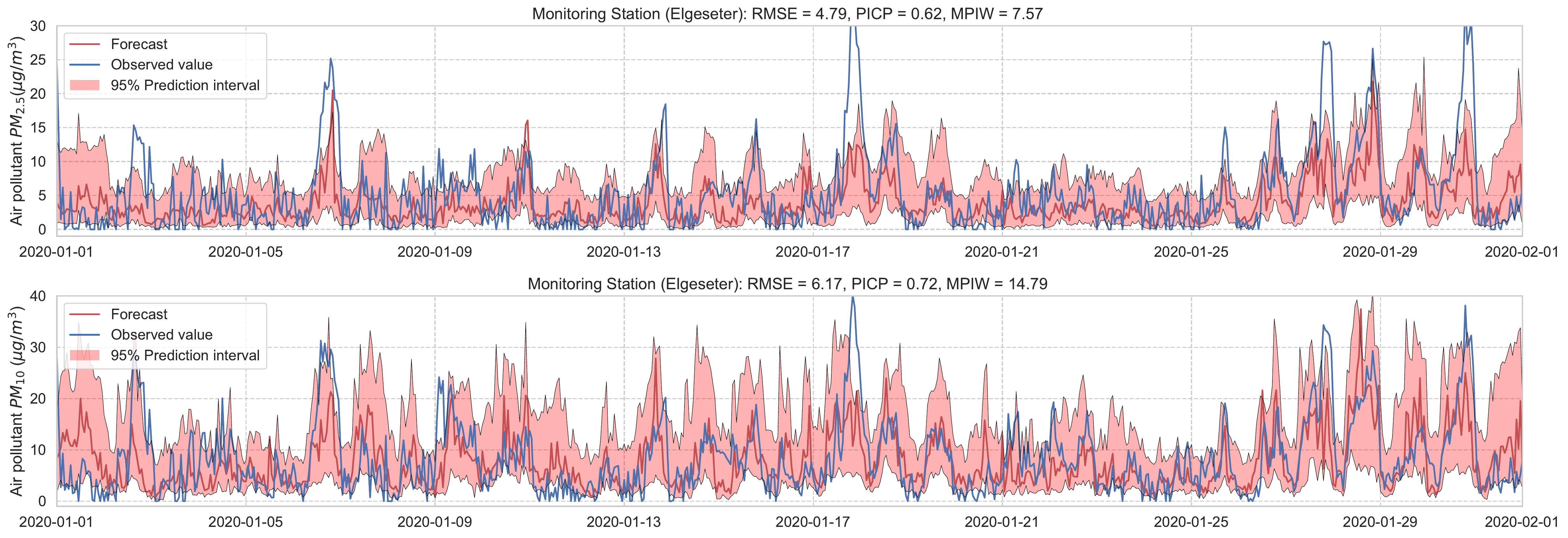}
    \caption{Air quality prediction interval using quantile regression of a Gradient Tree Boosting model.}
    \label{fig:gradient_boosting}
\end{figure}

Using an XGBoost model leads to an improved prediction accuracy (Figure \ref{fig:xgboost_reg}), but with a caveat that it only produces point estimates of the mean. To address this, we can use quantile regression \cite{koenker2001quantile} to estimate a specific percentile (i.e., quantile) of the target variable. The quantile regression estimates the conditional quantile by minimizing the absolute residuals. In contrast, regular regression estimates the conditional mean by minimizing the least squared residuals (focuses on central tendency). The advantages of using quantile regression are that it focuses more on dispersion or variability, does not assume homoscedasticity in the target variable (i.e., having the same variance independent of the input variable), and is more robust against outliers. We use a Gradient Tree Boosting model \cite{friedman2001greedy} to estimate the 5\% and 95\% percentiles, as shown in Figure \ref{fig:gradient_boosting}. 

Given a predicted lower $\hat{L}_t \in \mathbb{R} $ and upper bound $\hat{U}_t \in \mathbb{R}$, we can assess the quality of the generated prediction interval using metrics, such as Prediction Interval Coverage Probability ($PICP :\mathbb{R} \rightarrow [0, 1]$) and Mean Prediction Interval Width ($MPIW :\mathbb{R} \rightarrow [0, \infty)$):
 \begin{equation}
    \mathit{PICP} = \frac{1}{T}\sum_{t=1}^{T} \mathbbm{1}(y_{t}-\hat{L}_t) \times \mathbbm{1}(\hat{U}_t - y_{t})
    \label{eq:picp}
 \end{equation}
 \begin{equation}
    \mathit{MPIW} = \frac{1}{T}\sum_{i=1}^{T}(\hat{U}_t - \hat{L}_t)
    \label{eq:mpiw}
 \end{equation}
where $\mathbbm{1}$ is the Heaviside step function. $\mathit{PICP}$ indicates how often the prediction interval captures the true values, ranging from 0 (all outside) to 1 (all inside). Intuitively, we seek a model with a narrow prediction interval (i.e., smaller $\mathit{MPIW}$) while capturing the observed data points (i.e., larger $\mathit{PICP}$). For example, when forecasting $\mathit{PM}_{2.5}$ pollutants in Figure \ref{fig:gradient_boosting}, $\mathit{PICP} = 0.61$ indicates that in 61\% of the time-steps, the observed values are inside the predicted intervals, as compared to 72\% when forecasting $\mathit{PM}_{10}$. In contrast, the predicted intervals are smaller ($\mathit{MPIW}=7.49$) when forecasting $\mathit{PM}_{2.5}$ as compared to wider predicted intervals ($\mathit{MPIW}=14.67$) when forecasting $\mathit{PM}_{10}$. 

\section{Deep Probabilistic Forecast}
\label{sec:deep_probabilistic_forecast}

This section explores probabilistic models for air quality forecasting and describes how to quantify their predictive uncertainty. We assume we have a training dataset $\mathcal{D} = \{\mathbf{x}_t, y_t\}_{t=1}^{T}$ with an input feature $\mathbf{x}_t \in \mathbb{R}^{D}$ and a corresponding observation $y_t \in \mathbb{R}$ for each monitoring station. For the time-series forecasting task, we estimate the aleatoric uncertainty by employing the Mean-Variance Estimation method \cite{nix1994estimating}, in which we have a neural network with two output nodes (for every individual time-series) that estimates the mean $\hat\mu _t \in \mathbb{R}$ and variance $\hat\sigma_t^2 \in (0, \infty)$ of the target probability distribution. Additionally, we use the negative log-likelihood ($NLL: (0, \infty) \rightarrow (0, \infty)$) as a training criterion since the RMSE does not capture the predictive uncertainty. By treating the observed value $y_t$ as a sample from the target distribution, we can calculate $NLL$ as follows:
\begin{equation}
    \mathit{NLL} = 0.5 \left( \log{2\pi\hat\sigma_t^2 } + \frac{(y_t-\hat\mu _t)^2}{\hat\sigma_t^2} \right)
    \label{eq:nll}
\end{equation}
Notably, we treat the target distributions as having heteroscedastic uncertainty (i.e., non-constant variance) and diagonal covariance matrices, which simplifies the training criterion of multivariate time-series. For the threshold exceedance prediction task, we estimate the aleatoric uncertainty by having a neural network that outputs the predictive probability in terms of cross-entropy. 

Although epistemic uncertainty is model-dependent, it is essentially estimated by measuring the dispersion in predictions when running several inference steps over a specific data point. For the PM-value regression task, every single forward pass ($i$) outputs a normal distribution (with mean $\hat\mu_{t,i}$ and variance $\hat\sigma_{t,i}^2$). Thus, multiple forward passes result in a (uniformly weighted) mixture of normal distributions with the mean $\hat{\mu}_{t, mix} \in \mathbb{R}$ and variance $\hat\sigma_{t, mix}^2 \in (0, \infty)$ calculated as follows:
\begin{equation}
    \hat\mu_{t, mix} =\frac{1}{M}\sum_{i=1}^{M}\hat\mu _{t,i}
    \label{eq:mix_mu}
\end{equation}
\begin{equation}
 \hat\sigma_{t, mix}^2 =\frac{1}{M}\sum_{i=1}^{M}(\hat\sigma _{t,i}^2 + \hat\mu _{t,i}^2) -  \hat\mu_{t, mix}^2
 \label{eq:mix_sigma}
\end{equation}
Given a predicted mean and variance, we can then estimate a point prediction $\hat y_t = \hat\mu_{t, mix}$, a lower bound $\hat{L}_t= \hat\mu_{t, mix} - z \hat\sigma_{t, mix}$, and an upper bound $\hat{U}_t= \hat\mu_{t, mix} + z \hat\sigma_{t, mix}$, where $z$ is the standard score of a normal distribution. For example, with 95\% prediction interval (i.e., $P (L_t < y_t < U_t) = 0.95 $), we use $z =1.96$. For the threshold exceedance prediction task, every single forward pass outputs a predictive probability. Thus, we can combine predictions from multiple forward passes by simply averaging the predicted probabilities.

For empirical evaluation, we use the NLL and cross-entropy as evaluation metrics. Additionally, we use the Continuous Ranked Probability Score (CRPS) \cite{gneiting2007strictly}. It is a widely used metric to evaluate probabilistic forecasts that generalizes the MAE to a probabilistic setting. CRPS measures the difference in the cumulative distribution function (CDF) between the forecast and the true observation. It can be derived analytically for parametric distributions or estimated using samples if the CDF is unknown (e.g., originating from VI or MCMC). Given a forecast CDF $\mathbf{F}$ and an empirical CDF of scalar observation $y$:
\begin{equation}
    \mathit{CRPS}(\mathbf{F}, y) = \int_{\mathbb{R}} \left( \mathbf{F}(\hat{y}) - \mathbbm{1}(\hat{y} - y) \right)^2 d\hat{y}
    \label{eq:crps}
\end{equation}

\subsection{Bayesian Neural Networks (BNNs)}
\label{sec:bnn}
In BNNs, we use Bayesian methods for inferring a posterior distribution over the weights $p(\mathbf{w})$ rather than being constrained to weights of fixed values \cite{mackay1992practical}. These weight distributions are parameterized by trainable variables $\boldmath{\theta}$. For example, the trainable variable  can represent the mean and variance of a Gaussian distribution $\boldmath{\theta} = (\boldmath{\mu}, \boldmath{\sigma}^2)$, from which the weights can be sampled $\mathbf{w} \sim \mathcal{N}(\boldmath{\mu}, \boldmath{\sigma}^2)$. 

The objective of training is to calculate the posterior, but obtaining explicit posterior densities through Bayesian inference is intractable.
Additionally, using Markov chain Monte Carlo (MCMC) to estimate the posterior can be computationally prohibitive. Instead, we can leverage new techniques in variational inference \cite{hoffman2013stochastic, kingma2013auto} that make BNNs computationally feasible by using a variational approximation $q(\mathbf{w}|\boldmath{\theta})$ to the posterior. Thereby, the goal of learning is to find the variational parameters that minimize the Kullback–Leibler (KL) divergence between the variational approximation $q(\mathbf{w}|\boldmath{\theta})$ and the true posterior distribution given training data $p(\mathbf{w}|\mathcal{D})$. This can be achieved by minimizing the negative variational lower bound of the marginal likelihood \cite{graves2011practical}:
\begin{equation}
Loss (\theta)  = KL [q(\mathbf{w}|\theta) ||p(\mathbf{w})]-\mathop{{}\mathbb{E}}_{q(\mathbf{w}|\theta) }[\log p(\mathcal{D}|\mathbf{w}] 
\label{eq:variational_lower_bound}
\end{equation}
Assuming Gaussian prior and posterior, we can compute the $KL$ divergence in a closed-form:
\begin{equation}
    KL [q(\mathbf{w}|\theta) ||p(\mathbf{w})] = \log\frac{\pmb\sigma_p} {\pmb\sigma_q} + \frac{\pmb\sigma_q^2 + (\pmb\mu_q - \pmb\mu_p)^2}{2\pmb\sigma_p^2} - 0.5
\end{equation}
Additionally, the distribution over activations will also be Gaussian. Thus, we can take advantage of the local reparameterization trick \cite{kingma2015variational}, in which we sample from the distribution over activations rather than sampling the weights individually. Consequently, we reduce the variance of stochastic gradients, resulting in faster and more stable training. However, to allow for more flexibility and adaptations to a wide range of situations, we can use non-Gaussian distributions over the weights and MC gradients, as proposed by \cite{blundell2015weight}: 
\begin{equation}
Loss (\theta)  \approx \sum_{i=1}^{M} \log q(\mathbf{w}^i|\theta) - \log P(\mathbf{w}^i) - \log P(\mathcal{D}|\mathbf{w}^i)
\label{eq:estimate_kl_divergence}
\end{equation}
where $- \log P(\mathcal{D}|\mathbf{w}^i)$ is the $NLL$ for time-series forecasting (Equation \ref{eq:nll}) or the cross-entropy for the threshold exceedance prediction (Equation \ref{eq:ce}).

Our implementation uses Laplace distributions as priors and Gaussian as approximate posteriors in time-series forecasting, resulting in better empirical performance. For threshold exceedance prediction, using Gaussian for both the priors and posteriors and using the local reparameterization trick results in better performance. During inference, we sample the weight distributions and perform a forward pass to obtain a prediction. We use ($M=1000$) samples for every data point to estimate the model's uncertainty. This corresponds to sampling from infinite ensembles of neural networks. Therefore, combining the outputs from different samples gives information on the model's uncertainty. 

\begin{figure}[!htp]
    \centering
    \includegraphics[width=\linewidth]{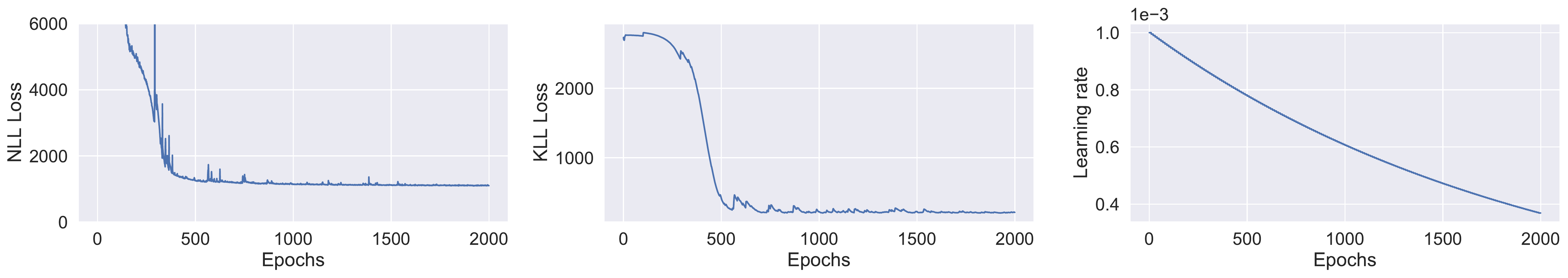}
    \caption{Learning curve of training a BNN model to forecast PM-values. \textbf{Left:} negative log-likelihood loss; \textbf{Center:} KL loss estimated using MC sampling; \textbf{Right:} learning rate of exponential decay.}
    \label{fig:bnn_reg_training}
\end{figure}

\begin{figure}[!htp]
    \centering
    \includegraphics[width=\linewidth]{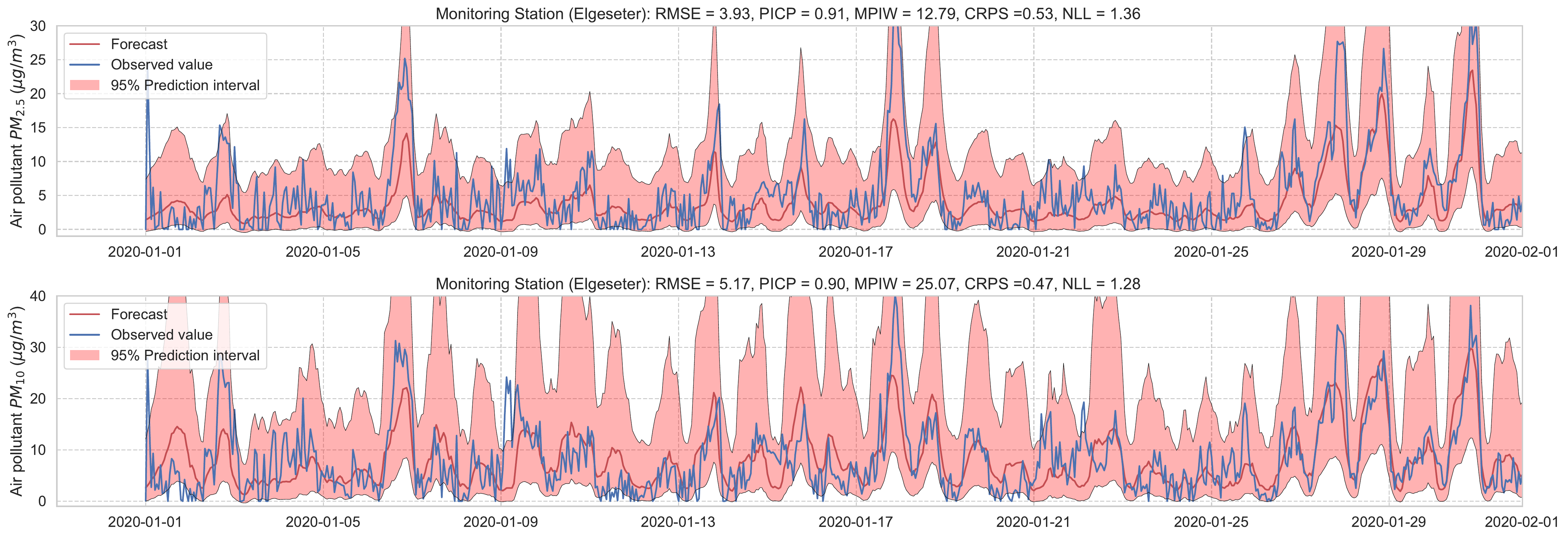}
    \caption{Probabilistic forecasting of multivariate time-series of air quality using a BNN model in one representative monitoring station.}
    \label{fig:bnn_reg}
\end{figure}

\begin{figure}[!htp]
    \centering
    \includegraphics[width=\linewidth]{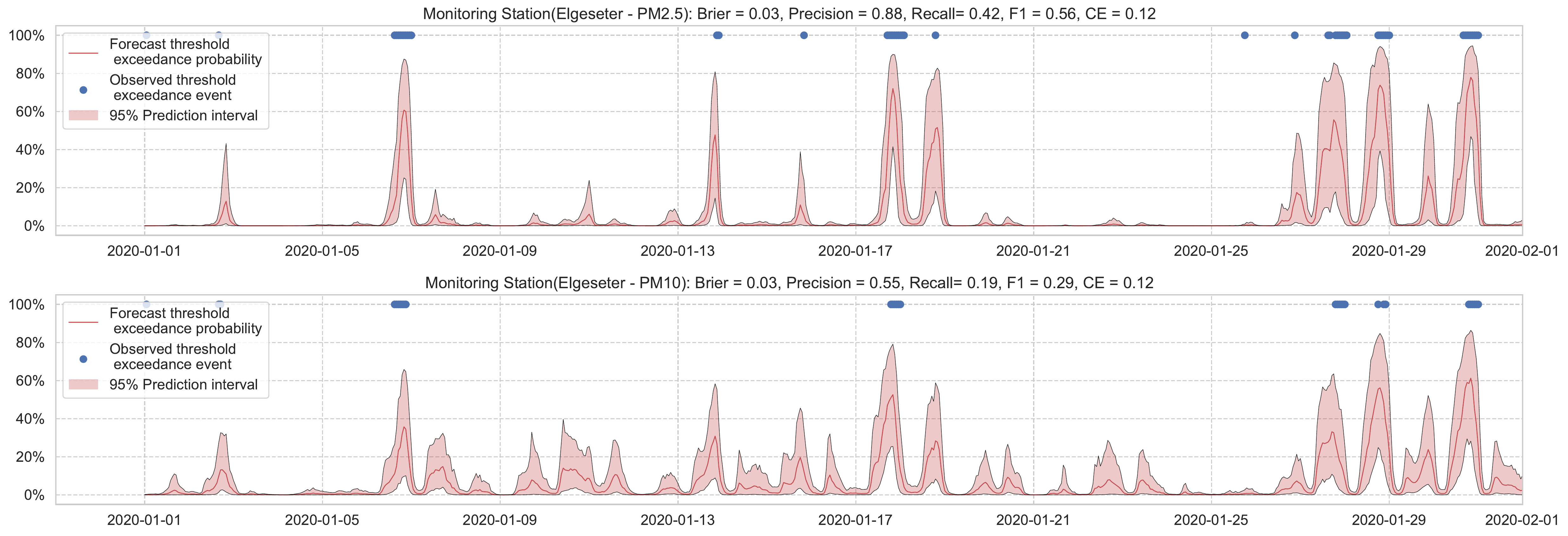}
    \caption{Predicting threshold exceedance probability of air pollutant level using a BNN model.}
    \label{fig:bnn_class}
\end{figure}

We train the BNN model using training data of one year (2019). Figure \ref{fig:bnn_reg_training} shows the learning curves for the PM-value regression task. Then we evaluate the model using data of one month (January 2020). Figures \ref{fig:bnn_reg} and \ref{fig:bnn_class} show the results of PM-value regression and threshold exceedance classification in one representative monitoring station. Table \ref{tab:bnn} shows a summary of performance results in all monitoring stations. The arrows alongside the metrics indicate which direction is better for that specific metric. Additionally, appendix \ref{sec:appendix_bnn} shows the performance results in all monitoring stations.


\begin{table}[!htp]
\small
\centering
\captionsetup{justification=centering}
\caption{Summary of performance results when forecasting the PM-value and threshold exceedance using a BNNs model.}
 \begin{tabular}{c c  c  c c c c  c  c c c c} 
 \hline
  \multirow{2}{*}{Station}& \multirow{2}{*}{Particulate}& \multicolumn{5}{c}{\text{PM-Value Regression}}  &  \multicolumn{5}{c}{\text{Threshold Exceedance Classification}} \\ 
 & &RMSE$\downarrow$&PICP$\uparrow$&MPIW$\downarrow$&CRPS$\downarrow$&NLL$\downarrow$& Brier$\downarrow$&Precision$\uparrow$&Recall$\uparrow$&F1$\uparrow$&CE$\downarrow$\\  

 \hline 
\multirow{2}{*}{Bakke Kirke}&$PM_{2.5}$ &4.81&0.99&17.62&0.51&1.29 &0.04&1.00&0.44&0.61&0.13\\

                             &$PM_{10}$ &5.86&0.94&26.12&0.50&1.28 &0.03&1.00&0.30&0.47&0.09\\

\hline 
\multirow{2}{*}{E6-Tiller}& $PM_{2.5}$ &3.77&0.92&13.25&0.54&1.39 &0.02&0.00&0.00&0.00&0.08\\

                            &$PM_{10}$ &9.40&0.92&34.18&0.48&1.26 &0.06&0.00&0.00&0.00&0.23\\

\hline 
\multirow{2}{*}{Elgeseter}  &$PM_{2.5}$ &3.93&0.91&12.79&0.53&1.36 &0.03&0.88&0.42&0.56&0.12\\

                            &$PM_{10}$  &5.17&0.90&25.07&0.47&1.28 &0.03&0.55&0.19&0.29&0.12\\

\hline 
\multirow{2}{*}{Torvet}     &$PM_{2.5}$ &4.07&0.90&10.83&0.48&1.30 &0.03&0.75&0.46&0.57&0.13\\

                            & $PM_{10}$ &5.25&0.93&18.47&0.43&1.17 &0.03&0.50&0.23&0.32&0.10\\ 
\hline
\end{tabular}
  \label{tab:bnn}
\end{table}

\subsection{Standard Neural Networks with MC Dropout}

Although BNNs are more flexible in reasoning about the model's uncertainty with Bayesian analysis, they are computationally less efficient and take longer to converge than standard (non-Bayesian) neural networks. Additionally, they require double the number of parameters at deployment compared to neural networks of the same size. A possible solution is to gracefully prune the weights with the lowest signal-to-noise ratio \cite{blundell2015weight}, but this leads to a loss in uncertainty information. Therefore, it would be more convenient to obtain uncertainty directly from standard neural networks. One simple approach we can use is Monte Carlo dropout as an approximate Bayesian method for representing model uncertainty. As shown in \cite{gal2016dropout}, MC dropout can be interpreted as performing variational inference. More specifically, it is mathematically equivalent to an approximation of a probabilistic deep Gaussian process.

\begin{figure}[!htp]
    \centering
    \includegraphics[width=\linewidth]{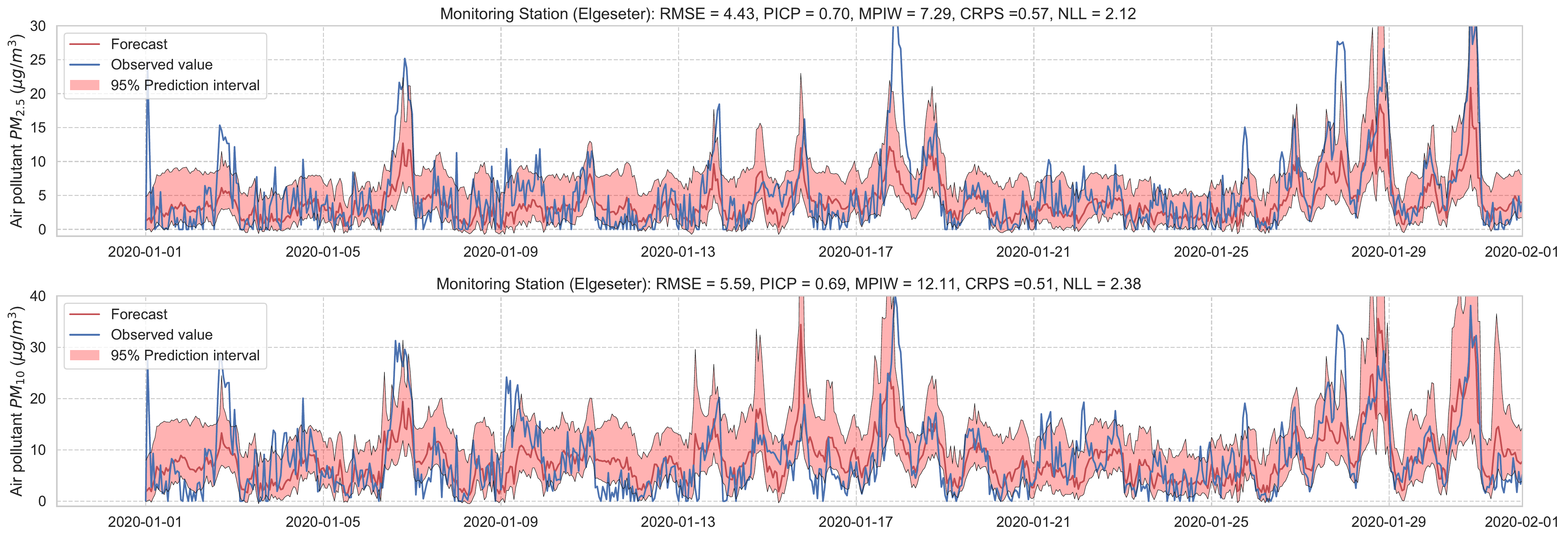}
    \caption{Probabilistic forecasting of multivariate time-series air quality using a standard neural network model with MC dropout.}
    \label{fig:mc_reg}
\end{figure}

\begin{figure}[!htp]
    \centering
    \includegraphics[width=\linewidth]{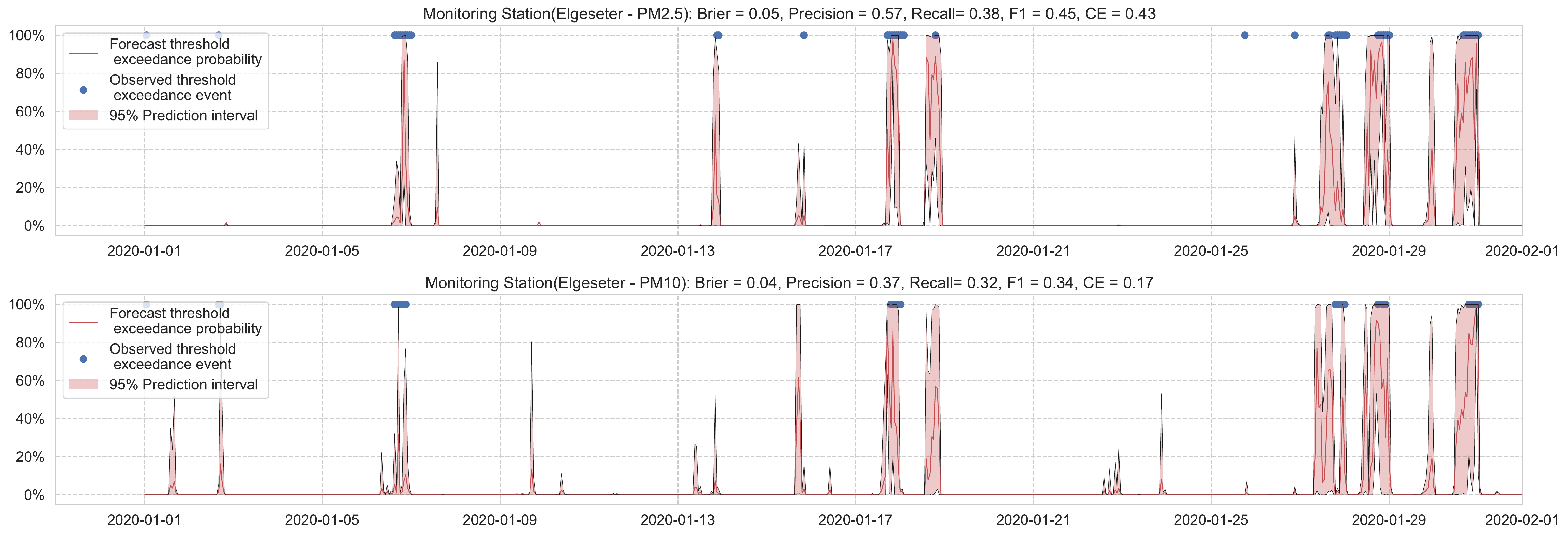}
    \caption{Predicting the threshold exceedance probability of air pollutants level using a standard neural network with MC dropout.}
    \label{fig:mc_class}
\end{figure}

Essentially, we train a standard neural network model with dropout. Then we keep the dropout during inference and run multiple forward passes using the same data input. This corresponds to sampling nodes, which is equivalent to sampling from ensembles of neural networks \cite{srivastava2014dropout}. By measuring the spread in predictions, we estimate the predictive uncertainty. In our implementations, we train a standard neural network model with a 50\% dropout rate. Then we evaluate it with 50\% dropout and ($M=1000$) samples. Figures \ref{fig:mc_reg} and \ref{fig:mc_class} show the results of the PM-value regression and threshold exceedance classification in one representative monitoring station. Table \ref{tab:mc} shows a summary of performance results in all monitoring stations, while Appendix \ref{sec:appendix_mc} contains the corresponding figures.

\begin{table}[H]
\small
\centering
\captionsetup{justification=centering}
\caption{Summary of performance results when forecasting the PM-value and threshold exceedance using a standard neural network with MC dropout.}
 \begin{tabular}{c c  c  c c c c  c  c c c c} 
 \hline
  \multirow{2}{*}{Station}& \multirow{2}{*}{Particulate}& \multicolumn{5}{c}{\text{PM-Value Regression}}  &  \multicolumn{5}{c}{\text{Threshold Exceedance Classification}} \\ 
 & &RMSE$\downarrow$&PICP$\uparrow$&MPIW$\downarrow$&CRPS$\downarrow$&NLL$\downarrow$& Brier$\downarrow$&Precision$\uparrow$&Recall$\uparrow$&F1$\uparrow$&CE$\downarrow$\\ 

 \hline 
\multirow{2}{*}{Bakke kirke}&$PM_{2.5}$ &5.34&0.69&9.40&0.60&2.30 &0.04&0.65&0.51&0.57&0.31\\

                             &$PM_{10}$ &6.42&0.66&12.45&0.59&3.35 &0.03&0.67&0.48&0.56&0.10\\

\hline 
\multirow{2}{*}{E6-Tiller}& $PM_{2.5}$ &3.75&0.72&7.26&0.60&2.24 &0.01&0.00&0.00&0.00&0.24\\

                            &$PM_{10}$ &9.49&0.71&16.62&0.51&2.30 &0.07&0.18&0.04&0.06&0.57\\

\hline 
\multirow{2}{*}{Elgeseter}  &$PM_{2.5}$ &4.43&0.70&7.29&0.57&2.12 &0.05&0.57&0.38&0.45&0.43\\

                            &$PM_{10}$  &5.59&0.69&12.11&0.51&2.38 &0.04&0.37&0.32&0.34&0.17\\

\hline 
\multirow{2}{*}{Torvet}     &$PM_{2.5}$ &4.60&0.55&5.26&0.57&2.91 &0.04&0.68&0.44&0.53&0.33\\

                            & $PM_{10}$ &5.63&0.62&8.94&0.51&2.51 &0.03&0.56&0.35&0.43&0.14\\ 
\hline
\end{tabular}
  \label{tab:mc}
\end{table}

\subsection{Deep Ensembles}

An established method to improve performance is to train an ensemble of neural networks with different configurations \cite{dietterich2000ensemble}. Additionally, many model types can be interpreted by using ensemble methods. For example, sampling weights of BNNs is equivalent to sampling from an infinite ensemble of networks, while sampling the nodes in MC dropout compares to sampling from a finite ensemble \cite{srivastava2014dropout}. Remarkably, we can also use (deep) ensembles to estimate the predictive uncertainty in neural network models \cite{lakshminarayanan2017simple}.

\begin{figure}[!htp]
    \centering
    \includegraphics[width=\linewidth]{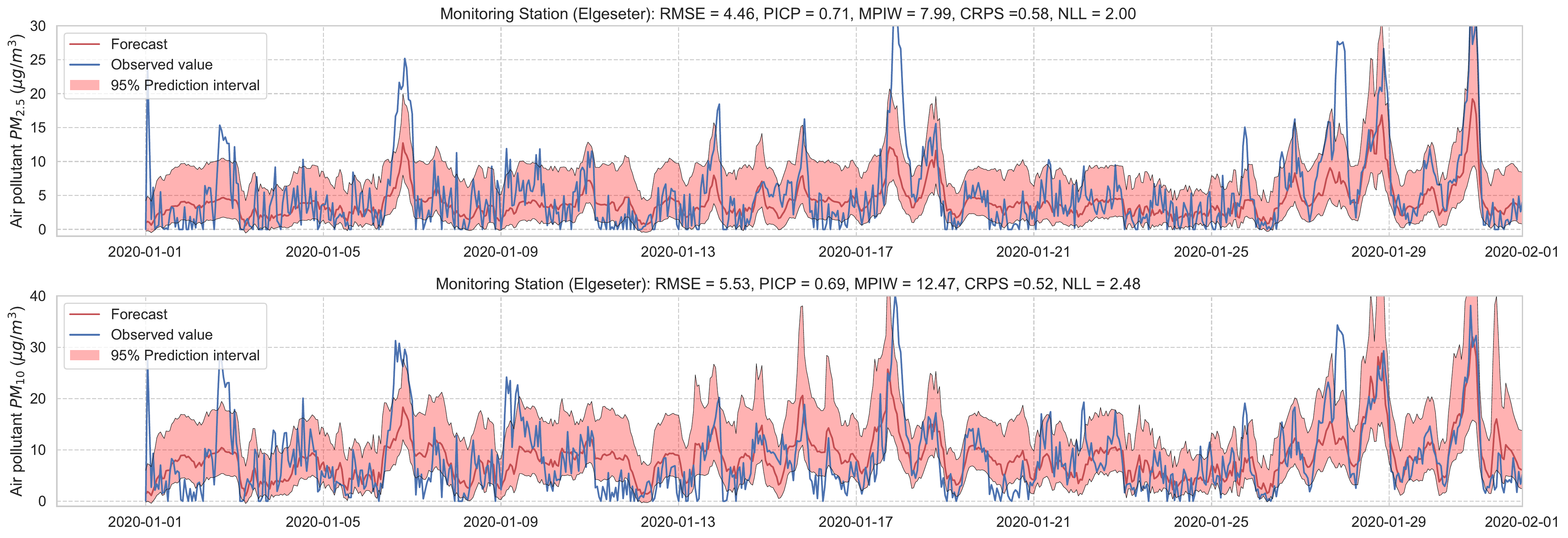}
    \caption{Probabilistic forecasting of multivariate time-series air quality using a deep ensemble.}
    \label{fig:ensemble_reg}
\end{figure}

\begin{figure}[!htp]
    \centering
    \includegraphics[width=\linewidth]{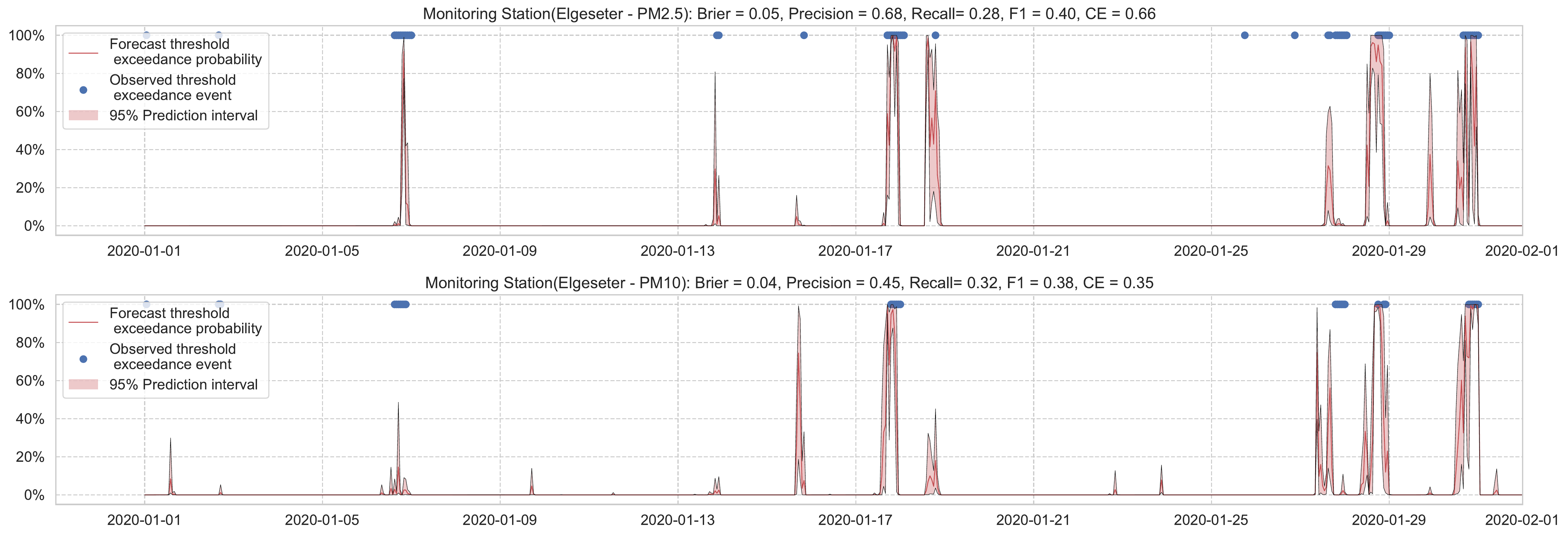}
    \caption{Predicting threshold exceedance probability of air pollutants level using a deep ensemble.}
    \label{fig:ensemble_class}
\end{figure}

Essentially, we train multiple neural networks with different parameter initialization on the same data. The stochastic optimization and random initialization ensure that the trained networks are sufficiently independent. During inference, we run a forward pass on the multiple neural networks using the same data input. We estimate the uncertainty by measuring the dispersion in predictions resulting from the multiple neural networks. In our implementation, we train an ensemble of 10 networks.  Figures \ref{fig:ensemble_reg} and \ref{fig:ensemble_class} show the results of PM-value regression and threshold exceedance classification in one representative monitoring station. Table \ref{tab:ensemble} shows a summary of performance results in all monitoring stations, while Appendix \ref{sec:appendix_ensemble} contains the corresponding figures.

The main drawback of the deep ensemble method is the computational cost and number of parameters required at run time compared to standard neural networks, which is an order of magnitude in our case. One solution is to use knowledge distillation \cite{hinton2015distilling} to compress the knowledge from the ensemble into a single model, but this also leads to a loss in uncertainty information.

\begin{table}[H]
\small
\centering
\captionsetup{justification=centering}
\caption{Summary of performance results when forecasting the PM-value and threshold exceedance using a deep ensemble.}
 \begin{tabular}{c c  c  c c c c  c  c c c c} 
 \hline
  \multirow{2}{*}{Station}& \multirow{2}{*}{Particulate}& \multicolumn{5}{c}{\text{PM-Value Regression}}  &  \multicolumn{5}{c}{\text{Threshold Exceedance Classification}} \\ 
 & &RMSE$\downarrow$&PICP$\uparrow$&MPIW$\downarrow$&CRPS$\downarrow$&NLL$\downarrow$& Brier$\downarrow$&Precision$\uparrow$&Recall$\uparrow$&F1$\uparrow$&CE$\downarrow$\\ 
 \hline

 \hline 
\multirow{2}{*}{Bakke kirke}&$PM_{2.5}$ &5.29&0.77&11.65&0.57&1.67  &0.05&0.69&0.53&0.60&0.55\\

                             &$PM_{10}$ &6.21&0.70&14.00&0.57&2.46 &0.03&0.60&0.36&0.45&0.26\\

\hline 
\multirow{2}{*}{E6-Tiller}& $PM_{2.5}$ &3.78&0.77&8.46&0.58&1.84 &0.01&0.00&0.00&0.00&0.34\\

                            &$PM_{10}$ &9.44&0.72&16.07&0.50&2.14 &0.07&0.31&0.08&0.12&1.16\\

\hline 
\multirow{2}{*}{Elgeseter}  &$PM_{2.5}$ &4.46&0.71&7.99&0.58&2.00 &0.05&0.68&0.28&0.40&0.66\\

                            &$PM_{10}$  &5.53&0.69&12.47&0.52&2.48 &0.04&0.45&0.32&0.38&0.35\\

\hline 
\multirow{2}{*}{Torvet}     &$PM_{2.5}$ &4.45&0.57&5.13&0.56&2.66 &0.04&0.73&0.31&0.43&0.55\\

                            & $PM_{10}$ &5.39&0.64&8.68&0.49&2.19 &0.03&0.62&0.19&0.29&0.30\\ 
\hline
\end{tabular}
  \label{tab:ensemble}
\end{table}

\subsection{Recurrent Neural Network with MC Dropout}

While standard neural networks are powerful models at representational learning, they do not exploit the inherent temporal correlation in air quality data since they act only on static, fixed contextual windows. To address this shortcoming, we can use recurrent neural networks (RNNs), which have cyclic connections from previous time steps, to learn the temporal dynamics of sequential data. Specifically, a hidden state from the last time step is stored and used in addition to the input to generate the current state and output. One class of RNNs is the long short-term memory (LSTM), which is used extensively in sequence modeling tasks, such as modeling language \cite{jozefowicz2016exploring}, forecasting weather \cite{chen2019hybrid} and traffic \cite{li2018diffusion}, recognizing human activity \cite{murad2017deep}, and recently forecasting COVID-19 transmission \cite{chimmula2020time}. An LSTM has gated memory cells to control how much information to forget from previous states and how much information to use to update current states \cite{sak2014long}. 

\begin{figure}[!htp]
    \centering
    \includegraphics[width=\linewidth]{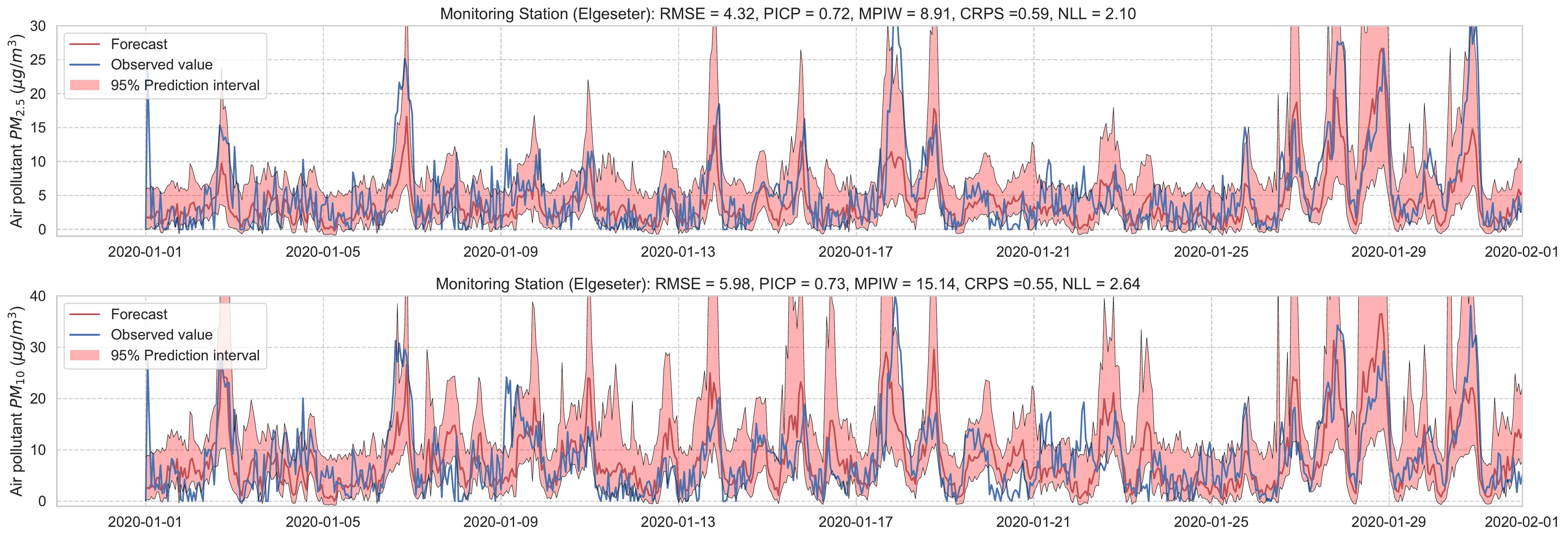}
    \caption{Probabilistic forecasting of multivariate time-series air quality using an LSTM model with MC dropout.}
    \label{fig:lstm_reg}
\end{figure}

\begin{figure}[!htp]
    \includegraphics[width=\linewidth]{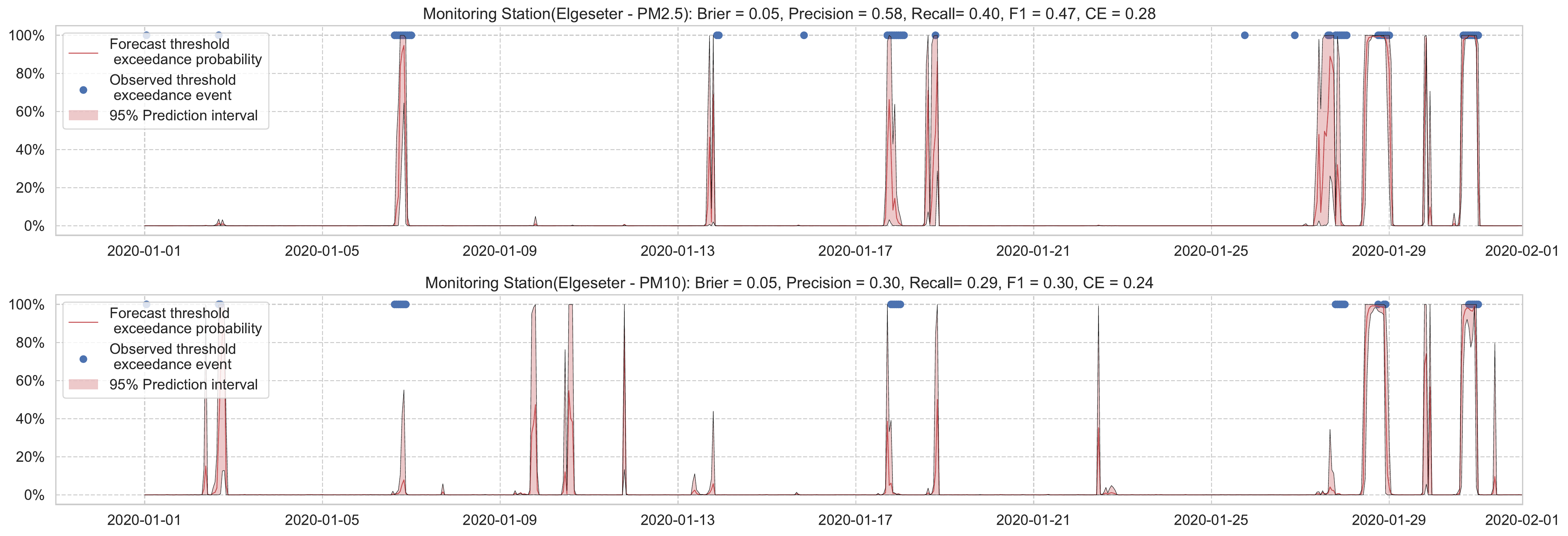}
    \caption{Predicting threshold exceedance probability of air pollutants level using an LSTM model with MC dropout.}
    \label{fig:lstm_class}
\end{figure}

A simple approach to capture uncertainty within an LSTM model is to use Monte Carlo dropout to approximate Bayesian inference \cite{zhu2017deep}. To implement dropout, we apply a mask that randomly drops some network units with their inputs, output and recurrent connections \cite{gal2016theoretically}. Our implementation trains a model of two LSTM layers with a 50\% dropout rate and evaluates it with 50\% dropout and ($M=1000$) samples. Figures \ref{fig:lstm_reg} and \ref{fig:lstm_class} show the results of the PM-value regression and threshold exceedance classification in one representative monitoring station. Table \ref{tab:lstm} shows a summary of performance results in all monitoring stations, while Appendix \ref{sec:appendix_lstm} contains the corresponding figures.

\begin{table}[!htp]
\small
\centering
\captionsetup{justification=centering}
\caption{Summary of performance results when forecasting PM-value or threshold exceedance using an LSTM model with MC dropout.}
 \begin{tabular}{c c  c  c c c c  c  c c c c}  

 \hline
  \multirow{2}{*}{Station}& \multirow{2}{*}{Particulate}& \multicolumn{5}{c}{\text{PM-Value Regression}}  &  \multicolumn{5}{c}{\text{Threshold Exceedance Classification}} \\ 
 & &RMSE$\downarrow$&PICP$\uparrow$&MPIW$\downarrow$&CRPS$\downarrow$&NLL$\downarrow$& Brier$\downarrow$&Precision$\uparrow$&Recall$\uparrow$&F1$\uparrow$&CE$\downarrow$\\ 

 \hline 
\multirow{2}{*}{Bakke kirke}&$PM_{2.5}$ &5.01&0.88&14.01&0.53&1.47   &0.05&0.66&0.53&0.58&0.25\\

                             &$PM_{10}$ &6.25&0.82&19.28&0.54&1.78 &0.03&0.59&0.48&0.53&0.14\\

\hline 
\multirow{2}{*}{E6-Tiller}& $PM_{2.5}$ &3.90&0.72&7.45&0.62&2.31 &0.02&0.00&0.00&0.00&0.11\\

                            &$PM_{10}$ &9.68&0.74&18.96&0.53&2.03 &0.08&0.24&0.12&0.16&0.43\\

\hline 
\multirow{2}{*}{Elgeseter}  &$PM_{2.5}$ &4.32&0.72&8.91&0.59&2.10 &0.05&0.58&0.40&0.47&0.28\\

                            &$PM_{10}$ &5.98&0.73&15.14&0.55&2.64 &0.05&0.30&0.29&0.30&0.24\\

\hline 
\multirow{2}{*}{Torvet}     &$PM_{2.5}$ &4.19&0.56&6.88&0.58&4.79  &0.05&0.58&0.42&0.49&0.30\\

                            & $PM_{10}$ &5.81&0.61&11.33&0.54&4.03 &0.03&0.43&0.35&0.38&0.1\\ 
\hline
\end{tabular}
  \label{tab:lstm}
\end{table}

\subsection{Graph Neural Networks with MC Dropout}

By using RNNs, we can capture the temporal (i.e., intra-series) correlations in the time-series of air quality data. However,  we need also to exploit the inherent structural (i.e., inter-series) correlations between multiple sensing stations. We can use Graph Neural Networks (GNNs) to address this, which operate on graph-structured data. In essence, GNNs update the features of a graph node by aggregating the features of its adjacent nodes. For example, a node can be a single monitoring station in our case. In the end, GNNs apply a shared layer on each node to obtain a prediction for each node.

In our setting, we assume each sensing station to be a node in a weighted and directed graph, represented by a learnable adjacency matrix. We use a slight variation of the GNNs suggested by Cao et al. \cite{cao2020spectral} to forecast multivariate time-series of air quality. The main idea is to learn a correlation graph directly from data (i.e., learn a graph of sensor nodes without a pre-defined typology) and then learn the structural and temporal correlation in the frequency domain.
To capture the temporal correlations, we use a layer of 1D convolution, and three sub-layers of Gated Linear Units (GLUs) \cite{dauphin2017language}, while we use Graph Convolutional Networks (GCNs) \cite{kipf2016semi} to capture the structural correlations. In the end, we use two shared layers of fully connected neural networks to predict each node's output.

\begin{figure}[!htp]
    \includegraphics[width=\linewidth]{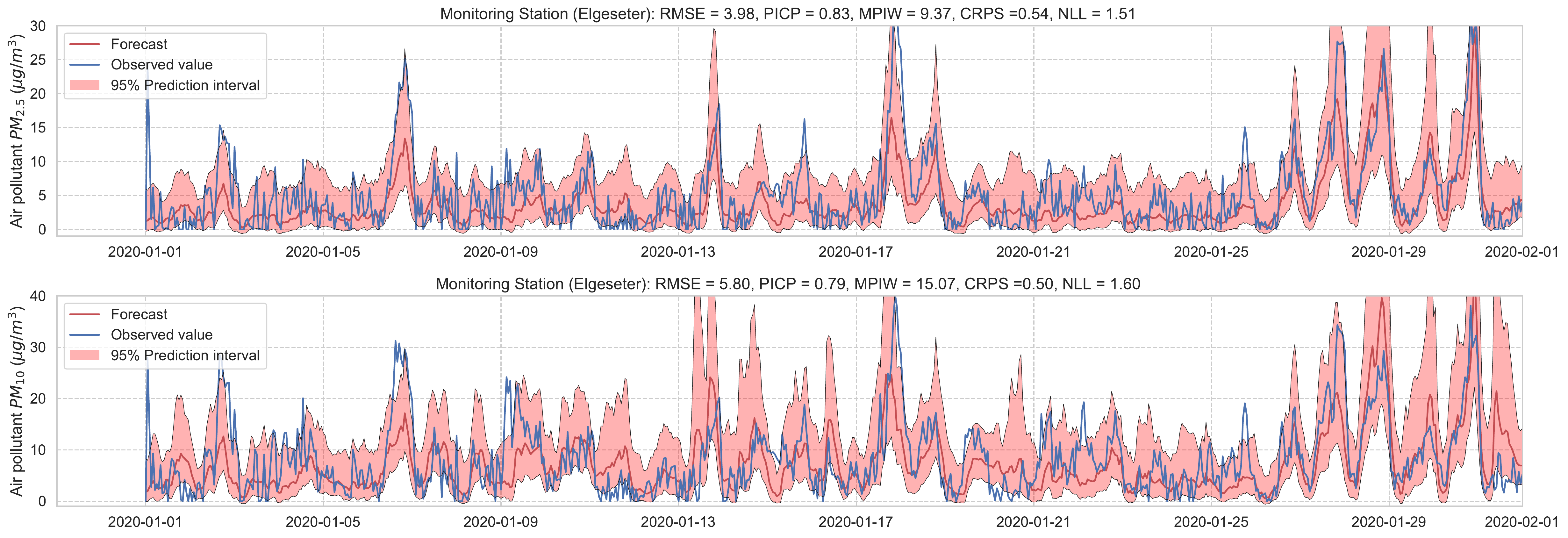}
    \caption{Probabilistic forecasting of multivariate time-series air quality using a GNN model with MC dropout.}
    \label{fig:gnn_reg}
\end{figure}

\begin{figure}[!htp]
    \includegraphics[width=\linewidth]{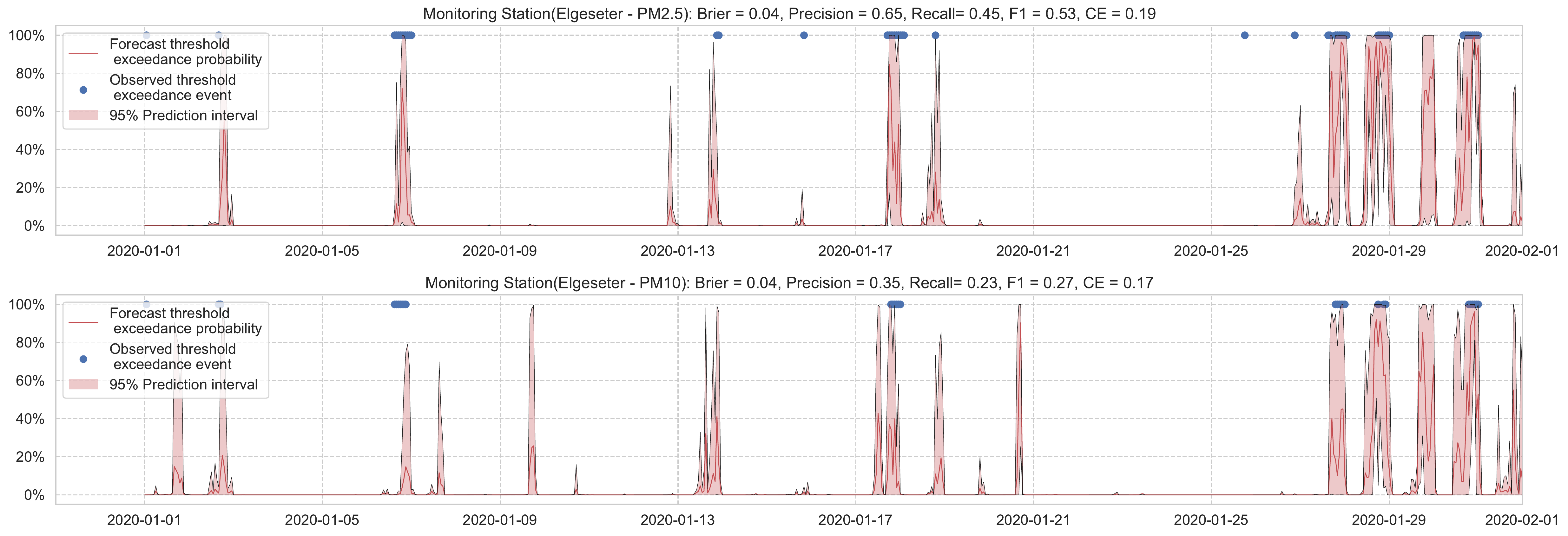}
    \caption{Predicting threshold exceedance probability of air pollutants level using a GNN model with MC dropout.}
    \label{fig:gnn_class}
\end{figure}

We use the MC dropout applied to the GLUs, GCNs, and to the fully connected layers to estimate model's uncertainty \cite{hasanzadeh2020bayesian}. We train the model with a 50\% dropout rate and evaluate it with 50\% dropout and ($M=1000$) samples. Figures \ref{fig:gnn_reg} and \ref{fig:gnn_class} show
 the results of the PM-value regression and threshold exceedance classification in one representative monitoring station. 
 Table \ref{tab:gnn} shows a summary of performance results in all monitoring stations, while Appendix \ref{sec:appendix_gnn} contains more-detailed figures.

\begin{table}[!htp]
\small
\centering
\captionsetup{justification=centering}
\caption{Summary of performance results when forecasting the PM-value or threshold exceedance using a GNN model with MC dropout.}
 \begin{tabular}{c c  c  c c c c  c  c c c c}  

 \hline
  \multirow{2}{*}{Station}& \multirow{2}{*}{Particulate}& \multicolumn{5}{c}{\text{PM-Value Regression}}  &  \multicolumn{5}{c}{\text{Threshold Exceedance Classification}} \\ 
 & &RMSE$\downarrow$&PICP$\uparrow$&MPIW$\downarrow$&CRPS$\downarrow$&NLL$\downarrow$& Brier$\downarrow$&Precision$\uparrow$&Recall$\uparrow$&F1$\uparrow$&CE$\downarrow$\\  
 \hline

 \hline 
\multirow{2}{*}{Bakke kirke}&$PM_{2.5}$ &4.70&0.88&12.33&0.52&1.41 &0.05&0.61&0.53&0.56&0.21\\

                             &$PM_{10}$ &6.26&0.79&16.10&0.54&1.83 &0.03&0.43&0.36&0.39&0.11\\

\hline 
\multirow{2}{*}{E6-Tiller}& $PM_{2.5}$ &3.80&0.83&9.14&0.57&1.60 &0.02&0.00&0.00&0.00&0.11\\

                            &$PM_{10}$ &9.46&0.80&19.89&0.48&1.59 &0.07&0.19&0.06&0.09&0.35\\

\hline 
\multirow{2}{*}{Elgeseter}  &$PM_{2.5}$ &3.98&0.83&9.37&0.54&1.51 &0.04&0.65&0.45&0.53&0.19\\

                            &$PM_{10}$ &5.80&0.79&15.07&0.50&1.60 &0.04&0.35&0.23&0.27&0.17\\

\hline 
\multirow{2}{*}{Torvet}     &$PM_{2.5}$ &4.27&0.68&6.19&0.50&2.04 &0.05&0.55&0.46&0.50&0.22\\

                            & $PM_{10}$ &5.55&0.70&10.39&0.47&1.83 &0.03&0.36&0.35&0.35&0.11\\ 
\hline
\end{tabular}
  \label{tab:gnn}
\end{table}

\subsection{Stochastic Weight Averaging--Gaussian (SWAG)}

An alternative approach to estimate uncertainty in a neural network is to approximate (Gaussian) posterior distributions over the weights using the geometric information in the trajectory of a stochastic optimizer. This approach is named Stochastic Weight Averaging--Gaussian (SWAG) \cite{maddox2019simple}. Notably, SWAG does not optimize the approximate distributions directly, such as in BNNs. Instead, it estimates the mean by calculating a running average of the weights traversed by the optimizer with a modified learning rate schedule (stochastic weight averaging \cite{izmailov2018averaging}). In addition, SWAG estimates the standard deviation by a diagonal covariance plus of a low-rank deviation matrix using information from a running average of the second moment of the traversed weights.

\begin{figure}[!htp]
    \centering
    \includegraphics[width=\linewidth]{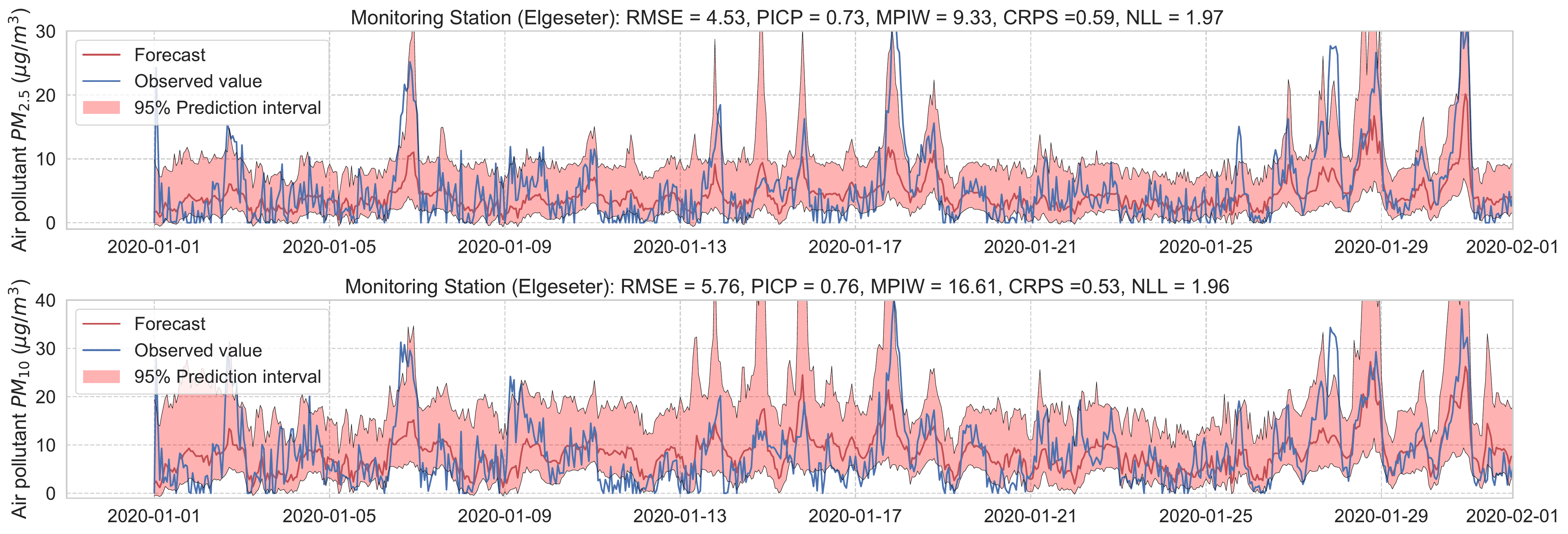}
    \caption{Probabilistic PM-value regression using a SWAG model.}
    \label{fig:swag_reg}
\end{figure}

\begin{figure}[!htp]
    \centering
    \includegraphics[width=\linewidth]{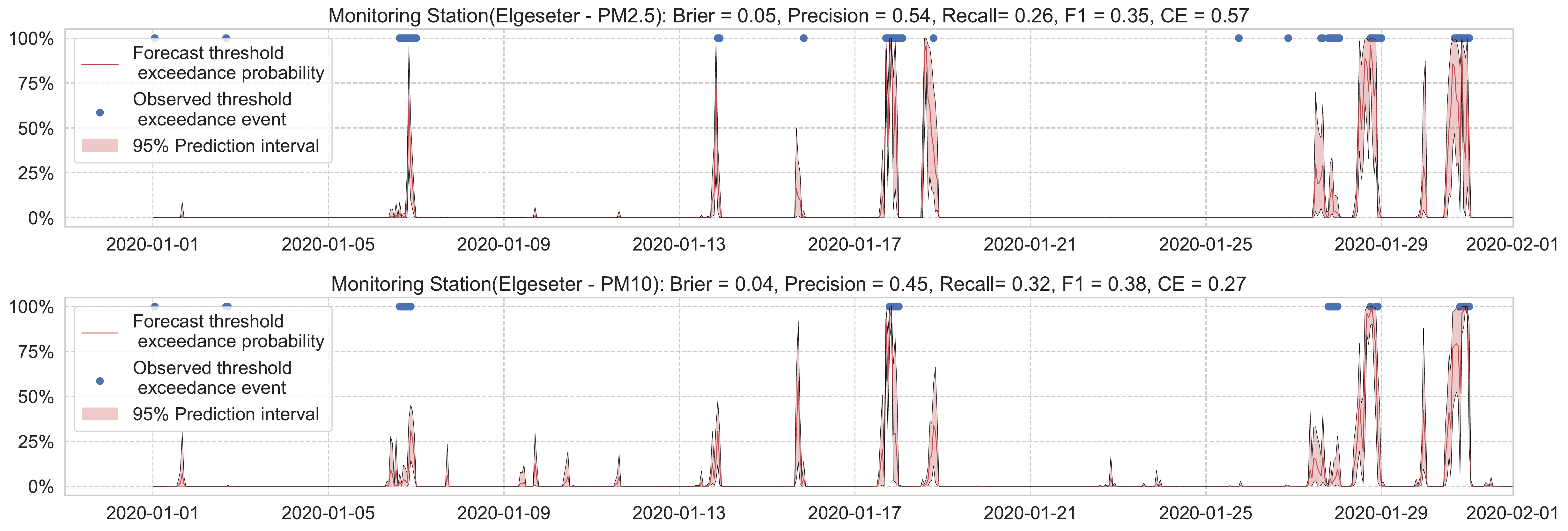}
    \caption{Probabilistic threshold exceedance classification using a SWAG model.}
    \label{fig:swag_class}
\end{figure}

During inference, we run multiple forward passes using the same data input while drawing samples of the weights from the approximate posterior. By measuring the spread in predictions, we estimate the predictive uncertainty. In our implementations, we train a simple feed-forward neural network model with SWAG and evaluate it with ($M=1000$) samples. Figures \ref{fig:swag_reg} and \ref{fig:swag_class} show the results of the PM-value regression and threshold exceedance classification in one representative monitoring station. Table \ref{tab:swag} shows a summary of performance results in all monitoring stations, and Appendix \ref{sec:appendix_swag} contains the corresponding figures.
 
\begin{table}[!htp]
\small
\centering
\captionsetup{justification=centering}
\caption{Summary of performance results when using a SWAG model with adversarial training.}

 \begin{tabular}{c c  c  c c c c  c  c c c c} 
 \hline
  \multirow{2}{*}{Station}& \multirow{2}{*}{Particulate}& \multicolumn{5}{c}{\text{PM-Value Regression}}  &  \multicolumn{5}{c}{\text{Threshold Exceedance Classification}} \\ 
 & &RMSE$\downarrow$&PICP$\uparrow$&MPIW$\downarrow$&CRPS$\downarrow$&NLL$\downarrow$& Brier$\downarrow$&Precision$\uparrow$&Recall$\uparrow$&F1$\uparrow$&CE$\downarrow$\\  
 \hline

 \hline 
\multirow{2}{*}{Bakke kirke}&$PM_{2.5}$ &5.51&0.79&13.13&0.58&1.64 &0.04&0.66&0.64&0.65&0.20\\

                             &$PM_{10}$ &6.66&0.78&17.95&0.57&2.03 &0.04&0.49&0.61&0.54&0.12\\

\hline 
\multirow{2}{*}{E6-Tiller}& $PM_{2.5}$ &3.76&0.79&9.25&0.59&1.82 &0.01&0.00&0.00&0.00&0.10\\

                            &$PM_{10}$ &9.35&0.82&21.28&0.49&1.73 &0.08&0.19&0.08&0.11&0.49 \\

\hline 
\multirow{2}{*}{Elgeseter}  &$PM_{2.5}$ &4.53&0.73&9.33&0.59&1.97 &0.04&0.60&0.45&0.52&0.21\\

                            &$PM_{10}$ &5.76&0.76&16.61&0.53&1.96 &0.04&0.37&0.45&0.41&0.18 \\

\hline 
\multirow{2}{*}{Torvet}     &$PM_{2.5}$ &4.58&0.79&10.33&0.54&1.63 &0.04&0.67&0.50&0.57&0.20 \\

                            & $PM_{10}$&5.62&0.71&12.48&0.50&1.76 &0.03&0.50&0.42&0.46&0.13\\ 
\hline
\end{tabular}
  \label{tab:swag}
\end{table}

\subsection{Improving Uncertainty Estimation with Adversarial Training}

Generally, it is desirable to have a model with a smooth conditional output distribution with respect to its input because most measured phenomena are inherently smooth. This idea of distributional smoothing has been used as a regularization technique by encouraging a model to be less overconfident, for example, using label smoothing \cite{szegedy2016rethinking, muller2019does} or virtual adversarial training \cite{miyato2018virtual}. 

For uncertainty estimation, distributional smoothing can improve the quality of the predictive uncertainty depending on the direction of smoothing. For example, smoothing along a random direction can be less effective while being computationally expensive in all directions. Lakshminarayanan et al. \cite{lakshminarayanan2017simple} propose using adversarial training with the fast gradient sign method \cite{goodfellow2014explaining} to smooth the predictive distribution along the direction where the loss is high. Qin et al. \cite{qin2020improving} investigate the relationship between adversarial robustness and predictive uncertainty. They show that inputs that are sensitive to adversarial perturbations are more likely to have unreliable predictive uncertainty. Based on this insight, they propose a new training approach that smooths training labels based on their input adversarial robustness.

\begin{figure}[!htp]
\centering
    \includegraphics[width=0.93\linewidth]{figures/mc_reg.pdf}
    \includegraphics[width=0.93\linewidth]{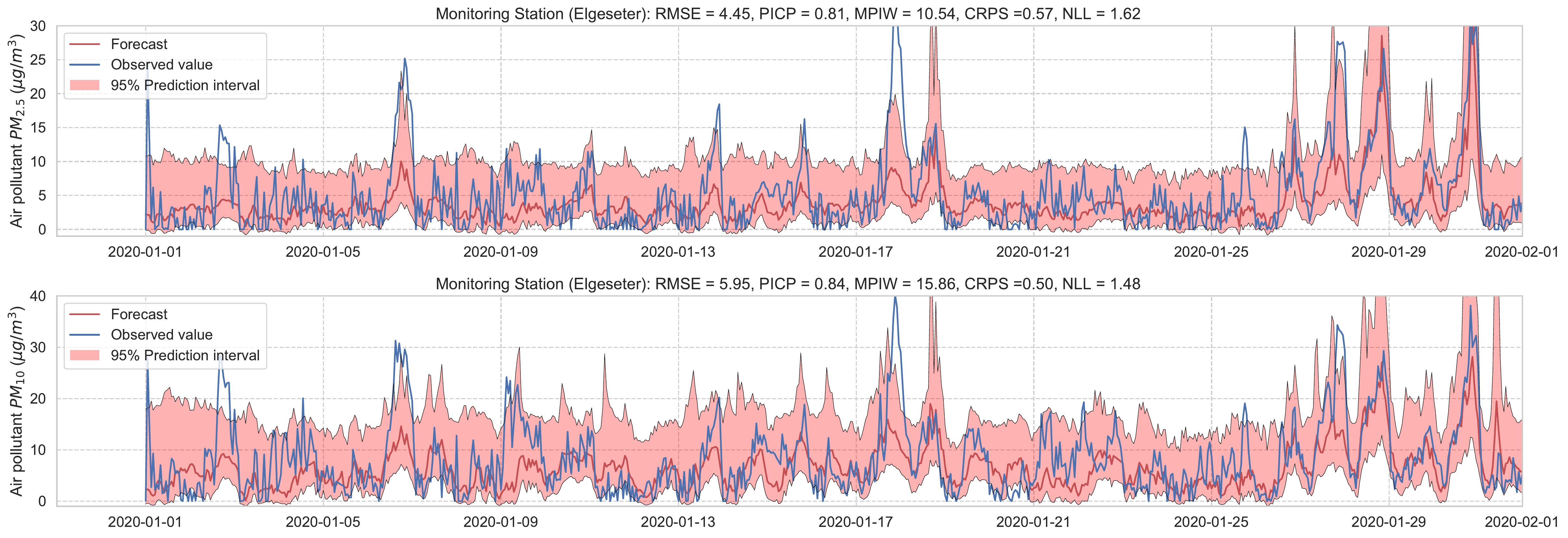}
    \caption{Comparison of uncertainty estimation in PM-value regression when training \textbf{(top)} without adversarial training versus \textbf{(bottom)} with adversarial training. Using Adversarial training leads to smoother predictive distribution; thus, lower NLL (less overconfident predictions).}
    \label{fig:mc_reg_adv}
\end{figure}

\begin{figure}[!htp]
 \centering
    \includegraphics[width=0.93\linewidth]{figures/mc_class.pdf}
    \includegraphics[width=0.93\linewidth]{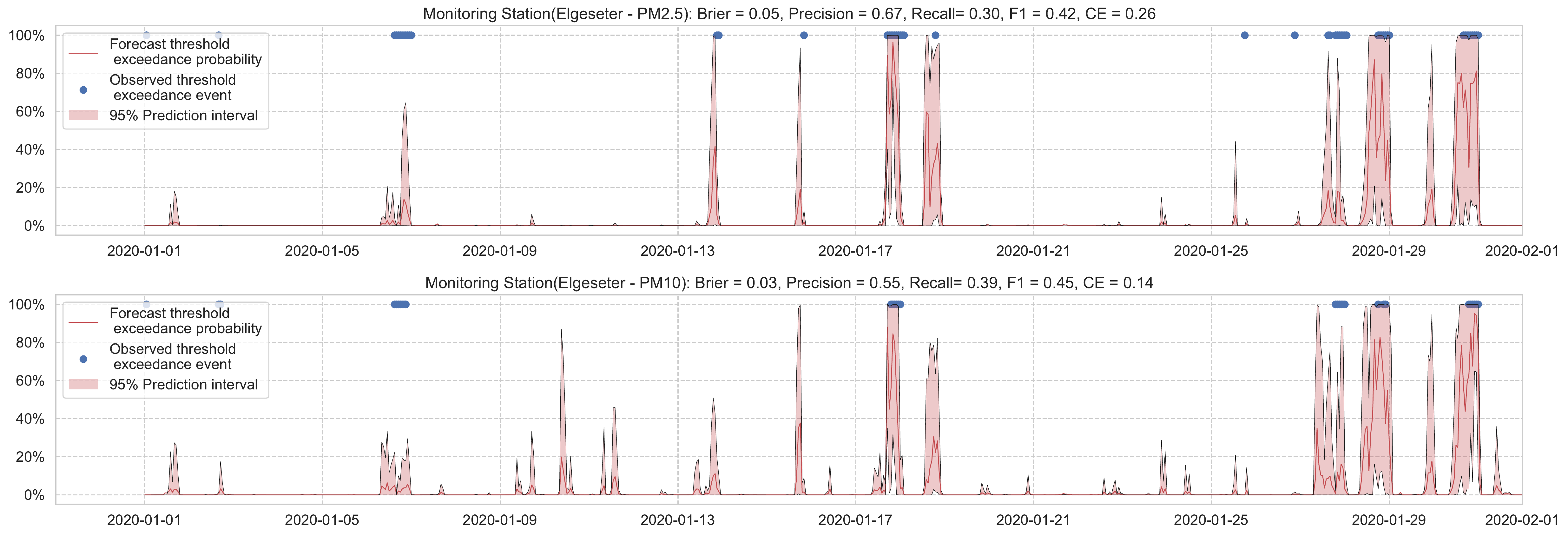}
    \caption{Comparison of uncertainty estimation in threshold exceedance classification when training \textbf{(top)} without adversarial training versus \textbf{(bottom)} with adversarial training. Using Adversarial training leads to smoother predictive distribution; thus, lower CE (less overconfident predictions).}
    \label{fig:mc_class_adv}
\end{figure}

In this paper, we instead propose using the ``free'' adversarial method \cite{shafahi2019adversarial}, which recycles the gradient information from regular training to quickly generate adversarial data. Thus, we locally smooth the prediction distribution along the adversarial direction with virtually no additional cost. Figures \ref{fig:mc_reg_adv} illustrates the improvement in uncertainty estimation when using adversarial training in PM-value regression task, while figure \ref{fig:mc_class_adv} in threshold exceedance classification (using MC dropout as an example).Generally, we observe that adversarial training improves the NLL and CE (i.e., making less overconfident predictions) with negligible effects on other metrics. This, of course, depends on the size of the adversarial perturbation, which can be tuned accordingly. We observe that increasing the perturbation size can have adverse effects on the accuracy metrics, which is expected \cite{zhang2019theoretically}. We also observe that adversarial training led to more improvements in the PM-value regression than in the threshold exceedance classification. This is because threshold exceedance events are rare, and smoothing the predictive distribution does not help catch these events.
\clearpage

\section{Discussion}
\label{sec:discussion}
In this section, we investigate some of the implications of the study. In particular, we perform a comparative analysis to evaluate the selected probabilistic models based on empirical performance, reliability of confidence estimate, and practical applicability. Then we close by investigating the practical impact of uncertainty quantification on decision-making.           

\subsection{Empirical Performance} 
\label{sec:empirical_performance}
In Section \ref{sec:deep_probabilistic_forecast}, we summarized the performance of each model in a tabular format. Here, we compare all the models according to their empirical performance in a single monitoring station. 

\begin{figure}[!htb]
    \centering
    \includegraphics[width=\linewidth]{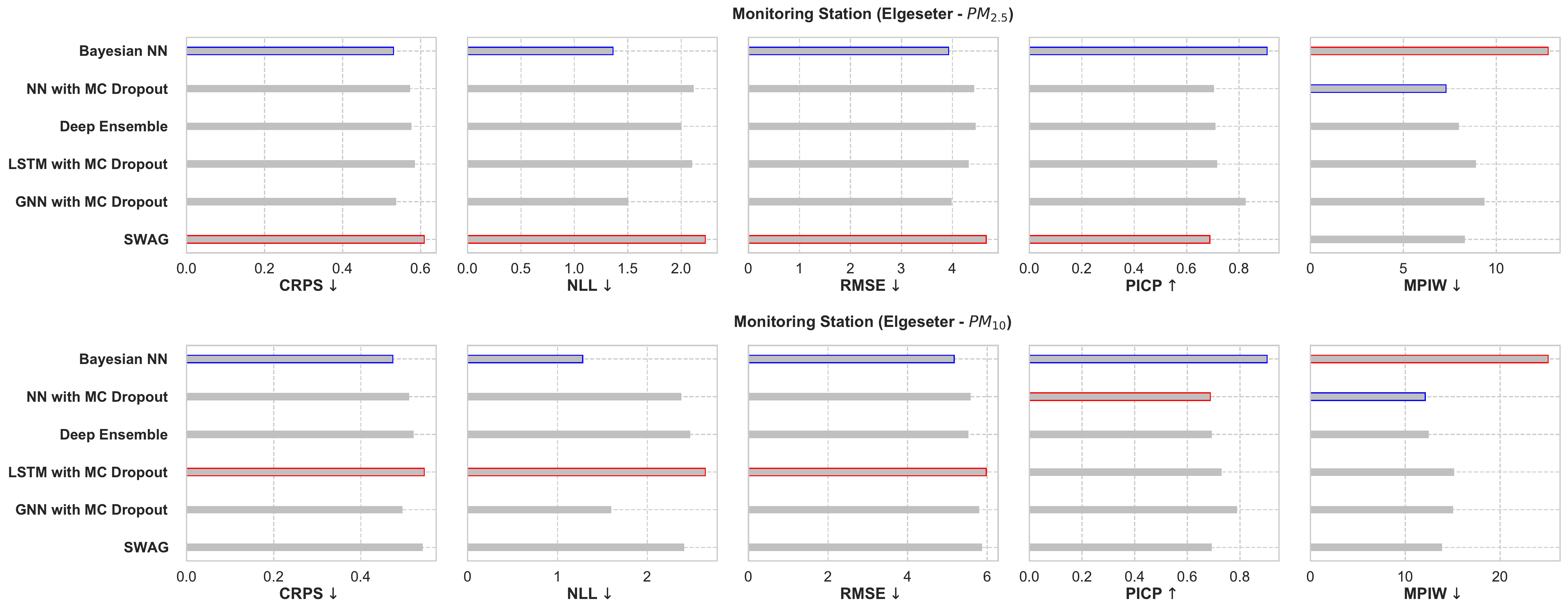}
    \caption{Comparison of empirical performance of the selected probabilistic models in the PM-value regression task.  The comparison is according to five performance metrics (left to right): CRPS, NLL, RMSE, PICP, and MPIW. Blue highlights the best performance, while red highlights the worst performance. The arrows alongside the metrics indicate which direction is better for that specific metric.}
    \label{fig:all_metrics_reg}
\end{figure}

Figure  \ref{fig:all_metrics_reg} shows a comparative summary of empirical performance in the PM-value regression task. We observe that all models perform consistently well with slight variations. The BNN model performs better in metrics that assess the quality of a probabilistic forecast (i.e., in CRPS and NLL). This is expected since BNNs provide the closest approximation to Bayesian inference, while other models provide only a crude approximation. Interestingly, the GNNs with MC dropout perform very closely to BNNs in CRPS and NLL.  By scoring better in CRPS, BNNs also score better in accuracy metrics (i.e., RMSE) since the CRPS generalizes the MAE to a probabilistic setting.

The PICP and MPIW are conflicting metrics that simultaneously assess the quality of the generated prediction interval. For example,  by increasing the width of a prediction interval (higher MPIW), more values will be inside the predicted intervals (higher PICP). Thus, we observe that BNNs perform well in PICP but poorly in MPIW.

\begin{figure}[!htb]
    \centering
    \includegraphics[width=\linewidth]{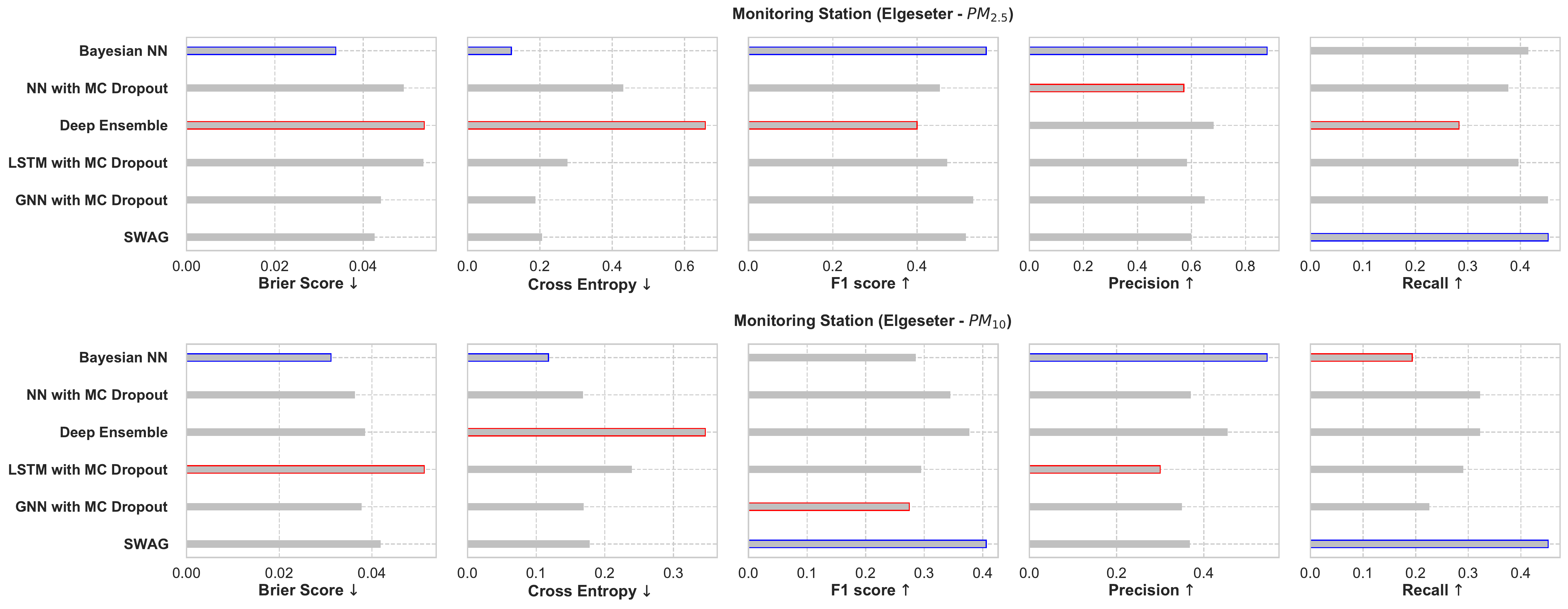}
    \caption{Comparison of empirical performance of the selected probabilistic models in the threshold exceedance classification task.  The comparison is according to five performance metrics (left to right): Brier score, cross-entropy, FI score, precision, and recall. Blue highlights the best performance, while red highlights the worst performance.}
    \label{fig:all_metrics_class}
\end{figure}

Figure \ref{fig:all_metrics_class} shows a comparative summary of empirical performance in the threshold exceedance classification task. We observe that the BNN model performs better in metrics that measure the quality of probabilistic predictions (scoring rule): Brier score and cross-entropy. We also observe that the performance is inconclusive in metrics intended for deterministic classification (F1, precision, recall). This shows that these metrics are not appropriate for probabilistic prediction. Additionally, these metrics are biased by class imbalance since threshold exceedance is a rare event. 

\subsection{Reliability of Confidence Estimate}

In Section \ref{sec:empirical_performance}, we evaluate the selected models using metrics of accuracy and predictive probabilities separately. However, for decision-making, it is crucial to avoid over-confident, incorrect predictions. Therefore, evaluating the reliability of a confidence estimate is indispensable when selecting probabilistic models. One approach to evaluate reliability is to measure the amount of loss (or the number of incorrect predictions) a model makes when its confidence is above a certain threshold. This is a slight variation of the \textit{accuracy-versus-confidence} technique suggested by Lakshminarayanan et al. \cite{lakshminarayanan2017simple} for the classification task. 

\begin{figure}[!htb]
    \centering
    \includegraphics[width=0.9\linewidth]{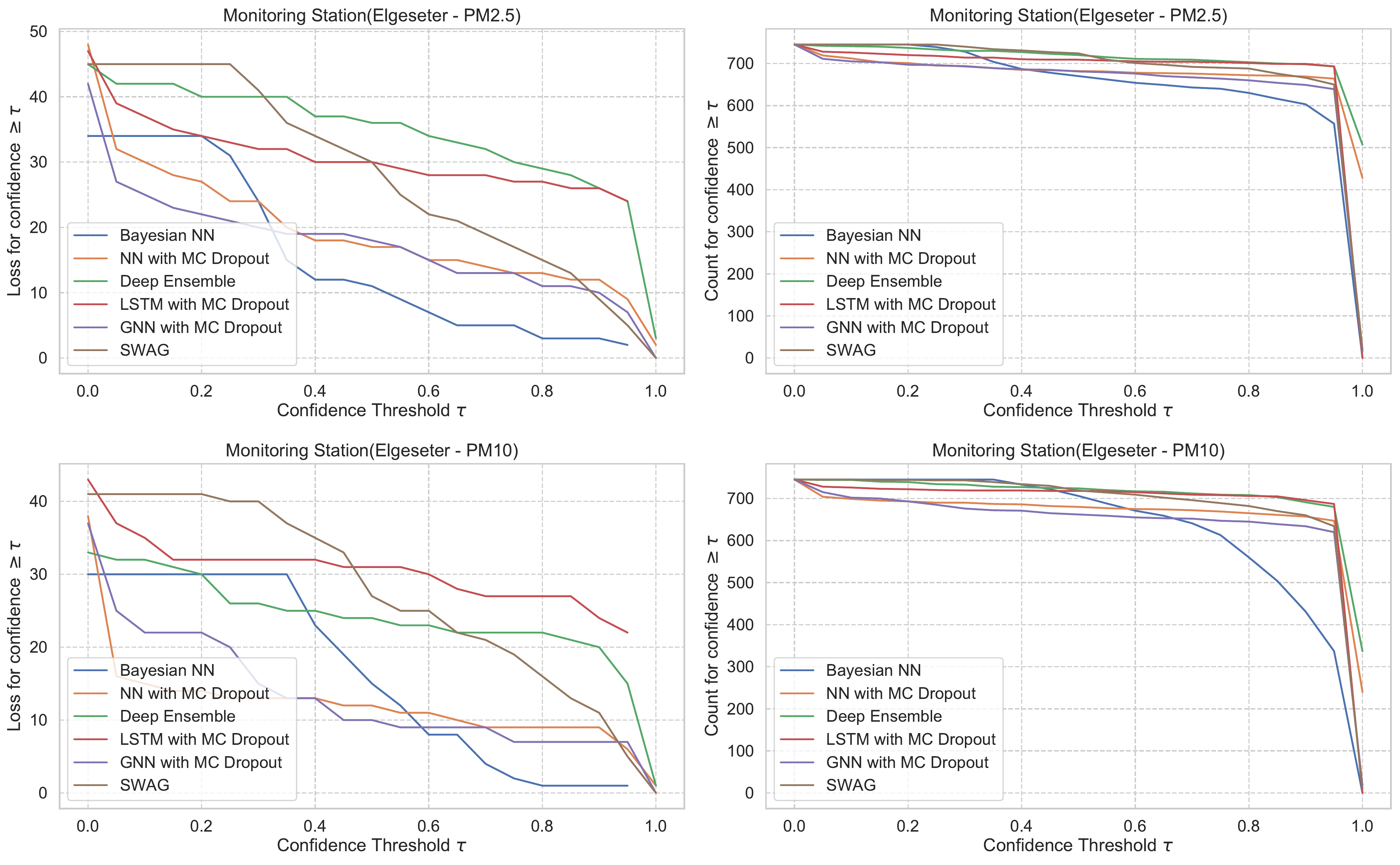}
    \caption{Comparison of confidence reliability for the selected probabilistic models in the threshold exceedance task. \textbf{Left:} loss versus confidence. \textbf{Right:} count versus confidence. The selected models produce are rational, which means the loss-vs-confidence curves are monotonically decreasing.}
    \label{fig:incorrect_predictions_vs_confidence}
\end{figure}

For the threshold classification task, the confidence interval is bounded between 0 and 1 ($0 \leq CI \leq 1 $). Thus, we can define a model confidence to be: $\mathit{confidence}= 1 - \mathit{CI} $ and a confidence threshold $ 0 \leq \tau \leq 1$. Then we plot the number of incorrect predictions the model makes when its confidence is above $\tau$. We expect the curve to be monotonically decreasing for a reliable confidence estimate since a rational model has fewer incorrect predictions at high confidence. Additionally, we plot the total number of predictions (in our case, number of hours in a month of test set) as a function of $\tau$. This curve is monotonically decreasing, but the decreasing rate indicates the amount of confidence in a model. The decreasing rate would be high in a model with lower confidence since it makes fewer predictions with high confidence. Figure \ref{fig:incorrect_predictions_vs_confidence} shows plots of loss vs. confidence and count vs. confidence in the threshold classification task. We observe that the BNN model is more reliable by making fewer incorrect predictions at high confidence but that it has the lowest confidence.

\begin{figure}[!htb]
    \centering
    \includegraphics[width=0.9\linewidth]{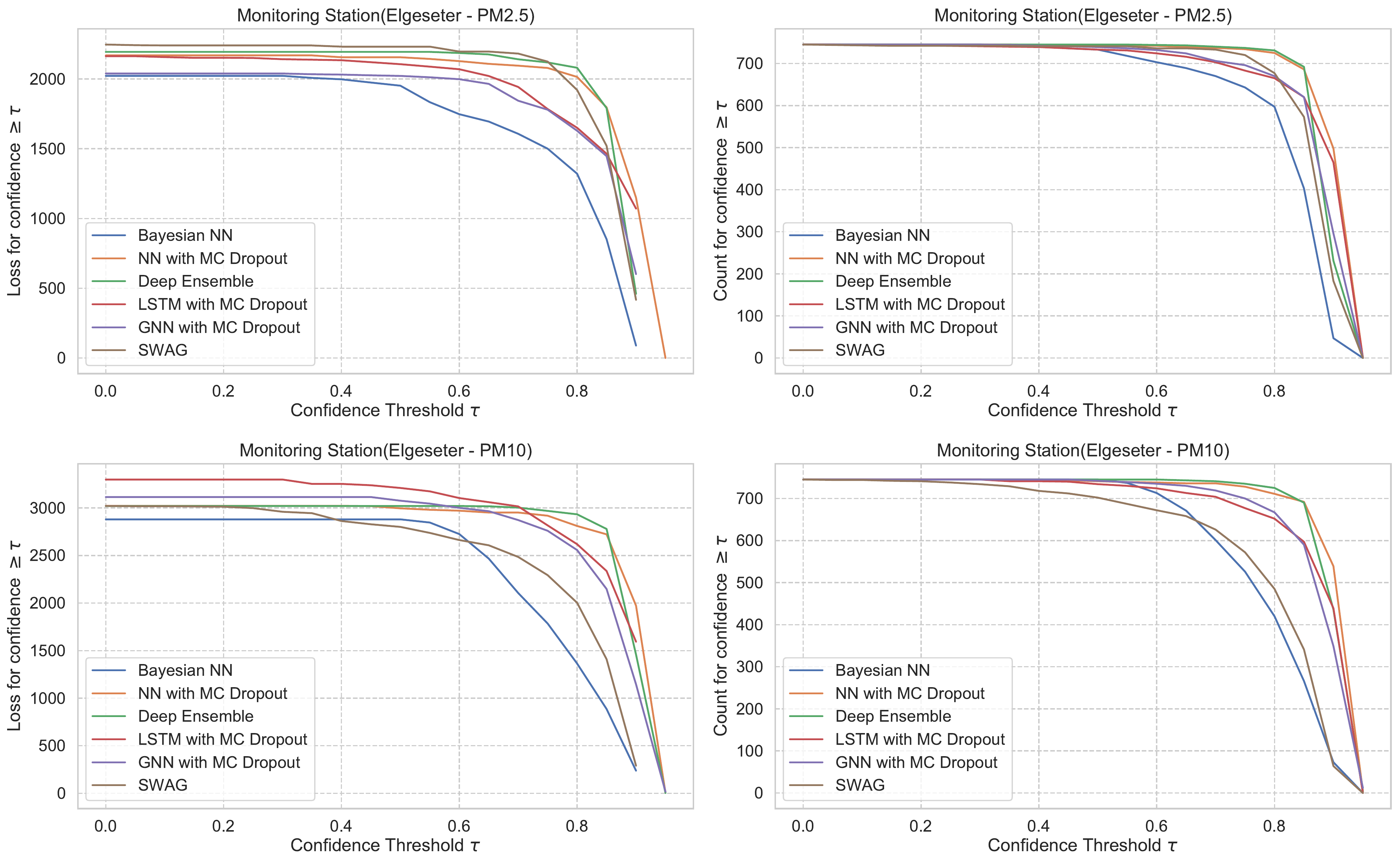}
    \caption{Comparison of confidence reliability for the selected probabilistic models in the PM-value regression task. \textbf{Left:} loss versus confidence. \textbf{Right:} count versus confidence.}
    \label{fig:loss_vs_confidence}
\end{figure}

For the PM-value regression task, the confidence interval is unbounded ($0 < CI < \infty $). However, to compare models, we can normalize and calculate their relative confidence to be bounded between 0 and 1. Then we plot the regression loss as a function of the confidence threshold $\tau$. Figure \ref{fig:loss_vs_confidence} show the plots of loss vs. confidence and count vs. confidence in the PM-value regression task. We observe that the selected models produce rational behaviors and that the BNN model is more reliable but has the lowest confidence. The curves are smooth since the loss is continuous in the regression task.

\begin{figure}[!htb]
    \centering
    \includegraphics[width=0.9\linewidth]{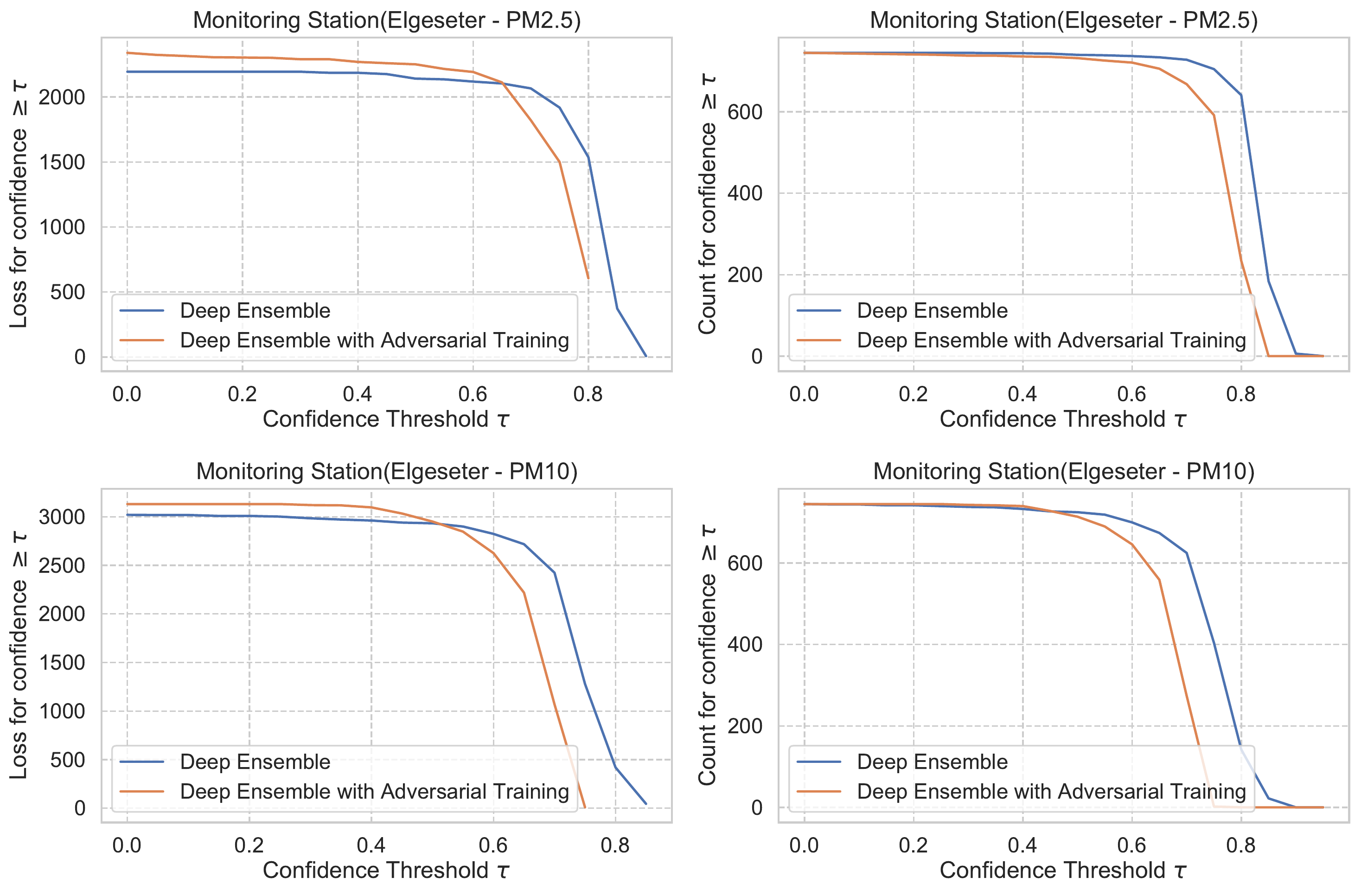}
    \caption{Impact of adversarial training on predictive uncertainty in PM-value regression, using deep ensemble as an example. \textbf{Left:} loss versus confidence. \textbf{Right:} count versus confidence.}
    \label{fig:loss_vs_confidence_adv_ensemble}
\end{figure}

Figure \ref{fig:loss_vs_confidence_adv_ensemble} shows the impact of adversarial training in the PM-value regression task, with deep ensembles as an example. We observe that adversarial training can reduce overconfident predictions.

\subsection{Risk-informed Decisions}
\label{sec:informed_decision_making}

The primary motivation of quantifying epistemic uncertainty is to represent how much the model does not know. Evaluating specific decision policies is out of the scope of this paper, as they depend on the specific costs of countermeasures or the cost of high pollution values. Nonetheless, to show the value of probabilistic models, we investigate the practical impact on decision-making using a case of urban management. Recently, the Norwegian government has proposed strict regulations for particle dust thresholds. That means if the particle dust exceeds a specific threshold, the government will impose a penalty on the municipality for violating pollution regulations. Therefore, the municipality has to decide whether to implement countermeasures in advance if a forecasting model predicts threshold exceedance events.

The challenge is that not all forecasts are correct and occasionally result in false positives and false negatives. A false positive is associated with implementing unnecessary countermeasures, while a false negative is associated with violating pollution regulations. The costs of these two scenarios are context-dependent. Therefore, it is helpful to have a forecasting model that also provides potential flexibility and tradeoffs depending on the cost of false positives and false negatives.

\begin{figure}%
    \subfloat[\centering Non-probabilistic model] {{\includegraphics[width=0.46\linewidth]{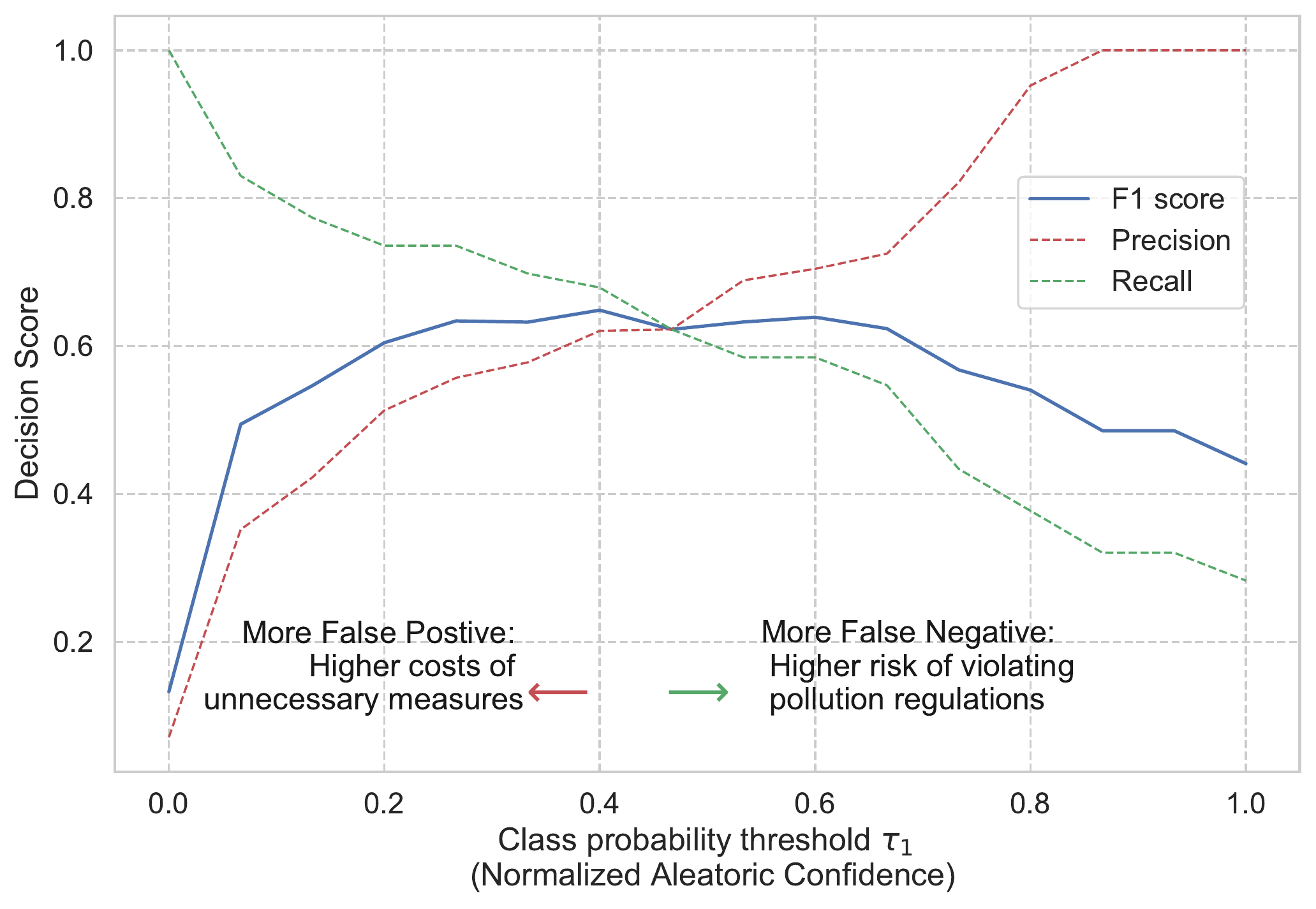} }}%
    \qquad
    \subfloat[\centering Probabilistic model]{{\includegraphics[width=0.46\linewidth]{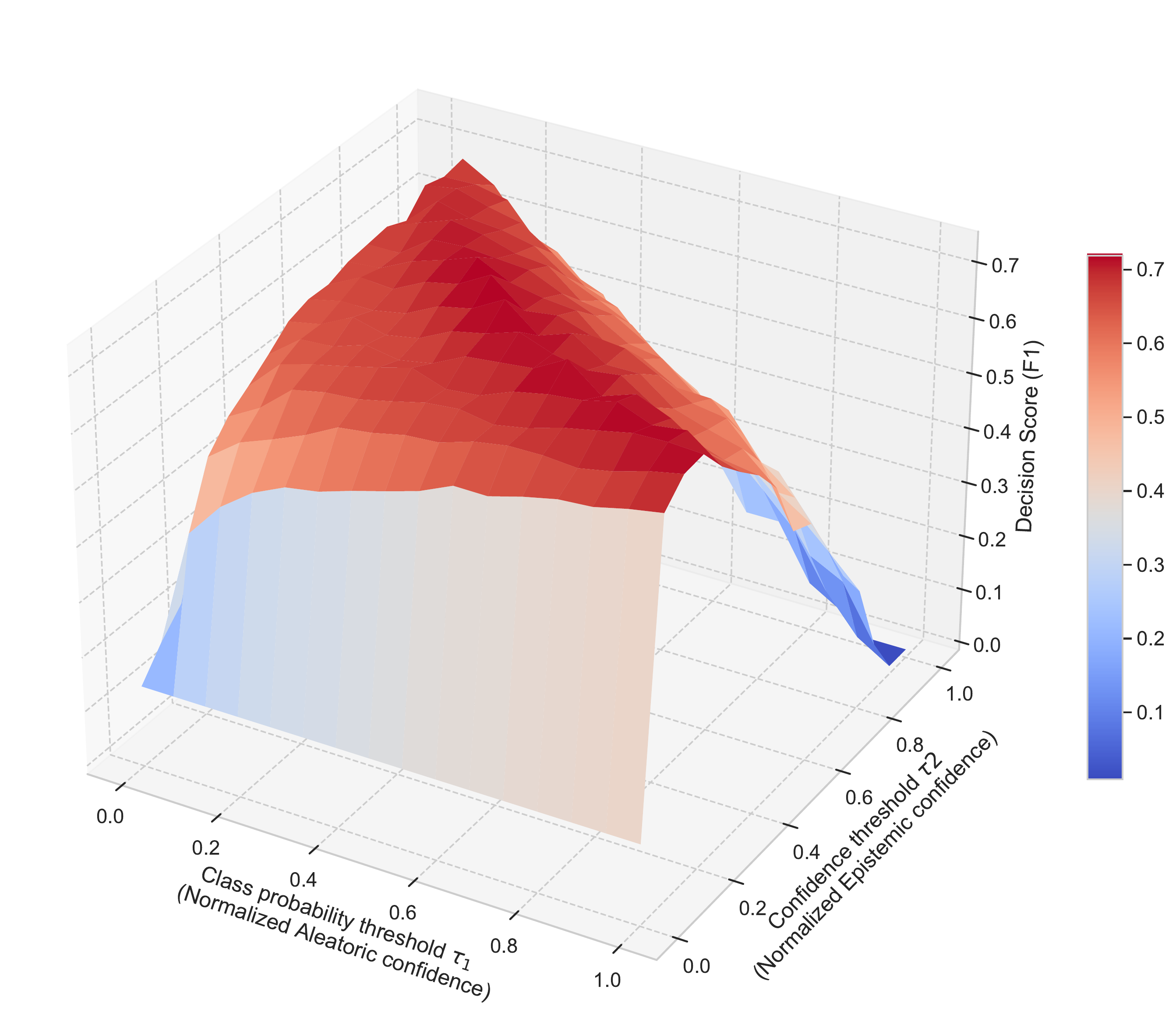} }}%
    \caption{Comparison of decision score in non-probabilistic and probabilistic models. \textbf{(a)} Decision score in a non-probabilistic model as a function of class probability threshold ($\tau_1$ corresponding to aleatoric confidence).
    \textbf{(b)} Decision score in a probabilistic model as a function of both class probability threshold ($\tau_1$) and model confidence threshold ( $\tau_2$ corresponding to epistemic uncertainty).}%
    \label{fig:informed_decision_making}%
\end{figure}

Non-probabilistic models already offer some aspect of this tradeoff by leveraging the aleatoric uncertainty. For example, we can adjust the probability threshold in a binary classification, i.e., implement countermeasures, only if the predictive probability is above $\tau_1$. Figure \ref{fig:informed_decision_making}a shows the decision score (in terms of F1, precision, and recall) as a function of aleatoric confidence in a non-probabilistic model. Accepting the model decision even when its predictive probability is low will result in more false positives and thus higher costs of unnecessary countermeasures. On the other hand, accepting the model decisions only when its predictive probability is high will result in more false negatives, thus increasing the risk of violating pollution regulations. 

In probabilistic models, we can also leverage epistemic uncertainty to obtain a higher degree of tradeoffs. That means we implement countermeasures only if the predictive probability is above $\tau_1$ \textbf{and} the model confidence is above a certain threshold $\tau_2$. Figure \ref{fig:informed_decision_making}b shows the resulting decision score as a function of aleatoric and epistemic confidence in a probabilistic model. We use a BNN model as a representative example. To allow a fair comparison, we use a non-probabilistic model with the same architecture and trained with the same conditions. We observe that the probabilistic model provides a wider area of control over the risk profile based on the costs of false positives versus false negatives. In this case, the probabilistic model scores better over a wider range of $\tau_1$ and $\tau_2$ than a non-probabilistic model. This confirms that probabilistic models are more suitable for making more informed and risk-aware decisions.

\subsection{Practical Applicability}

In Section \ref{sec:empirical_performance}, we observe that a BNN model performs better in a variety of metrics. Further, BNNs offer an elegant approach to represent model uncertainty without strong assumptions. However, they are computationally demanding, require more parameters, are sensitive to hyperparameter choices, and take longer to converge than non-Bayesian neural networks. Thus, BNNs are difficult to work with, especially on larger datasets. Additionally, the nature of variational approximation and the choice of the prior or posterior distributions bias the predictive uncertainty.

From a practical perspective, MC Dropout, SWAG, and deep ensemble are convenient and scalable approaches to obtain uncertainty directly from standard neural networks without changing the optimization. Additionally, they have fewer hyperparameters and can be applied independently from the underlying network architecture (such as standard neural network, LSTM, and GNNs). However, these approaches require strong assumptions and provide a crude approximation of the Bayesian inference. During inference, they have high computational costs during inference (proportional to the number of samples $M$). Further, the uncertainty is estimated by measuring the diversity of predictions that depends on the network's architecture, size, and properties of the training data and initialization parameters. Accordingly, there is no guarantee that they produce a diverse prediction for an uncertain input.

The deep ensemble is a convenient non-Bayesian approach that captures aspects of multi-modality. It is better than the bootstrap ensemble in which independent models are trained using independent datasets. The deep ensemble trains every model using the whole dataset since stochastic optimization and random initialization make the models sufficiently independent. However, training deep ensemble is more expensive than SWAG or MC dropout since we need to train $M$ different models. 

The underlying assumption used in MC dropout is that it approximates a probabilistic deep Gaussian process. This assumption is based on the idea that a single-layer neural network tends to converge to a Gaussian process in the limit of an infinite number of hidden units \cite{williams1997computing}. Our results show that MC dropout provides less reliable uncertainty estimates (Figures \ref{fig:incorrect_predictions_vs_confidence} and \ref{fig:loss_vs_confidence}), which confirms the results recently presented in other domains, such as computer vision \cite{gustafsson2020evaluating} and molecular property prediction \cite{scalia2020evaluating}. 

The related works that address quantifying uncertainty in a data-driven forecast of air quality uses either quantile regression (QR) \cite{aznarte2017probabilistic}, Gaussian processes (GP) \cite{pucer2018bayesian}, or ConvLSTM with MC dropout \cite{mokhtari2021uncertainty}. We have reimplemented these methods in our problem setting to compare them to the proposed models since we have different datasets and different data inputs. We use a Gradient Tree Boosting \cite{friedman2001greedy} for quantile regression and GPyTorch \cite{gardner2018gpytorch} for Gaussian processes. Table \ref{tbl:comparision_to_previous_work} summarizes the comparison results in the PM-value regression task in one representative monitoring station. We observe that the proposed models perform better, especially in metrics that assess the quality of a probabilistic forecast (i.e., in CRPS and NLL). Quantile regression only estimates the conditional quantile and not a probabilistic distribution. Thus, we can only assess the quality of its prediction interval and not a probabilistic forecast. Gaussian processes offer a Bayesian formalism to reason about uncertainty. However, they have the highest computational cost since they require an inversion of the kernel matrix, which has an asymptotic complexity of $\mathcal{O}(n^3)$, where $n$ is the number of training examples.

\begin{table}[!htb]
\small
\centering
\captionsetup{justification=centering}
 \caption{Comparison of the previous works and the proposed models when quantifying uncertainty in data-driven forecast of air quality}
 \begin{tabular}{lccccccc}
 \hline
 \multicolumn{1}{c}{\bf Metric} & \multicolumn{1}{c}{\bf QR\cite{aznarte2017probabilistic}} & \multicolumn{1}{c}{\bf GP \cite{pucer2018bayesian}} & \multicolumn{1}{c}{\bf ConvLSTM\cite{mokhtari2021uncertainty}}& \multicolumn{1}{c}{\bf BNN } & \multicolumn{1}{c}{\bf Deep Ensembles } & \multicolumn{1}{c}{\bf GNN } & \multicolumn{1}{c}{\bf SWAG } \\
  \hline

 \bf RMSE$\downarrow$  &6.17 &6.45     &6.46 &\bf{5.17}  &5.59      &5.80 &5.76\\
 \bf PICP$\uparrow$    &0.72 &\bf{0.92}&0.71 &0.90     &0.69      &0.79 &0.76\\ 
 \bf MPIW$\downarrow$  &14.79&36.96    &12.18&25.07    &\bf{12.11} &15.07&16.61\\
 \bf CRPS$\downarrow$  &NA   &0.53     &0.56 &\bf{0.47}&0.51      &0.50 &0.53\\
 \bf NLL$\downarrow$   &NA   &1.36     &2.48 &\bf{1.28}&2.38      &1.60 &1.96\\
\hline
 \end{tabular}

 \label{tbl:comparision_to_previous_work}
 \end{table}

\section{Conclusion}
\label{sec:conclusion}

This work presents a broad empirical evaluation of the relevant state-of-the-art deep probabilistic models applied in air quality forecasting. Through extensive experiments, we describe training these models and evaluating their predictive uncertainties using various metrics for regression and classification tasks. We introduce a new state-of-the-art example for air quality forecasting by defining the problem setup and selecting proper input features and models. Then, we apply uncertainty-aware models that exploit the temporal and spatial correlation inherent in air quality data using recurrent and graph neural networks. We propose improving uncertainty estimation using "free" adversarial training to locally smooth the prediction distribution along the adversarial direction with virtually no additional cost. Finally, we show how data-driven probabilistic models can improve the current practice using a real-world example of air quality forecasting in Norway. Particularly, we show the practical impact of uncertainty quantification and demonstrate that probabilistic models are more suitable for making informed decisions.

The results show that the proposed models perform better than previous works in quantifying uncertainty in data-driven air quality forecasts. BNNs provide a more reliable uncertainty estimate but with challenging practical applicability. Additionally, the results demonstrate that MC Dropout, SWAG, and deep ensemble have less reliable uncertainty estimates, but they are more convenient in practical applicability. Our contribution hereof is not about specific model selection but more about navigating the larger design space and the different tradeoffs offered by various probabilistic models. 

Future work includes exploring hybrid air quality forecasting combining physics-based and data-driven probabilistic models. Additionally, future work should address uncertainty estimation with out-of-domain or seasonal variations in air quality forecasting. In our experiments, we used a reasonable search space for hyperparameters and network architectures. Thus, future work could target an exhaustive search of hyperparameters, additional data sets, and models (e.g., transformers).

\begin{Funding}
This work is funded by the European Union’s Horizon 2020 research and innovation program, project AI4EU, grant agreement No. 825619.
\end{Funding}

\begin{ack}
We would like to thank Sigmund Akselsen for reading a draft of the paper and providing detailed feedback on the work. We also thank Tiago Veiga for helping with a script to fetch the datasets.
\end{ack}

\bibliographystyle{unsrt}


\appendix

\section{Datasets}
 \label{sec:appendix_dataset}
 
 This section shows additional figures explaining the used dataset of air quality, meteorological data, traffic, and street-cleaning reports from the municipality.
 \subsection{Air Quality Data}
 
 \setcounter{figure}{0}
 
\begin{figure}[H]
    \centering
    \includegraphics[width=\linewidth]{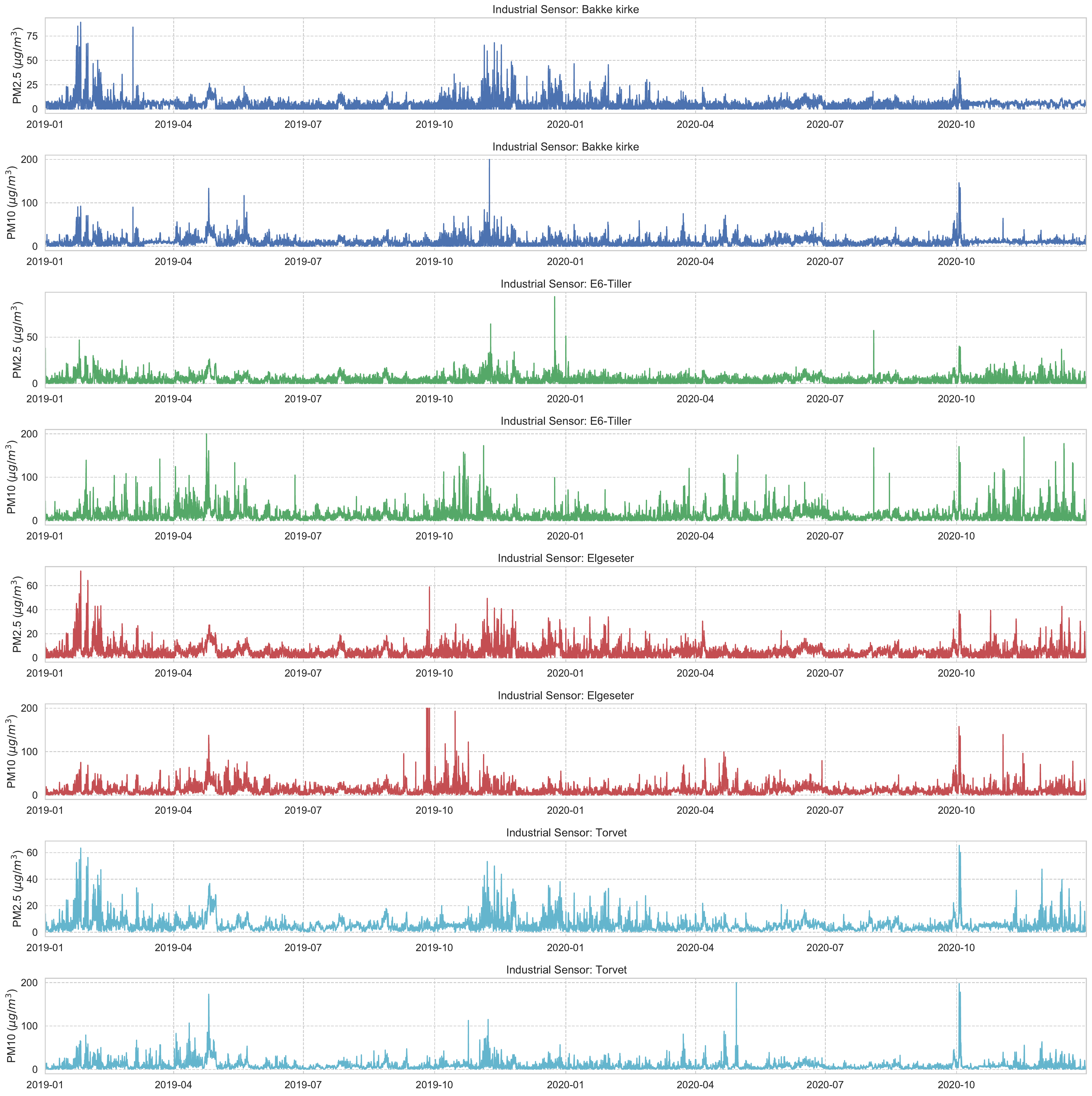}
    \caption{Air quality data of $PM_{2.5}$ and $PM_{10}$, measured over two years in four different sensing stations in the city of Trondheim. These data are offered by the Norwegian Institute for Air Research (NILU) (\url{https://www.nilu.com/open-data/}).}
    \label{fig:appendix_aq_data}
\end{figure}

Figure \ref{fig:appendix_aq_data} shows the air quality data of $PM_{2.5}$ and $PM_{10}$, measured over two years in four different sensing stations in the city of Trondheim (\ref{fig:NILU_Sensors}). The figure shows a clear temporal and spatial correlations between the four sensing stations.

\subsection{Weather data}
\begin{figure}[H]
    \centering
    \includegraphics[width=\linewidth]{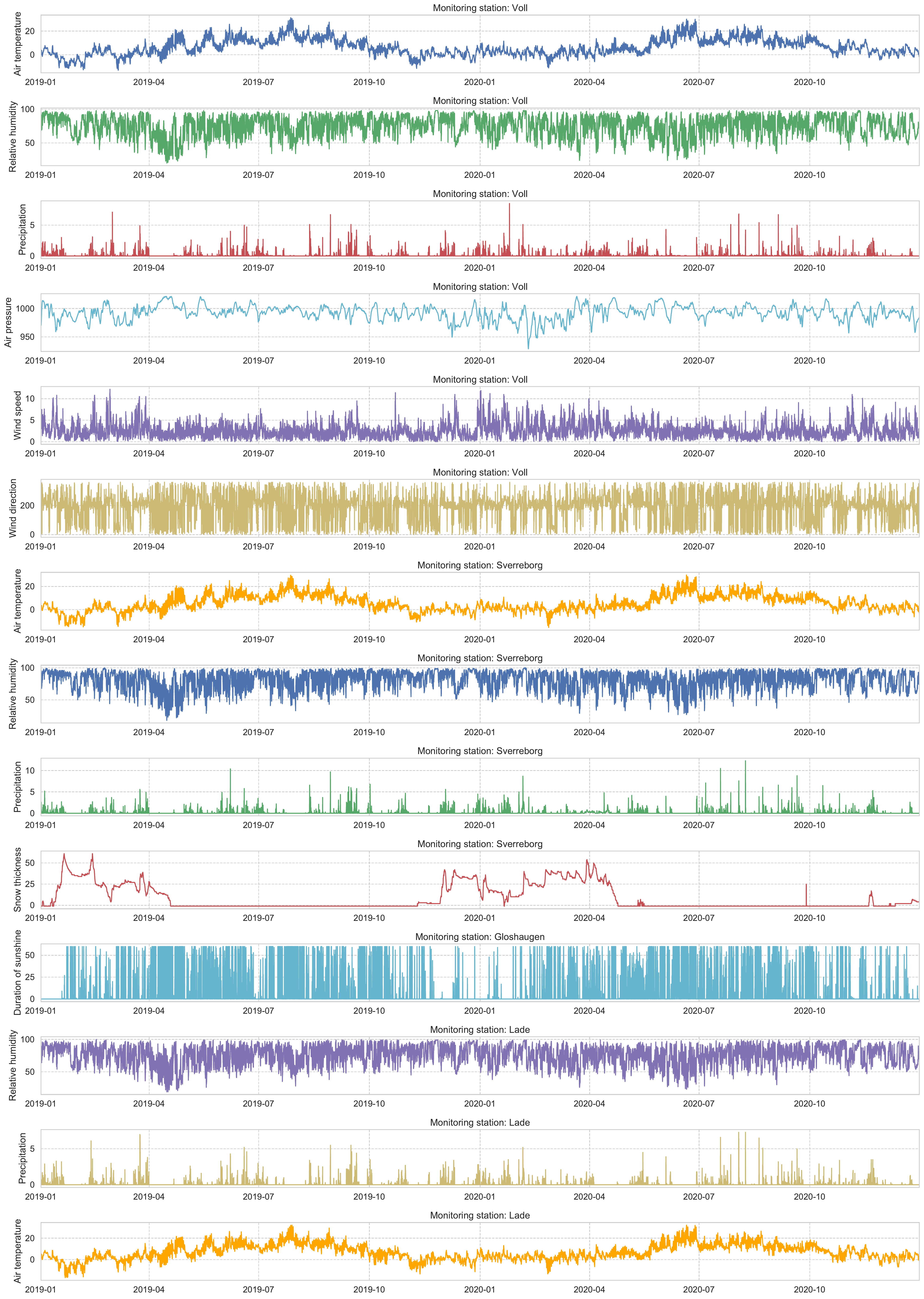}
    \caption{Weather data observations over two years at four monitoring station in the city of Trondeheim (Voll, Sverreborg, Gloshaugen, Lade). These data are offered by the Norwegian Meteorological Institute (\url{https://frost.met.no}).}
    \label{fig:appendix_weather_data}
\end{figure}

We use weather data observations at four monitoring stations in the city of Trondeheim (Voll, Sverreborg, Gloshaugen, Lade). These observations include: air temperature, relative humidity, precipitation, air pressure, wind speed, wind direction, snow thickness, and duration of sunshine. Figure \ref{fig:appendix_weather_data} summarize all these observations.
                 
\subsection{Traffic data}
\begin{figure}[H]
    \centering
    \includegraphics[width=\linewidth]{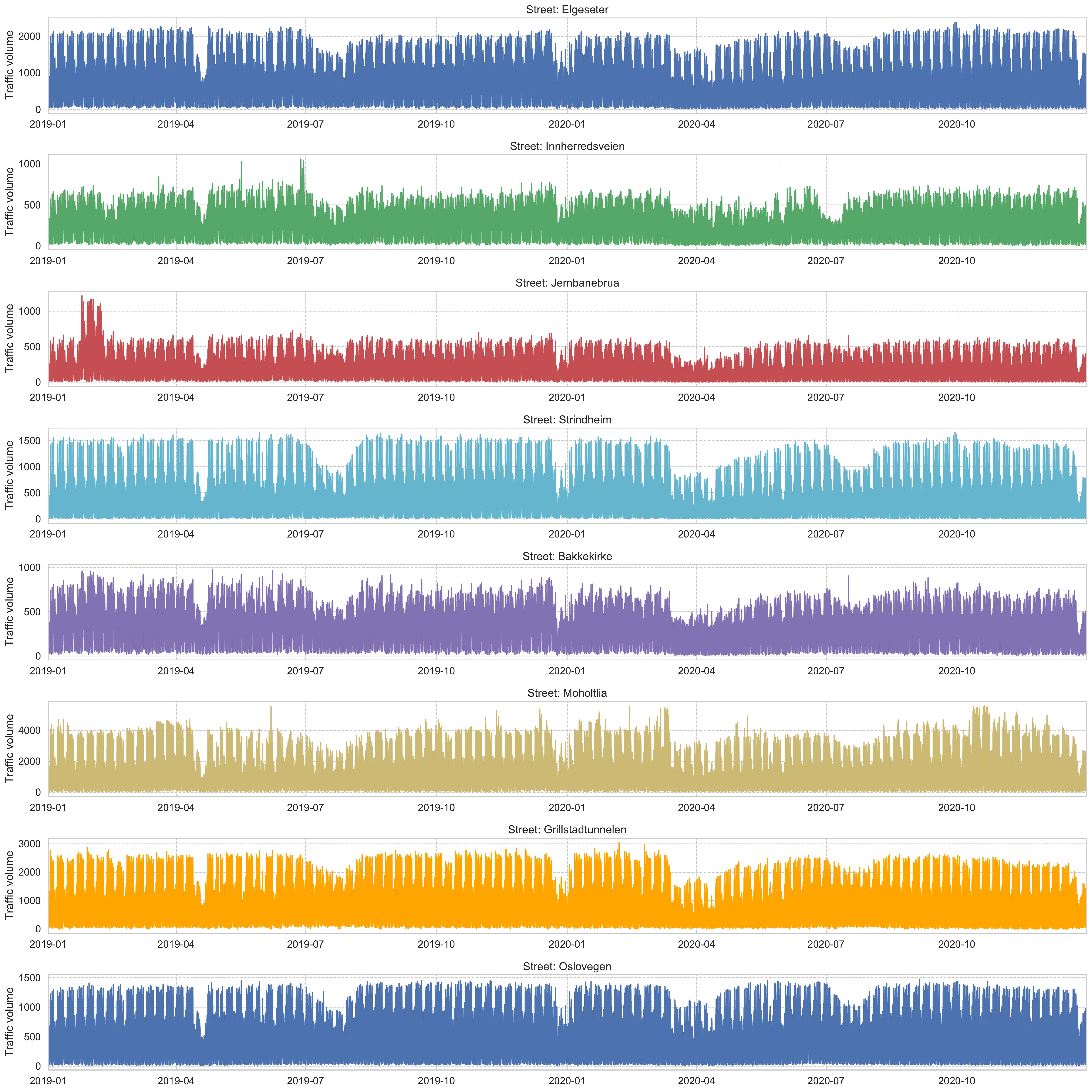}
    \caption{Traffic volume recorded at eight streets of Trondheim over two years. These data are offered by the Norwegian Public Roads Administration (\url{https://www.vegvesen.no/trafikkdata/start/om-api}). }
    \label{fig:appendix_traffic_data}
\end{figure}

We also used traffic volume as an input feature for the forecasting models. We used the data of the traffic volume recorded at eight streets of Trondheim. Figure \ref{fig:appendix_traffic_data} shows the traffic volume recorded over two years. The figure shows a clear correlation of traffic between streets. It also shows the drop in traffic during the pandemic lockdown in Trondheim (after March 12, 2020).

\subsection{Streets-cleaning
data}

\begin{figure}[H]
    \centering
    \includegraphics[width=\linewidth]{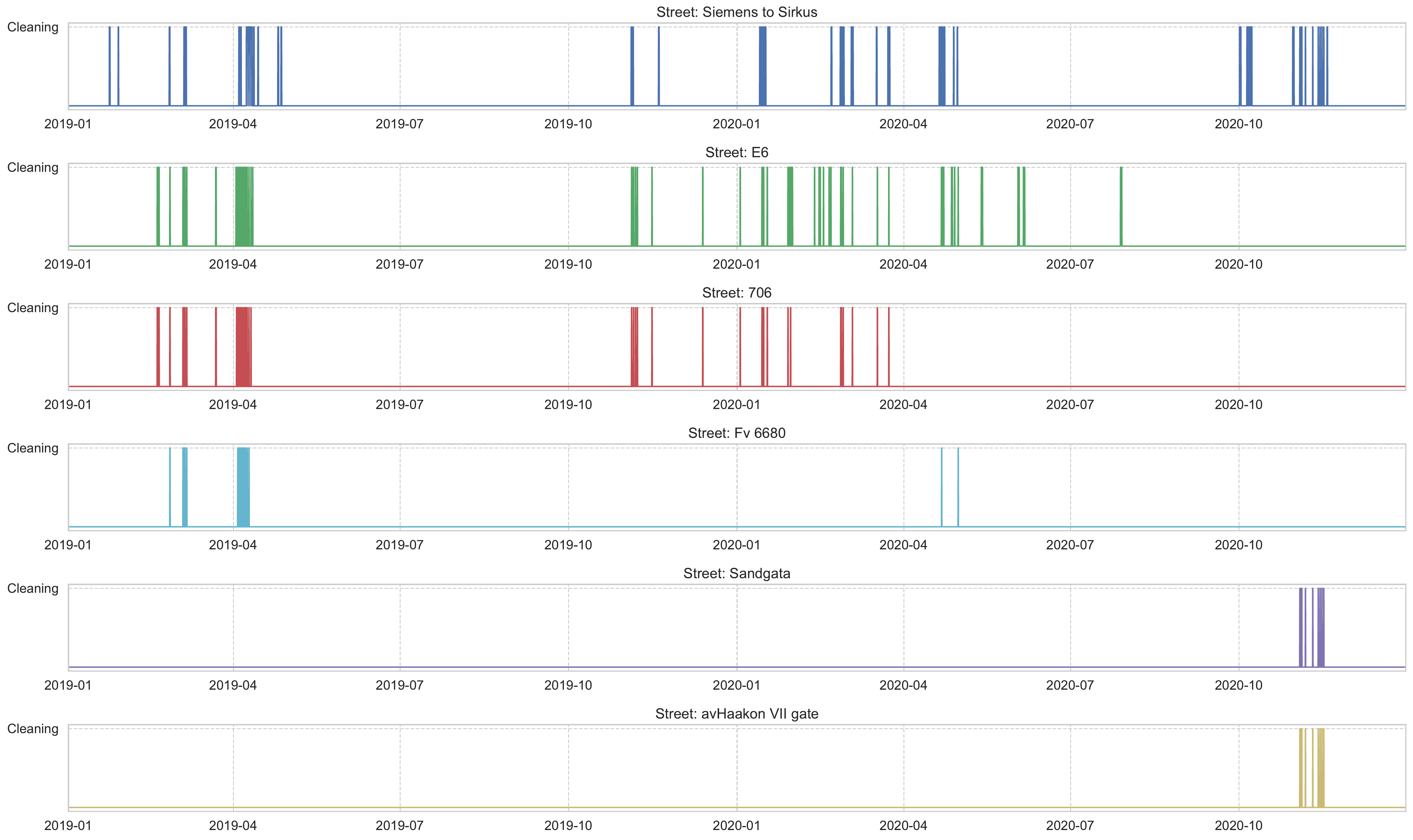}
    \caption{Data of the duration of time in which a street-cleaning is taking place on the main streets of Trondheim, reported by the municipality.}
    \label{fig:appendix_street_cleaning_data}
\end{figure}

We also used data of street-cleaning at main streets of Trondheim, as shown in Figure \ref{fig:appendix_street_cleaning_data}. These data are reported by the municipality and include the duration of time in which a street-cleaning is taking place. 

\section{Reliability of Confidence Estimate: Additional Plots}
\begin{figure}[H]
    \centering
    \includegraphics[width=0.94\linewidth]{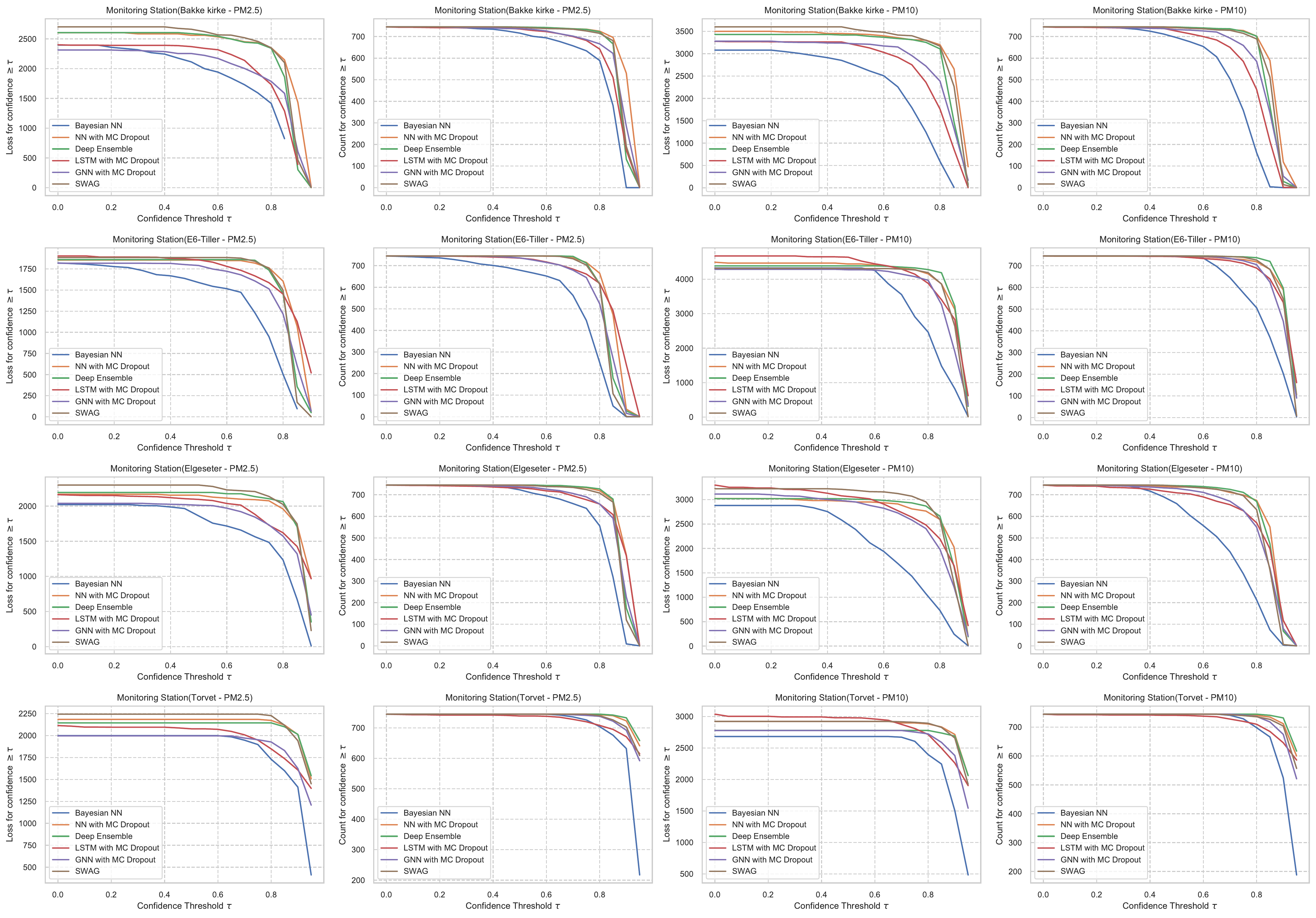}
    \caption{Comparison of confidence reliability for the selected probabilistic models in the PM-value regression task in all monitoring stations.}
    \label{fig:appendix_loss_vs_confidence}
\end{figure}

\begin{figure}[H]
    \centering
    \includegraphics[width=0.94\linewidth]{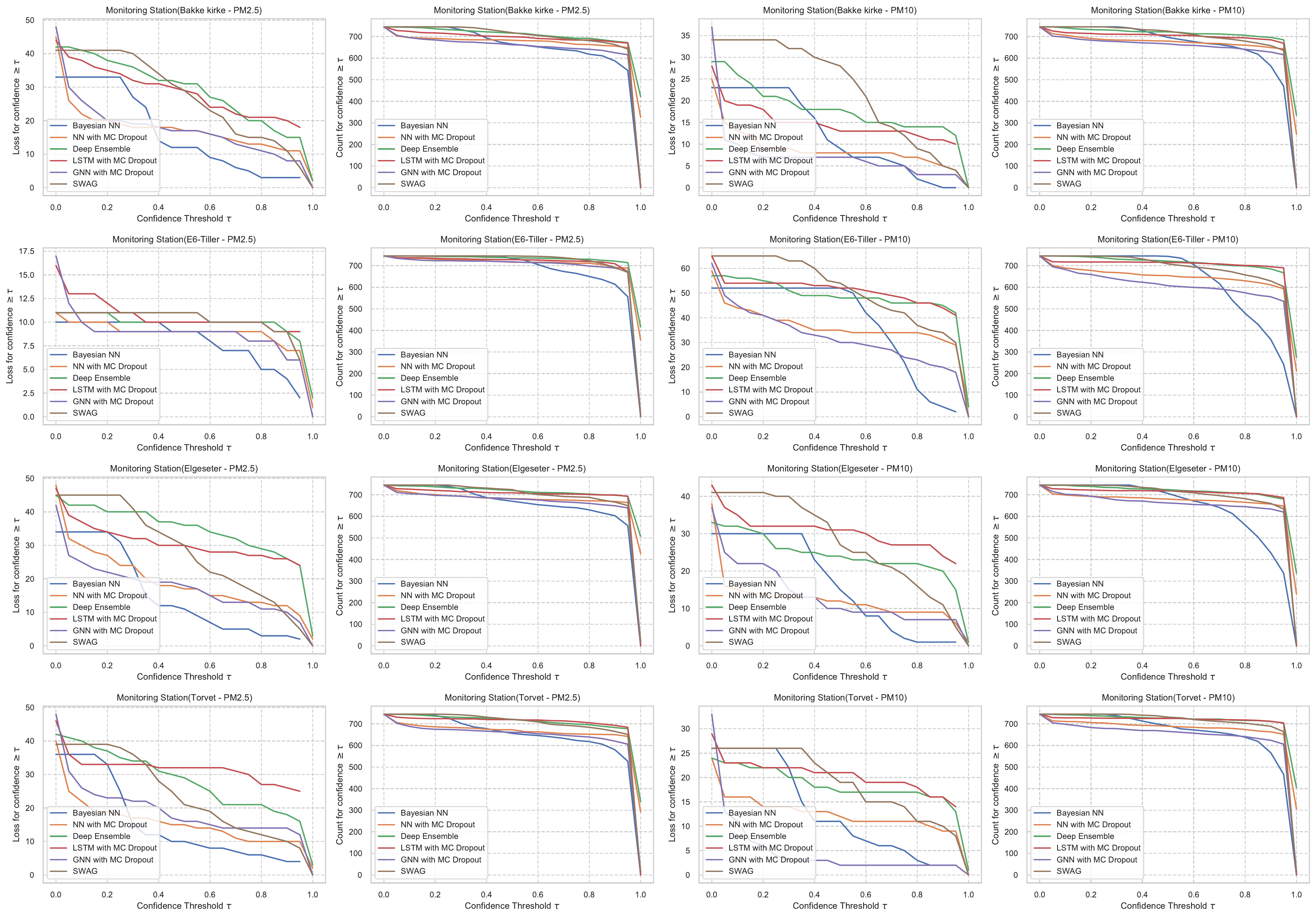}
    \caption{Comparison of confidence reliability for the selected probabilistic models in the threshold exceedance classification task in all monitoring stations.}
    \label{fig:appendix_incorrect_predictions_vs_confidence}
\end{figure}

This section presents additional plots for comparison of confidence reliability in all monitoring stations. Figure \ref{fig:appendix_loss_vs_confidence} shows the comparison of confidence reliability in the PM-value regression task while Figure \ref{fig:appendix_incorrect_predictions_vs_confidence} in the threshold exceedance classification task.

\section{Justification for Threshold-Exceedance Classification}

Recently, the Norwegian government has proposed even stricter regulations for particle dust thresholds. Therefore, we want to estimate if the concentration of air particles exceeds certain thresholds following the Common Air Quality Index (CAQI) used in Europe \cite{CiteairII2012}, as shown in Table \ref{tab:CAQI}.

\begin{figure}[H]
    \centering
    \includegraphics[width=\linewidth]{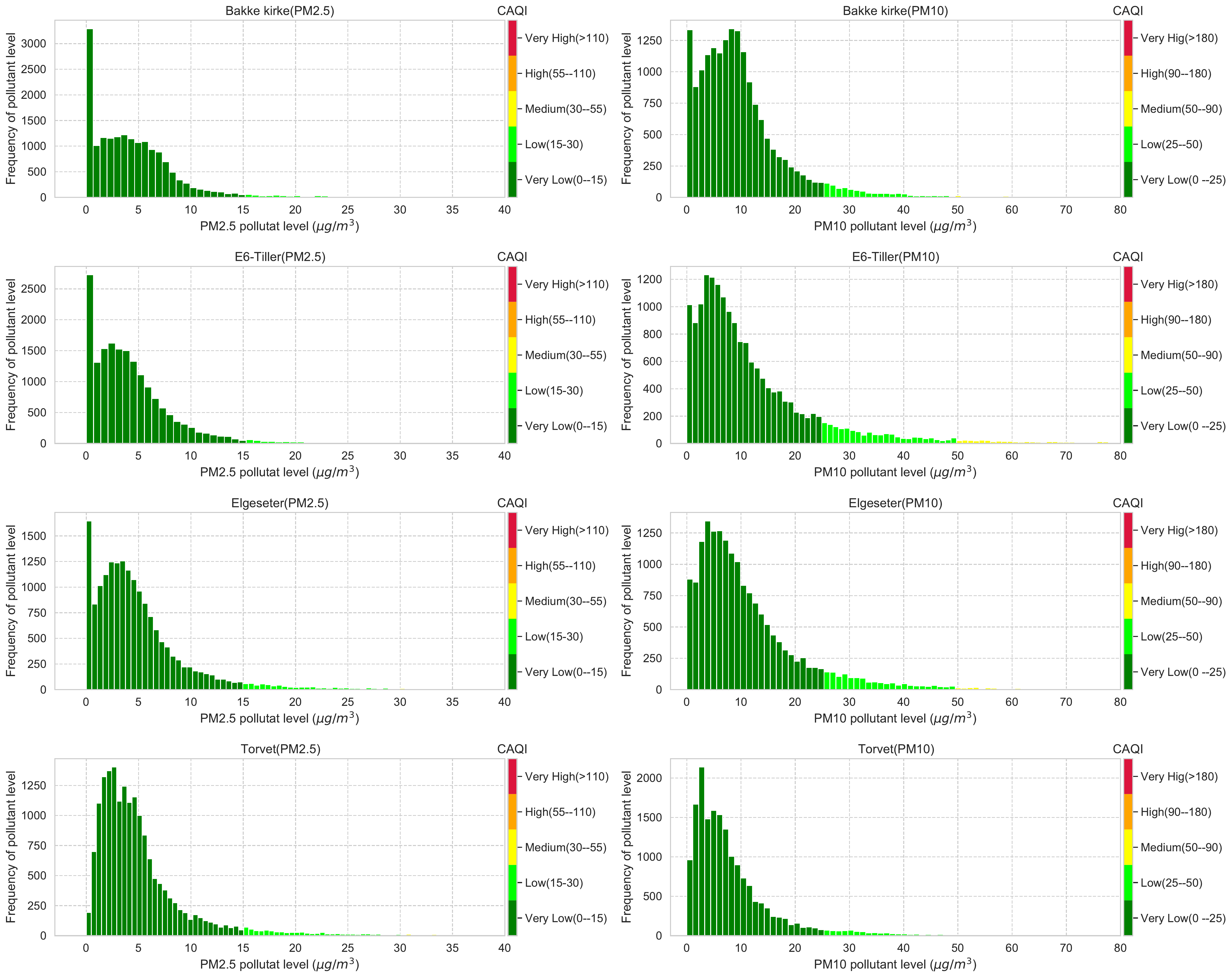}
    \caption{Histograms that approximately represent the distribution of air quality at four monitoring stations in the city of Trondheim. The air quality data come from  heavily right-skewed distributions, in which higher CAQI classes are under-represented.}
    \label{fig:appendix_right_skewed}
\end{figure}

\begin{figure}[H]

    \includegraphics[width=\linewidth]{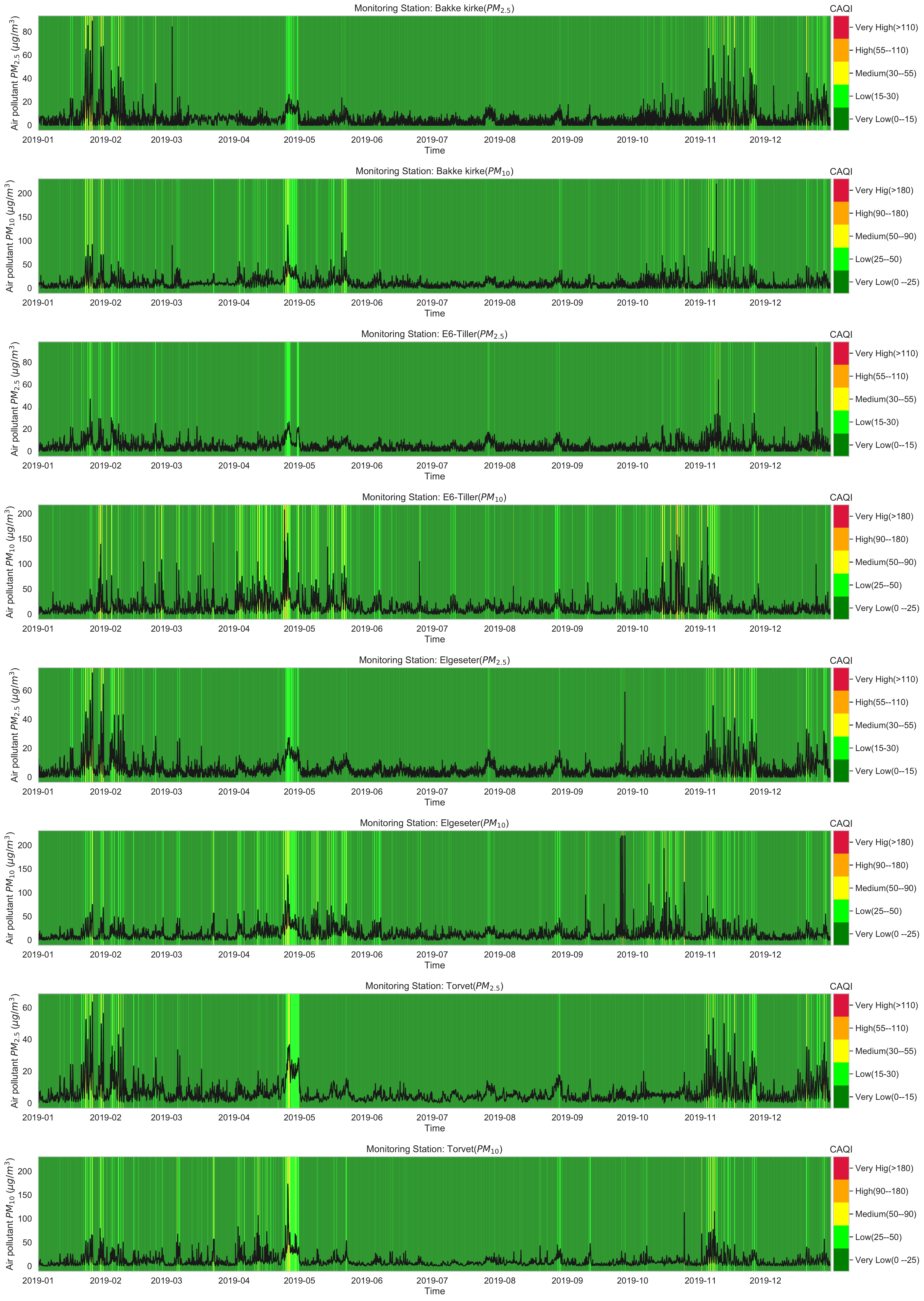}

    \caption{Air quality in the city of Trondheim over one year in all monitoring stations. The data are decomposed into the five CAQI levels of air pollutants. It is usually at \textit{Very Low} and rarely exceeds the \textit{Medium} level.}

    \label{fig:appendix_aq_index_cbar}
\end{figure}

The CAQI specifies five levels of air pollutants, but the air quality in the city of Trondheim is usually at \textit{Very Low} and rarely exceeds the \textit{Medium} level. Figure \ref{fig:appendix_aq_index_cbar} shows the air quality level over one year in all monitoring stations. Generally, air quality data have right-skewed distributions, with the degree of skewness (asymmetry) differing among different cities. Figure \ref{fig:appendix_right_skewed} shows histograms that approximately represent the distribution of air quality at four monitoring stations in the city of Trondheim. The figure clearly shows heavily right-skewed distributions, in which higher CAQI classes are under-represented. Therefore, instead of a multinomial classification task (with five classes), we transform the problem into a threshold exceedance forecast task in which we try to predict the points in time where the air quality exceeds the \textit{Very Low} level.

\vspace{-0.8cm}
\section{Experimental Details of Deep Probabilistic Forecasting Models}
 \vspace{-0.3cm}
 In this section, we show the results of the PM-value regression and threshold exceedance classification in all monitoring stations using the corresponding probabilistic model.
 
\vspace{-0.44cm}
\subsection{Bayesian Neural Networks (BNNs)}
\vspace{-0.8cm}
\label{sec:appendix_bnn}

\begin{figure}[H]
\centering
    \subfloat[\centering PM-value regression] {{\includegraphics[width=0.42\linewidth]{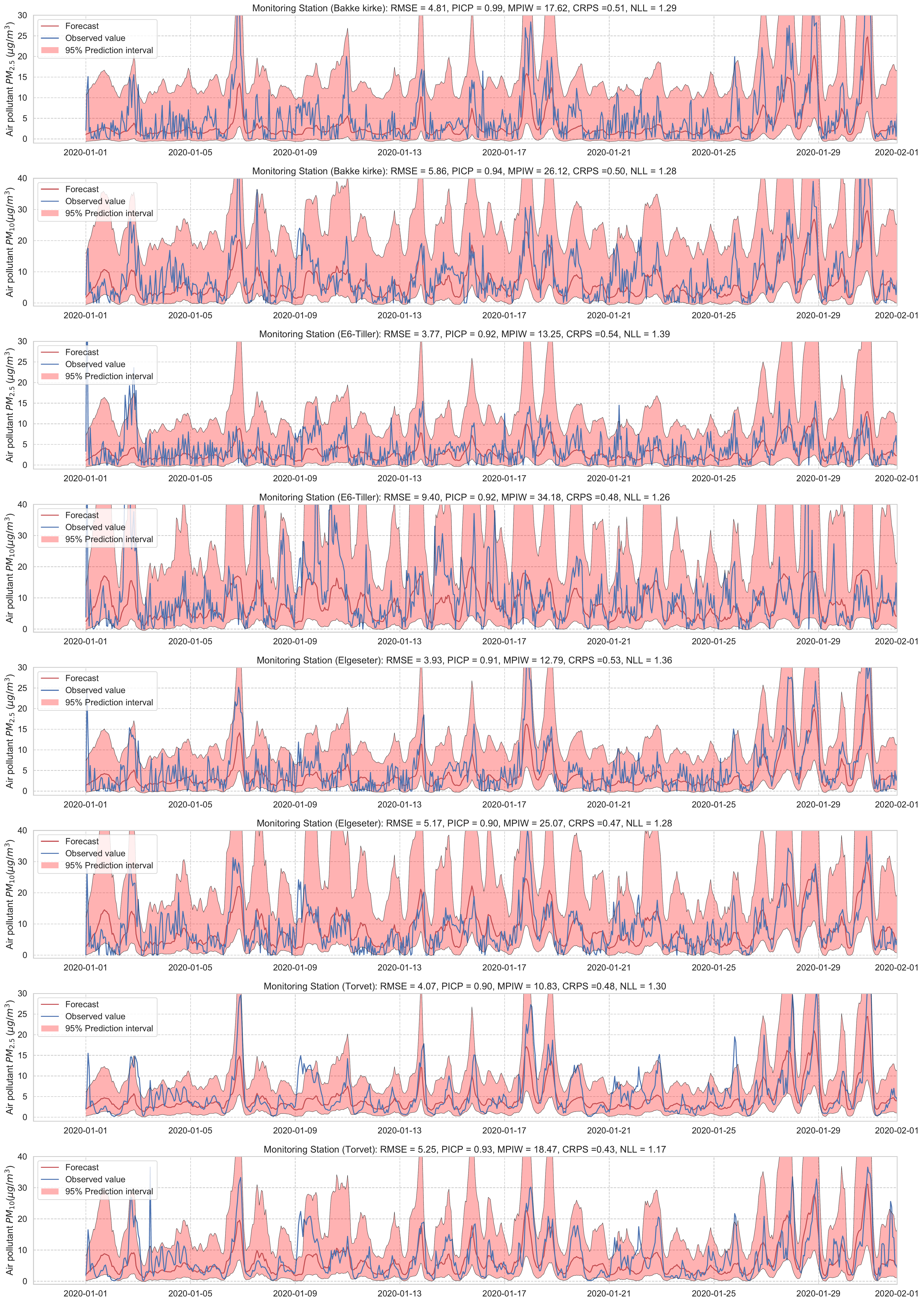} }}%
    \qquad
    \subfloat[\centering Threshold exceedance classification]{{\includegraphics[width=0.42\linewidth]{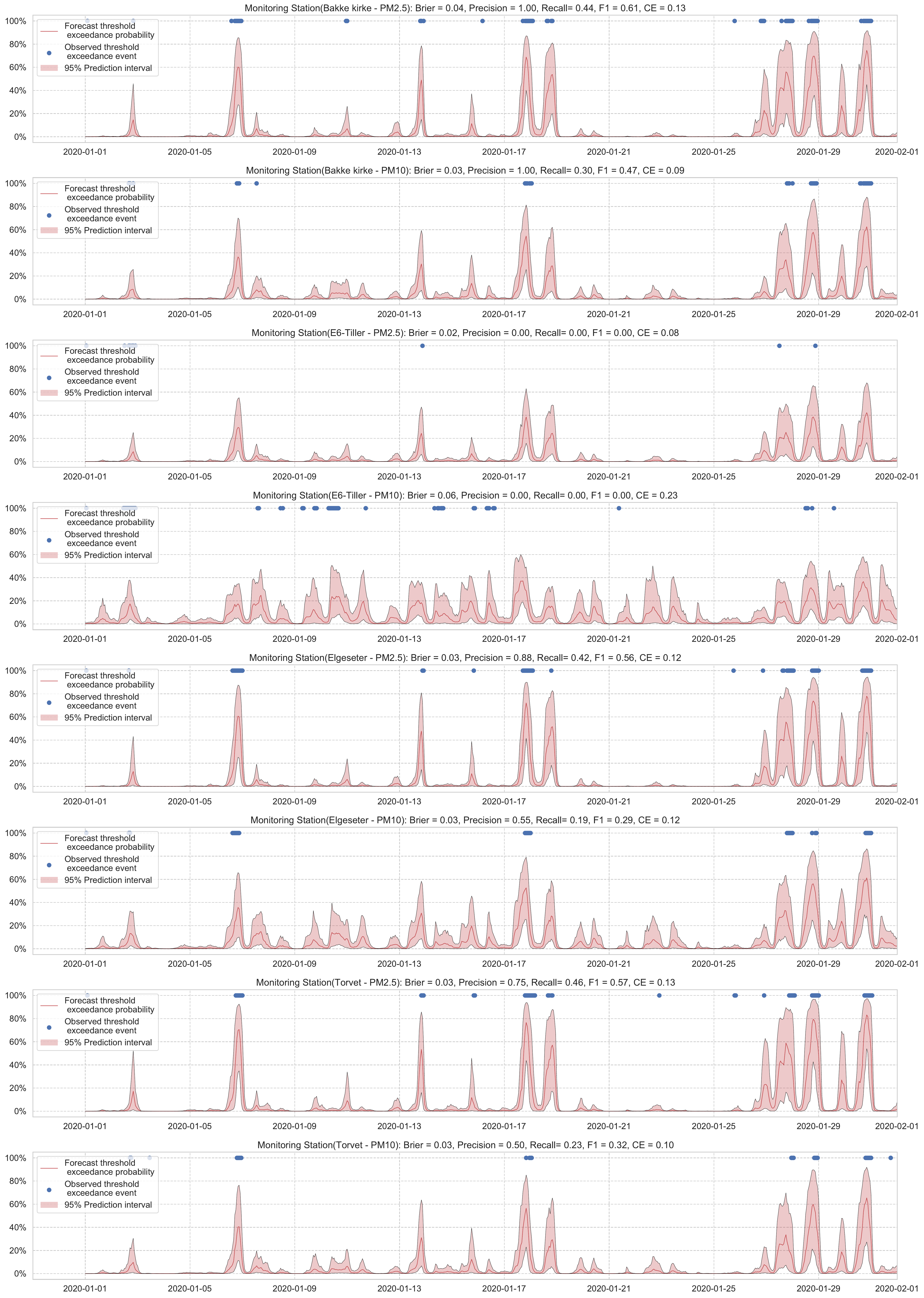} }}%
    \caption{Deep probabilistic forecast of air quality using BNNs at four monitoring stations.}%
    \label{fig:appendix_bnn}%
\end{figure}
\vspace{-0.5cm}
\subsection{Standard NNs with MC Dropout}
\label{sec:appendix_mc}
\vspace{-0.8cm}
\begin{figure}[H]
\centering
    \subfloat[\centering PM-value regression] {{\includegraphics[width=0.42\linewidth]{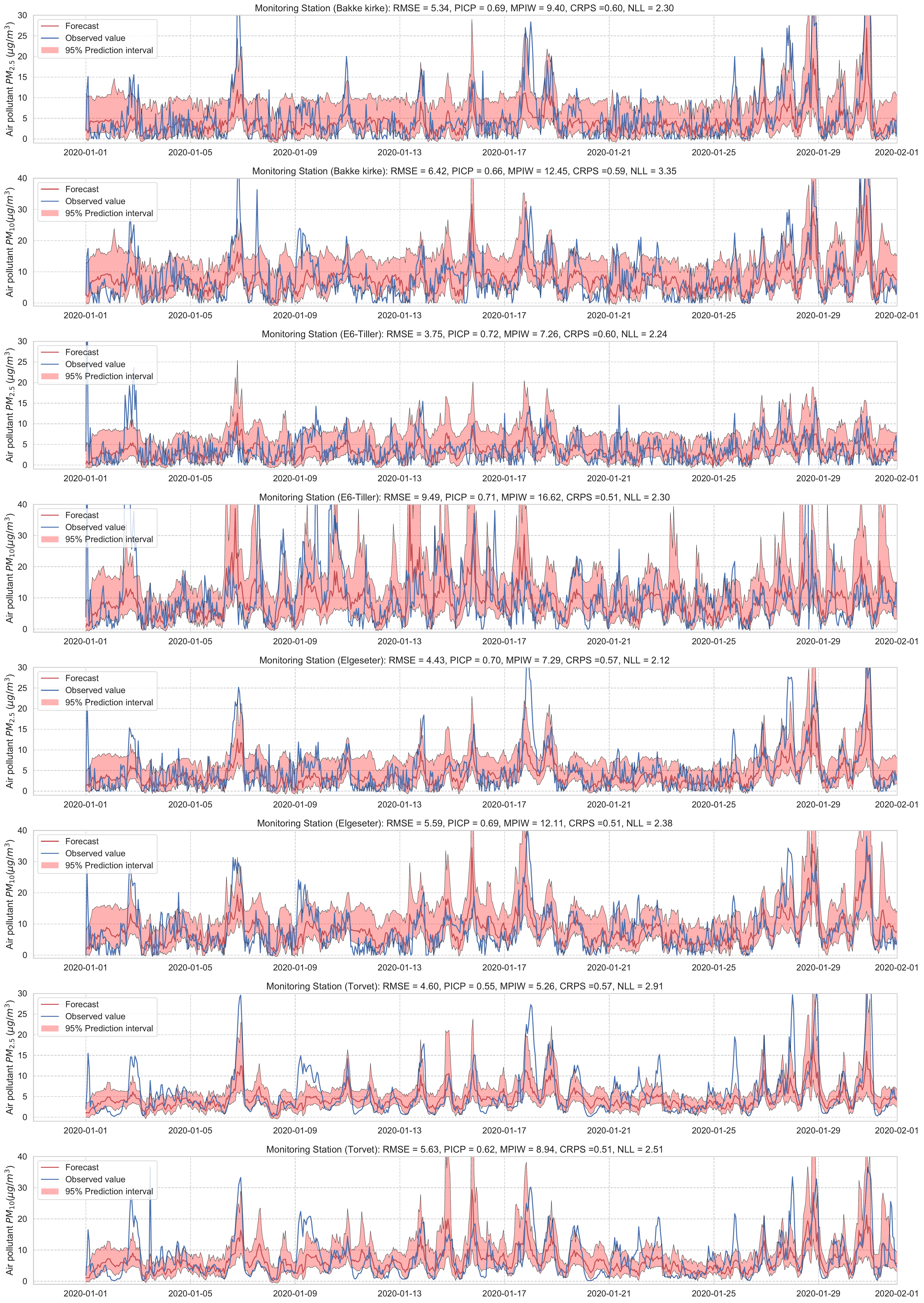} }}%
    \qquad
    \subfloat[\centering Threshold exceedance classification]{{\includegraphics[width=0.42\linewidth]{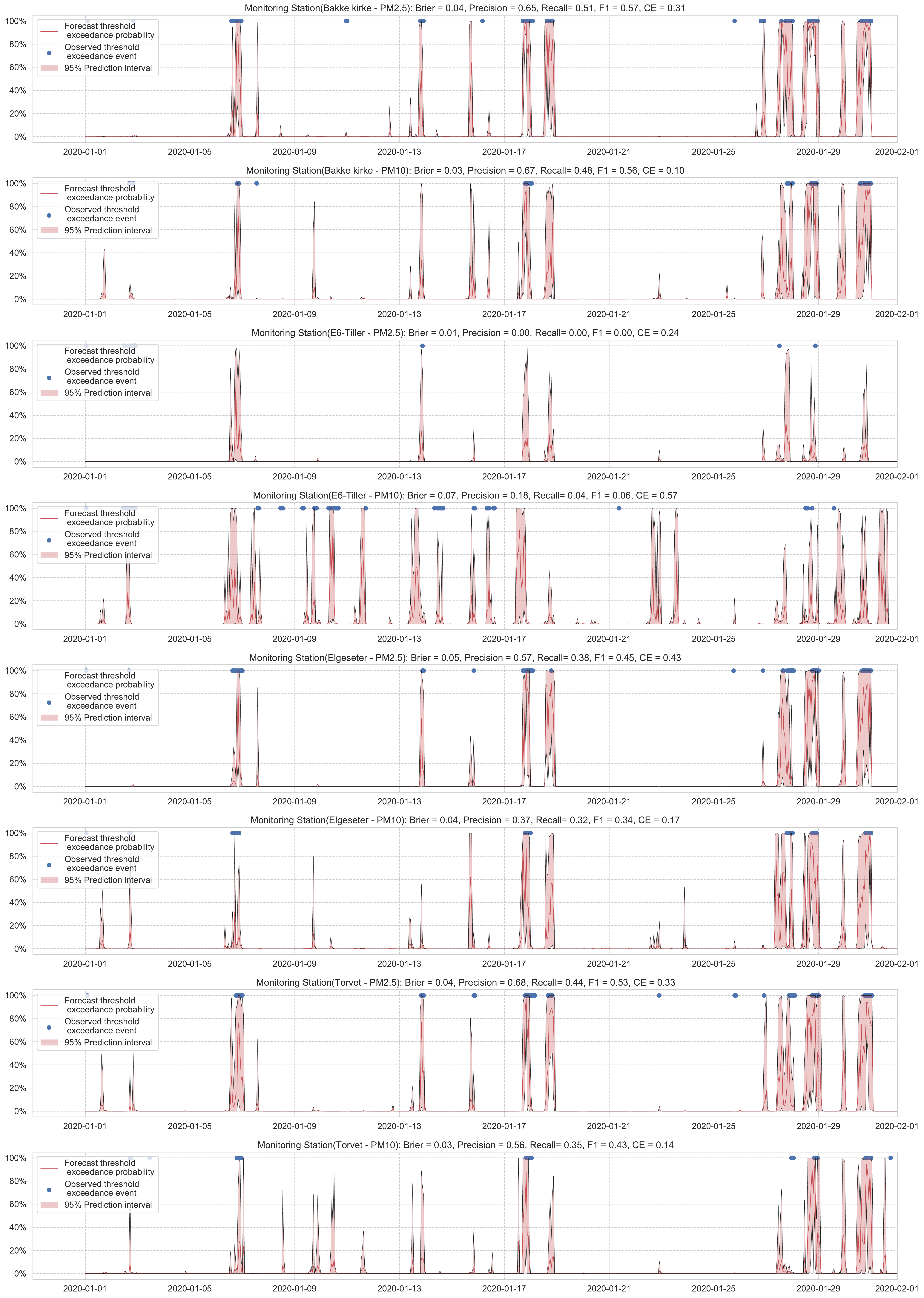} }}%
    \caption{Deep probabilistic forecast of air quality using NNs with MC dropout at four monitoring stations.}%
    \label{fig:appendix_mc}%
\end{figure}

\clearpage
\vspace{-0.8cm}
\subsection{Deep Ensemble}
\label{sec:appendix_ensemble}
\vspace{-0.8cm}

\begin{figure}[H]
\centering
    \subfloat[\centering PM-value regression] {{\includegraphics[width=0.45\linewidth]{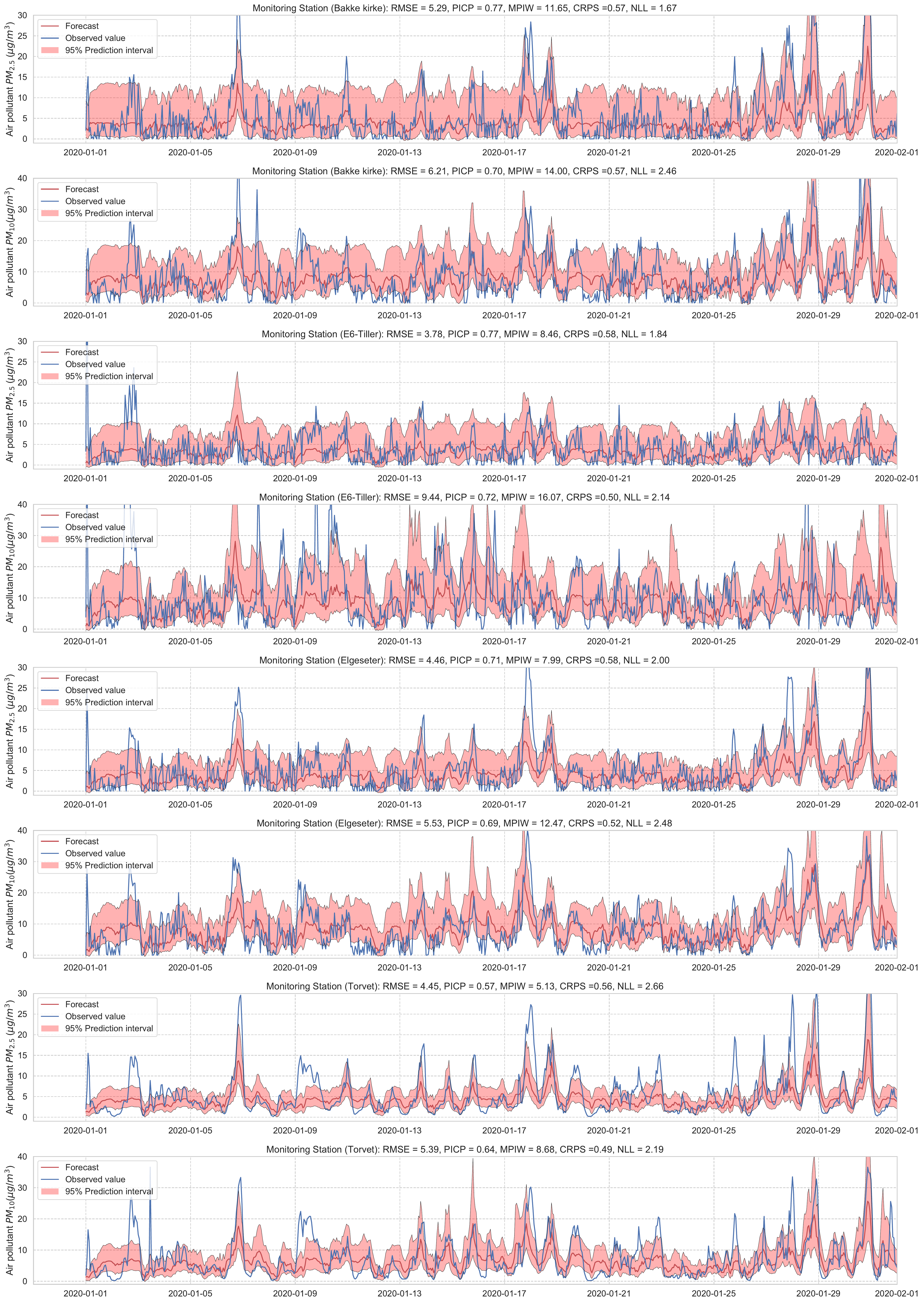} }}%
    \qquad
    \subfloat[\centering Threshold exceedance classification]{{\includegraphics[width=0.45\linewidth]{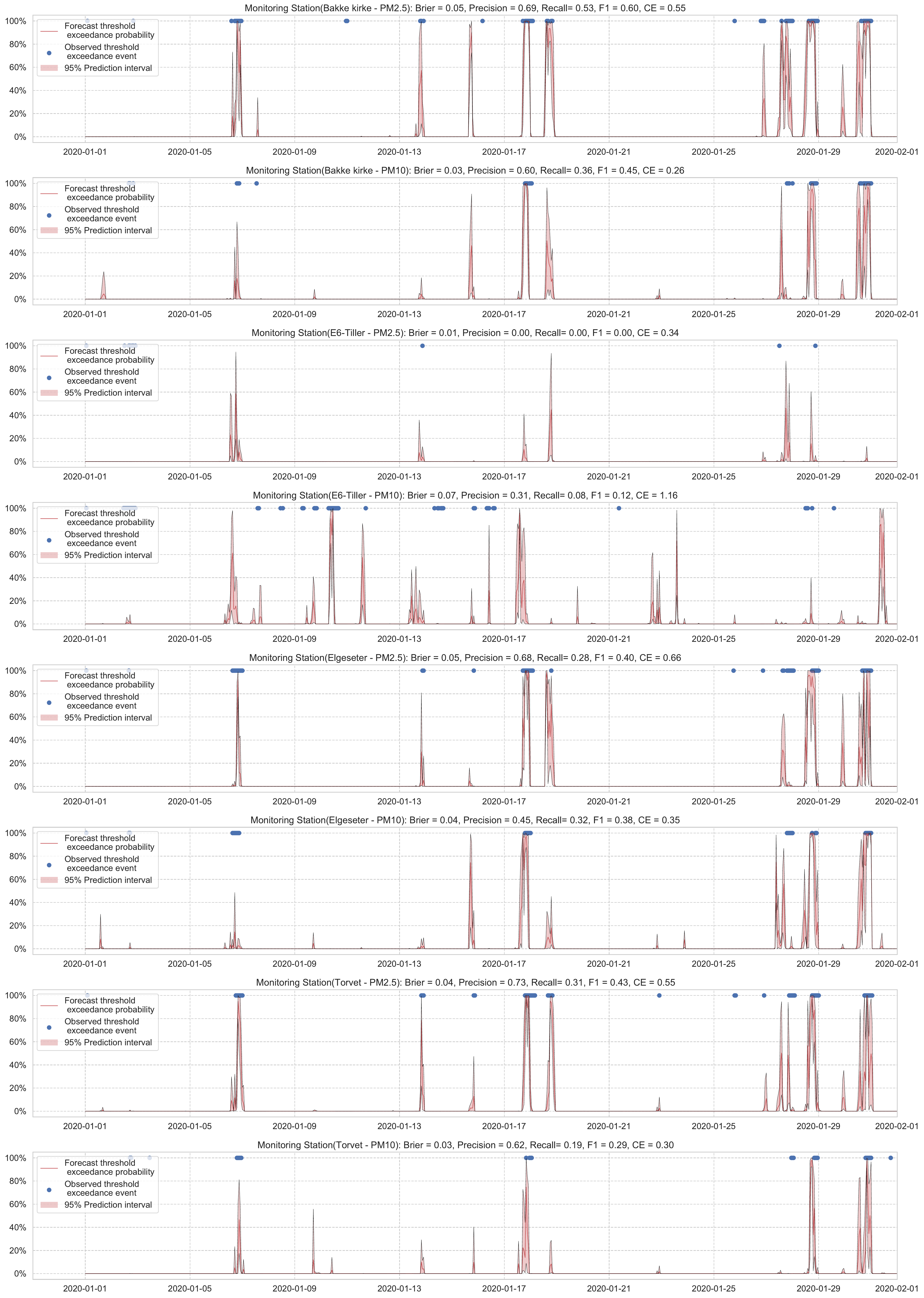} }}%
    \caption{Deep probabilistic forecast of air quality using deep ensemble at four monitoring stations.}%
    \label{fig:appendix_ensemble}%
\end{figure}

\vspace{-0.4cm}
\subsection{LSTM with MC Dropout}
\label{sec:appendix_lstm}
\vspace{-0.8cm}
\begin{figure}[H]
\centering
    \subfloat[\centering PM-value regression] {{\includegraphics[width=0.45\linewidth]{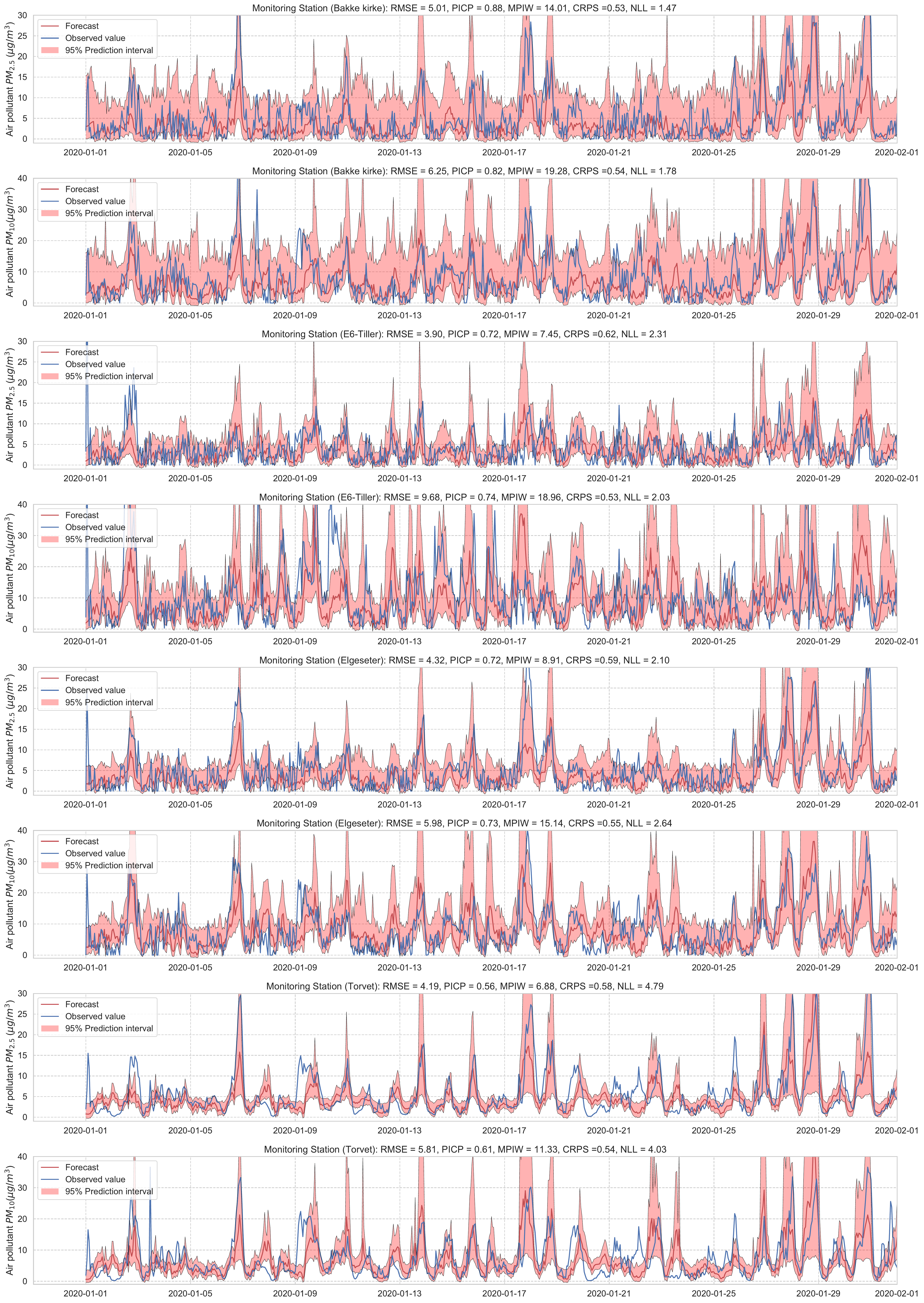} }}%
    \qquad
    \subfloat[\centering Threshold exceedance classification]{{\includegraphics[width=0.45\linewidth]{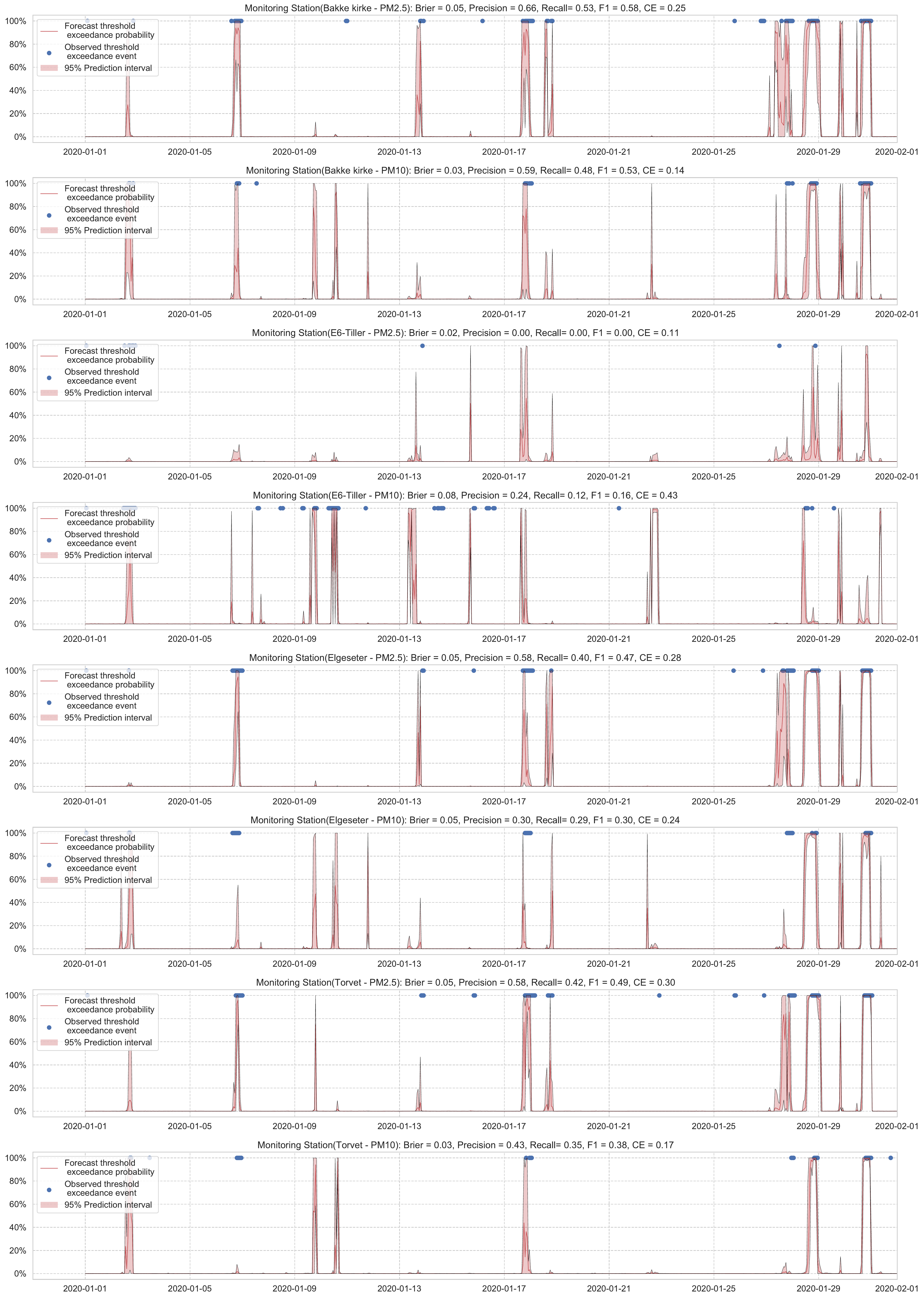} }}%
    \caption{Deep probabilistic forecast of air quality using LSTM with MC dropout at four monitoring stations.}%
    \label{fig:appendix_lstm}%
\end{figure}

\clearpage

\vspace{-0.8cm}
\subsection{GNNs with MC Dropout }
\label{sec:appendix_gnn}
\vspace{-0.8cm}
\begin{figure}[H]
\centering
    \subfloat[\centering PM-value regression] {{\includegraphics[width=0.45\linewidth]{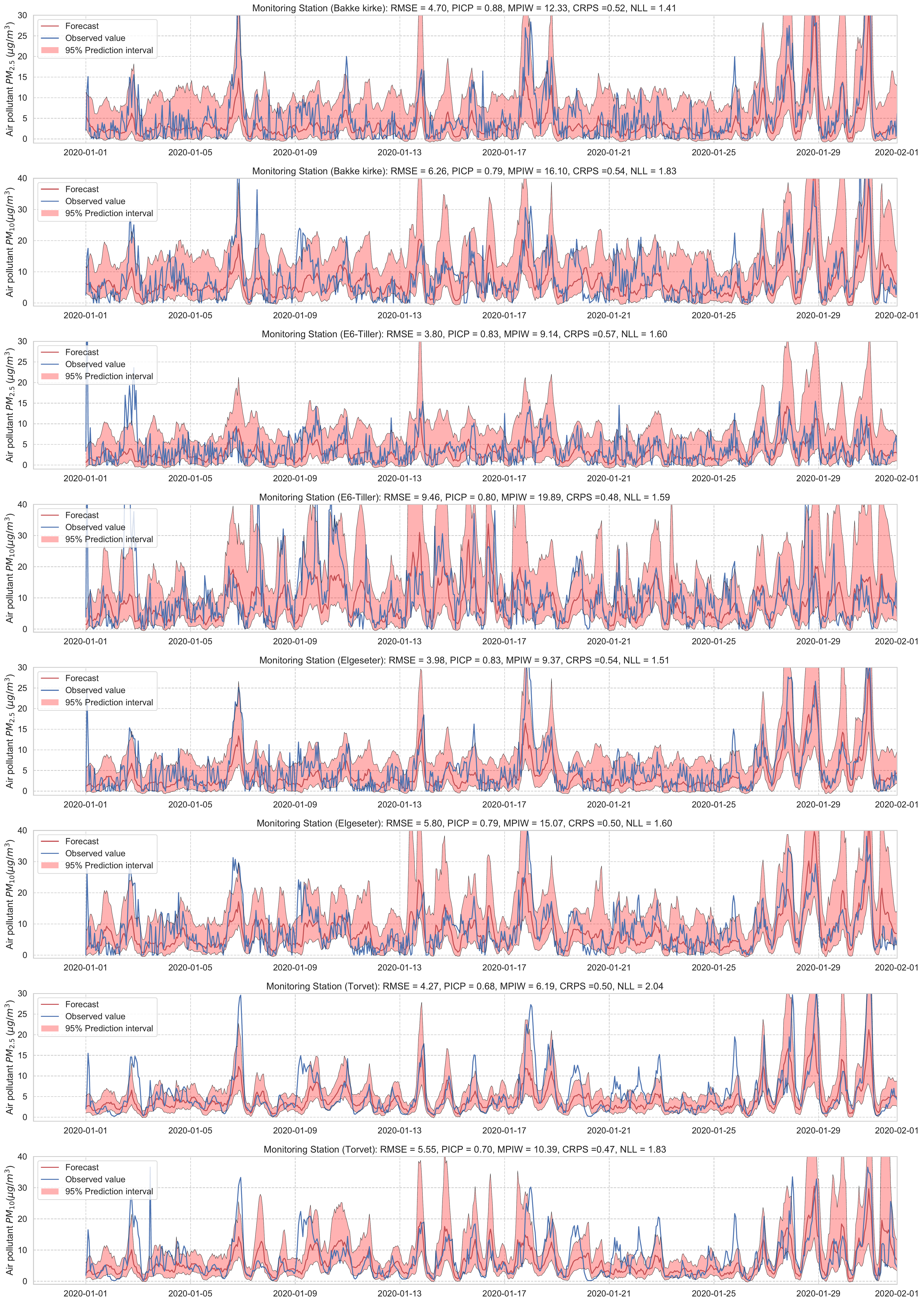} }}%
    \qquad
    \subfloat[\centering Threshold exceedance classification]{{\includegraphics[width=0.45\linewidth]{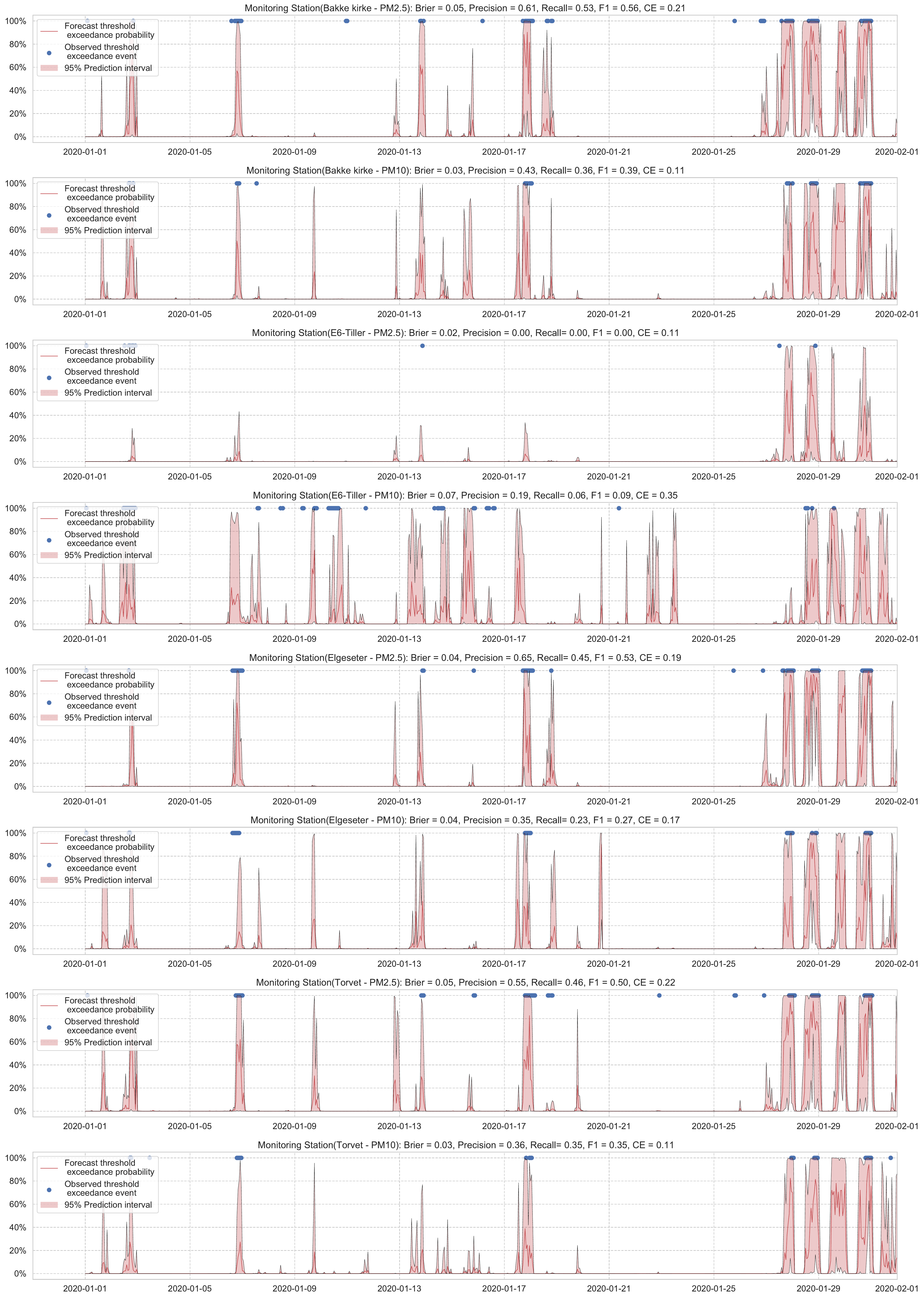} }}%
    \caption{Deep probabilistic forecast of air quality using GNNs with MC dropout at four monitoring stations.}%
    \label{fig:appendix_gnn}%
\end{figure}
\vspace{-0.4cm}
\subsection{SWAG}
\label{sec:appendix_swag}
\vspace{-0.8cm}

\begin{figure}[H]
\centering
    \subfloat[\centering PM-value regression] {{\includegraphics[width=0.45\linewidth]{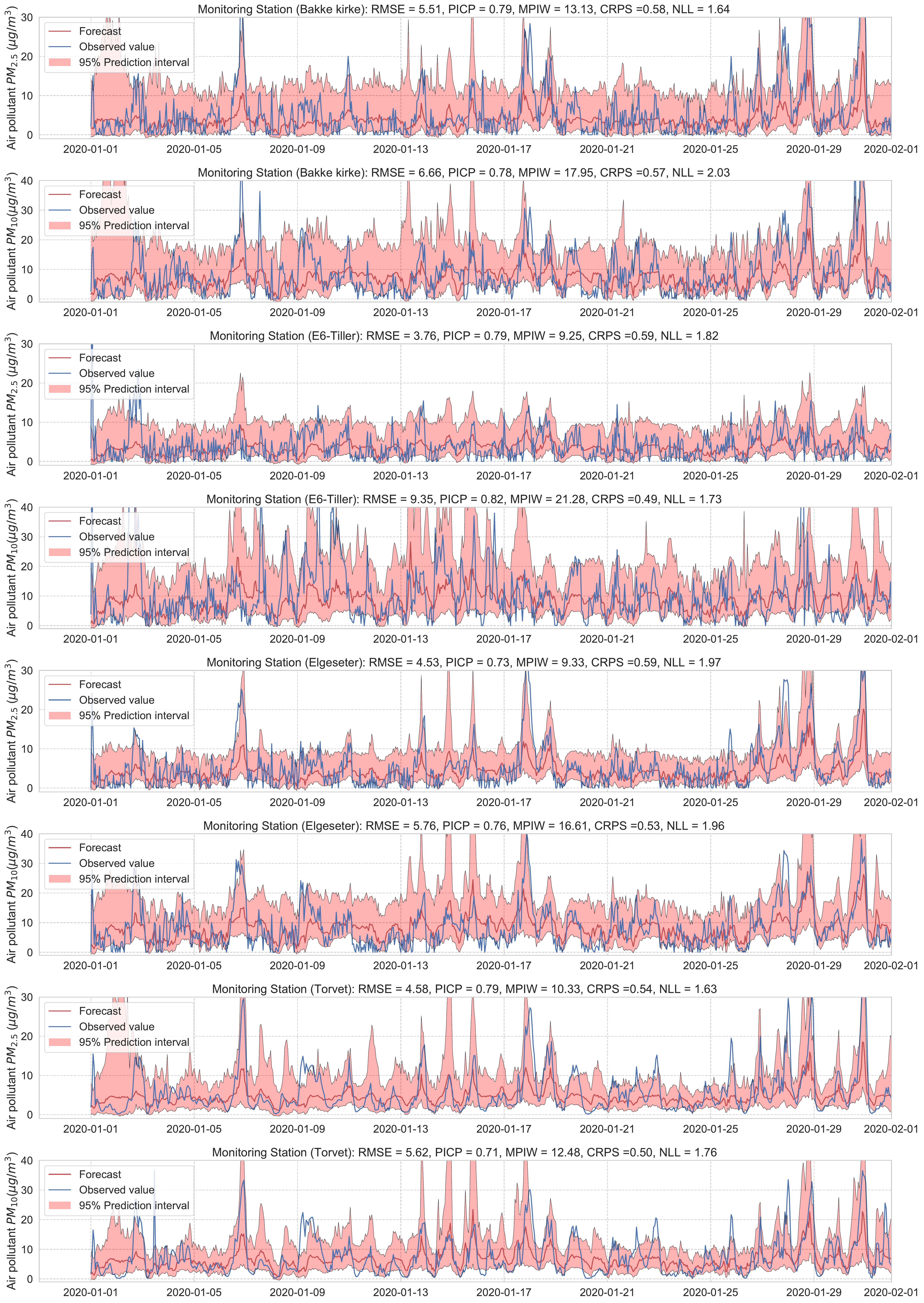} }}%
    \qquad
    \subfloat[\centering Threshold exceedance classification]{{\includegraphics[width=0.45\linewidth]{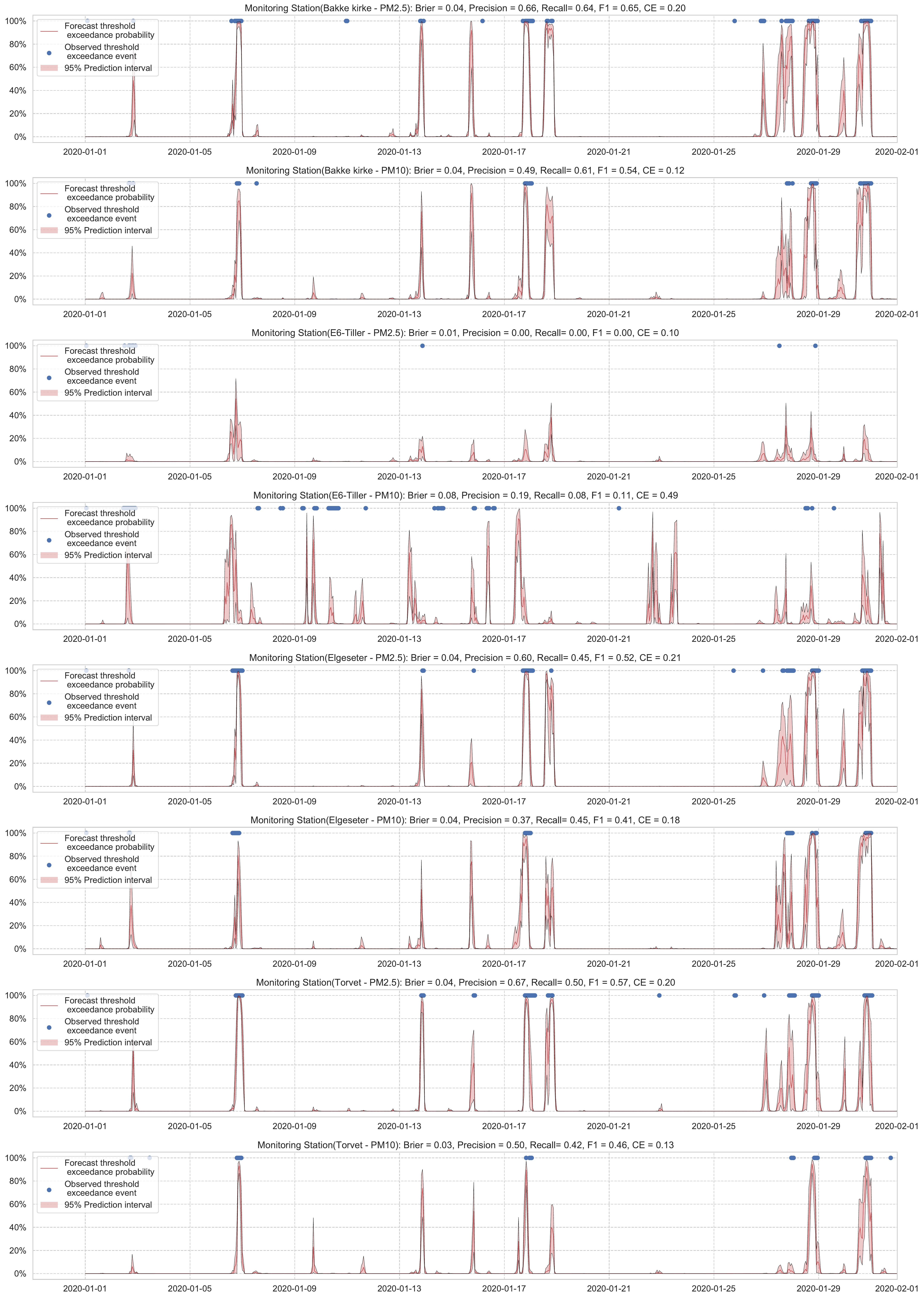} }}%
    \caption{Deep probabilistic forecast of air quality using a SWAG model at four monitoring stations.}%
    \label{fig:appendix_gnn}%
\end{figure}

\end{document}